\documentclass[pdflatex,sn-mathphys-num]{sn-jnl}

\usepackage{graphicx}%
\usepackage{multirow}%
\usepackage{amsmath,amssymb,amsfonts}%
\usepackage{amsthm}%
\usepackage{mathrsfs}%
\usepackage[title]{appendix}%
\usepackage{xcolor}%
\usepackage{textcomp}%
\usepackage{manyfoot}%
\usepackage{booktabs}%
\usepackage{algorithm}%
\usepackage{algorithmicx}%
\usepackage{algpseudocode}%
\usepackage{listings}%

\usepackage{subfiles}
\usepackage{cleveref}
\usepackage[section]{placeins}

\newcommand{\FIG}[1]{\cref{#1}}
\newcommand{\TABLE}[1]{\cref{#1}}
\newcommand{\EQUA}[1]{\cref{#1}}
\newcommand{\SECTION}[1]{\cref{#1}}

\newcommand{\MODEL}{Unnamed Model}

\newcommand{\ETAL}{{\emph{et al.}}}
\newcommand{\EG}{{\emph{e.g.}}}
\newcommand{\IE}{{\emph{i.e.}}}

\newcommand{\YNET}{$\textsf{Y}$-net}
\newcommand{\YNETCITE}{\cite{mangalam2020s}}

\newcommand{\SPECTGNN}{SpecTGNN}
\newcommand{\SPECTGNNCITE}{\cite{cao2021spectral}}

\newcommand{\MSN}{MSN}
\newcommand{\MSNCITE}{\cite{wong2021msn}}

\newcommand{\SEEM}{SEEM}
\newcommand{\SEEMCITE}{\cite{wang2022seem}}

\newcommand{\NSPSFM}{NSP-SFM}
\newcommand{\NSPSFMCITE}{\cite{yue2022human}}
\newcommand{\MUSEVAE}{MUSE-VAE}
\newcommand{\MUSEVAECITE}{\cite{lee2022muse}}

\newcommand{\EVMODEL}{E-V$^2$-Net}

\newcommand{\EVCITE}{\cite{wong2023another}}
\newcommand{\EQMOTION}{EqMotion}
\newcommand{\EQMOTIONCITE}{\cite{xu2023eqmotion}}

\newcommand{\FLOWCHAIN}{FlowChain}
\newcommand{\FLOWCHAINCITE}{\cite{maeda2023fast}}
\newcommand{\IMP}{IMP}
\newcommand{\IMPCITE}{\cite{shi2023representing}}
\newcommand{\SCMODEL}{SocialCircle}
\newcommand{\SCCITE}{\cite{wong2023socialcircle}}

\newcommand{\LGTRAJ}{LG-Traj}
\newcommand{\LGTRAJCITE}{\cite{chib2024lg}}
\newcommand{\RAN}{RAN}
\newcommand{\RANCITE}{\cite{dong2024recurrent}}

\newcommand{\UPDD}{UPDD}
\newcommand{\UPDDCITE}{\cite{liu2024uncertainty}}
\newcommand{\SCPMODEL}{SocialCircle+}
\newcommand{\SCPCITE}{\cite{wong2024socialcircle+}}

\newcommand{\REMODEL}{\emph{Resonance}}
\newcommand{\RECITE}{\cite{wong2024resonance}}

\newcommand{\C}[1]{{\textcolor[HTML]{0000FF}{#1}}}
\newcommand{\R}[1]{{\textcolor[HTML]{FF0000}{#1}}}

\newcommand{\TODO}[1]{\colorbox{yellow}{TODO}}
\newcommand{\hatbf}[1]{\hat{\mathbf{#1}}}

\newcommand{\NEWHALFLINE}{\vspace{0.5em}\noindent}

\newcommand{\NEWMODEL}{\emph{Socialality}}
\newcommand{\NEWKERNEL}{{Socialality}}
\newcommand{\ANCHOR}{\emph{socialality}}
\renewcommand{\MODEL}{GPCC}
\renewcommand{\FIG}[1]{Fig. \ref{#1}}
\renewcommand{\TABLE}[1]{Table. \ref{#1}}

\theoremstyle{thmstyleone}%
\theoremstyle{thmstyletwo}%

\theoremstyle{thmstylethree}%

\begin{document}

\title[Article Title]{Socialality Anchors: Towards Group-bounded Trajectory Prediction}


\author[]{\fnm{Ziqian} \sur{Zou}}\email{ziqianzoulive@icloud.com}

\author[]{\fnm{Conghao} \sur{Wong}}\email{conghaowong@icloud.com}

\author[]{\fnm{Qinmu} \sur{Peng}}\email{pengqinmu@hust.edu.cn}

\author*[]{\fnm{Xinge} \sur{You*}}\email{youxg@mail.hust.edu.cn}

\affil*[]{\orgname{Huazhong University of Science and Technology}, \orgaddress{\city{Wuhan}, \postcode{430074}, \state{Hubei}, \country{China}}}




\abstract{
Trajectory prediction is a key component for understanding human behavior patterns in dynamic scenes.
Researchers have devoted substantial efforts to modeling social interactions, especially group-wise interactions, since group membership often reflects shared intention, coordinated motion, and stable mutual adaptation, thus providing a persistent and semantically meaningful social prior for forecasting.
However, existing group modeling methods may rely on a fixed threshold and infer groups mainly from agents' relative positions within the observation window, overlooking the fact that grouping rules should be agent-specific, temporally coherent, and context-adaptive across diverse personalities, culturalities, and evolving interaction contexts.
Inspired by human social perception that alternates between interpersonal distance in boundary-sensitive situations and relative speed consistency in dynamic interactions, we propose \emph{Socialality}, a human-inspired trajectory prediction framework with interpretable \emph{socialality} anchors and an extended grouping window for stable, context-aware grouping inference.
Concretely, \emph{Socialality} introduces a duo-scalar-controlled grouping kernel Socialality that jointly leverages historical observations and short-term future trajectory previews to learn agent-specific grouping rules, and employs a group-wise perception mechanism to model in-group and out-of-group interactions in an intuitive and explainable manner.
Furthermore, we conduct extensive experiments on standard benchmarks to demonstrate the performance gains of \emph{Socialality}, and provide qualitative analyses and statistical studies of anchor distributions to verify the interpretability and stability of the proposed \emph{socialality} anchors.
Code repo: \url{https://github.com/LivepoolQ/Socialality}
    }




\maketitle

\section{Introduction}
\label{sec_introduction}

Understanding and forecasting intelligent agents' behaviors in dynamic scenes is a critical capability for human cognition and for many vision-based intelligent systems. 
Trajectory prediction, as a representative task, aims to forecast socially acceptable future trajectories for each target agent from a short history of observations, while accounting for the temporal regularities of motion and the potential interactions among surrounding agents \cite{alahi2016social}. 
Such forecasting ability supports a broad range of applications, including behavior analyses \cite{alahi2017learning,chai2019multipath}, navigation and planning \cite{lee2017desire}, transportation and autonomous driving \cite{chen2022scept}, as well as detection and multi-target tracking \cite{pellegrini2009youll,saleh2020artist}.

\begin{figure}[h]
\centering
\includegraphics[width=1.0\textwidth]{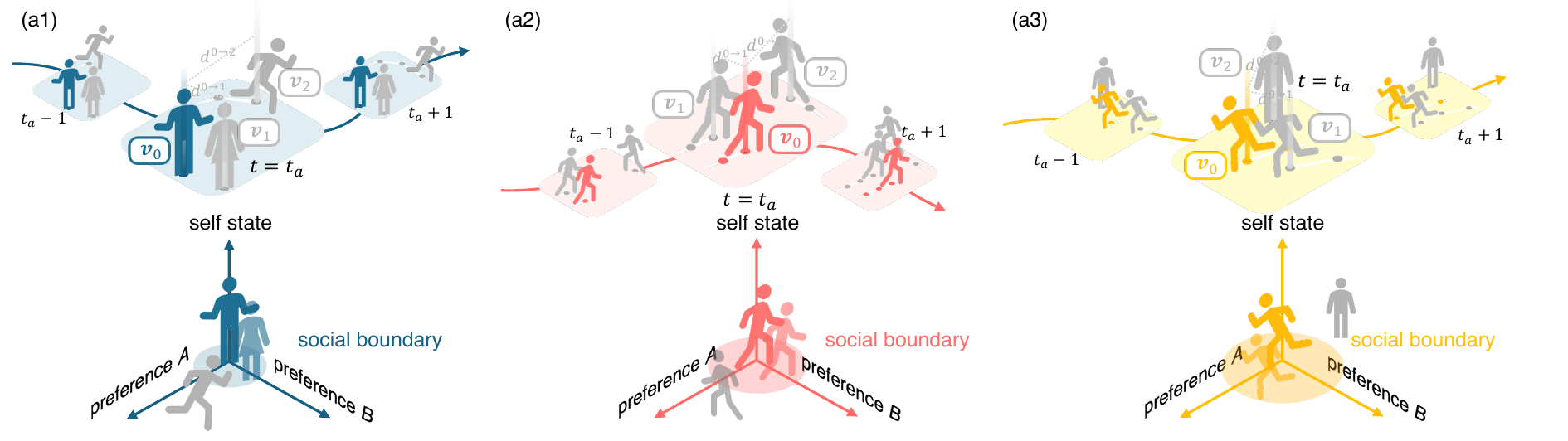}
\caption{
    By observing the agent-wise preferences when grouping with others over a period of time, we mainly focus on how to infer each target agent's social boundary, which will anchor its group affiliation.
        }\label{fig_intro}
\end{figure}

Despite recent progress, trajectory prediction remains challenging in crowded scenes \cite{gupta2018social}, where explicit grouping phenomenon could be widely observed.
Empirical observations show that 55-70\% of pedestrians walk in groups of two or more members in crowds\cite{moussaid2010walking}.
In this manuscript, we define group affiliation for a target agent as which neighboring agent is considered to be a group member of this target agent \footnote{Different from some other group-based methods \cite{qiu2010modeling,bae2022learning}, it should be noted that in our formulations the group affiliation is asymmetric, which means that agent $i$ considering agent $j$ as a group member does not necessarily mean vice versa.}.
When forecasting the target agent's trajectory in crowded scenes, its group affiliation provides an important prior for how it interacts with surrounding agents \cite{moussaid2010walking,bae2022learning}.
For example, group members tend to adjust their speeds and directions jointly when avoiding obstacles, rather than going separate ways.
We refer to such interactions conditioned by the group affiliation as the \emph{group-bounded} social interactions.
Modeling group-bounded social interactions between pedestrians has long been a central theme in this field. 

To explicitly determine a target agent's group affiliation, a natural thought is to infer its \emph{social boundary} \cite{lamont2002study}, which defines the acceptable range of viewing neighboring agents that are regarded as group members and reflects the target agent's interaction preference with others.
However, it should take a period of time to infer such social boundary or interaction preference, such as acceptable interpersonal distance and preferred motion pattern.
Moreover, proxemics studies suggest that social boundaries often vary with different agents from diverse cultural backgrounds and are context-dependent\cite{hall1973hidden,sorokowska2017preferred}.
Considering the difficulty of modeling such heterogeneous social boundaries of diverse agents, most existing trajectory prediction methods choose to skip the social boundary inferring and directly model social interactions among pedestrians before forecasting.
However, this modeling shortcut neglects the priors provided by explicit group affiliations \emph{anchored} by social boundaries, flattening the distinction between social interactions among in-group pedestrians and those among pedestrians from different groups.
This suggests that trajectory prediction requires structural priors provided by explicitly modeling pedestrians' agent-specific, context-adaptive, and temporally coherent social boundaries to anchor their specific group affiliations.

Specifically, agent-specific indicates that social boundaries should be explicitly established in an agent-wise manner, rather than being embedded in node-level or feature-level representations of each agent.
Context-adaptive indicates the adaptability of even the same agent's social boundary, which could be completely different under diverse interaction contexts or in-group relations. 
Temporally coherent indicates that such social boundaries around the current time step are not instantaneously determined from isolated observations, but inferred from continuous motion consistency over a period of time, including observation and anticipation.

Current researchers have attempted to model such social boundaries or groups in two main ways, representing each agent with different nodes in graph-based methods or setting certain grouping rule or train additional networks to determine group membership.
On the one hand, some graph-based methods use a leveled graph or a directed graph \cite{bae2022learning,xu2022groupnet} to model different group structures while handling social interactions simultaneously among agents.
However, in many graph-based formulations, edges mainly encode pair-wise interaction intensity or latent relational dependency, which is utilized as the evidence of whether two corresponding agents belonging to the same group.
The instantaneous interaction strength or latent relational dependency should not be regarded equal to being in the same group, since simply avoiding fast-moving bypassing pedestrian obviously indicates considerable interactions strength.
Moreover, though these methods could indeed represent certain group-like structures through graph-based message passing or latent relation learning, such structures could only be implicitly reflected in these representations, serving as a by-product of the modeling of social interactions rather than a dedicated inference target. 
As a result, these methods may succeed in describing who interacts with whom, but they do not explicitly explain what social boundaries anchor different groups.

On the other hand, some methods try to represent groups as explicit structures.
Among these methods, some rely on manually annotated group ground truths to additionally train grouping networks \cite{yamaguchi2011who,solera2015socially}, which output group structures first before final trajectory prediction training or evaluation.
Such annotated methods take lots of effort and largely rely on annotators' own subjective judgments of whether two agents belong to the same group.
Our conference paper, \MODEL~\cite{zou2024who}, proposes an end-to-end training paradigm without human annotation process, which determines group membership through a long-term distance kernel according to a manually set grouping rule. 
Such group membership results provide grouping priors for the subsequent perception and interaction modeling process.
However, its fixed threshold implicitly assumes that a universal social boundary is shared by all agents. 
Moreover, \MODEL's grouping assignment is purely observation-based, relying only on historical motion cues while ignoring anticipation of short-term future tendencies. 
These assumptions overlook the agent-specific, context-adaptive, and temporally coherent nature of social boundaries in real crowded scenes.

To address the above requirements, as shown in \FIG{fig_intro}, interpreting different agents' specific and context-adaptive preferred social boundaries to explicitly anchor pedestrian groups with temporal coherence has become the main consideration of this manuscript.
Considering trajectory prediction as a human-centric task, it could be much easier to establish social boundaries in a human-inspired way that simulates or emulates how humans perceive social context.
In social psychology \cite{campbell1958common}, perceived groupness is often characterized by entitativity cues, including proximity, similarity, and common fate, which connects the social boundary requirements mentioned above.
Here, the agent-specific requirement aligns with proxemics studies suggesting that interpersonal distance is shaped by both personal preferences and cultural norms \cite{sorokowska2017preferred}. 
For example, people from different cultural backgrounds may maintain notably different distances in public walking scenarios, distinctively forming different privacy boundaries.
The context-adaptive requirement also aligns with human intuition.
For example, when walking with colleagues or friends, pedestrians may preserve a moderate interpersonal distance and move relatively efficiently towards a shared destination. 
When walking with intimate companions, they may naturally keep a closer distance and move at a more relaxed pace. 
These suggest that social boundaries are conditioned on the agent-level social contexts, \IE, social relations, roles, and coordination patterns.
Finally, the time-coherent requirement further ensures the social boundaries are modeled in a human-inspired manner.
For example, a pair of agents may appear close over the whole observation window but show a tendency of gradual separation near the current time, suggesting that their group relation may be weakening.
Similarly, agents that gradually approach and align their motion patterns may be forming a group, adopting the other agent into its own boundary space.
In other words, it means that such boundary preference requires a ``temporal thickness'' \cite{husserl2012phenomenology}, \IE, it requires both observation and anticipation to finalize.

Similar to physical or territorial boundaries that are often anchored through observable indicators, as shown in \FIG{fig_intro}, social boundaries can also be inferred from behavioral cues that reveal how pedestrians maintain and adjust their grouping relations with others.
Motivated by the social psychology theories mentioned above, we introduce a set of learnable group-bounded anchors, referred to as \ANCHOR~anchors, to simulate such grouping cues, which enables the model to learn and capture different agents' various ways of anchoring their groups before making trajectory decisions.
In relatively static or boundary-sensitive situations, interpersonal distance is often prioritized, because spatial proximity serves as an immediate cue of social boundary and affiliation in proxemics and group perception \cite{hall1973hidden}. 
By contrast, in dynamic interaction scenes, humans tend to rely more on relative speed or motion consistency to infer whether multiple agents share a common fate or collective intention \cite{moussaid2010walking}.
Such human-inspired observations motivate us to treat distance and speed as the primary behavioral factor for learning different anchors.
These anchors could map various boundary preferences into quantified scalars, each of which is supposed to represent a single property for forming or keeping agents in a group, thus finally modulating and customizing their own unique social boundaries.
In addition, we extend the original observation-only grouping window in the long-term distance kernel to contain both observation and anticipation temporal segments to coherently capture motion tendency with temporal thickness, thus forecasting trajectories with unique grouping preferences for different agents under various interaction contexts.
In essence, the proposed group-bounded anchors transform social boundary modeling into a grouping cues interpretation problem.
Instead of justifying whether two pedestrians are close enough to be grouped, the model learns which cues should be activated to explain the group affiliation of each agent under the current context.

This manuscript is an extension of our previous conference paper \cite{zou2024who}.
The \MODEL~Model in our conference paper introduces a group-based trajectory prediction paradigm, where social interactions are handled based on the inferred group membership using the original long-term distance kernel.
However, with a manually fixed threshold in the long-term distance kernel, \MODEL's grouping rule cannot adapt to different forecasting scenes or even agent-wise social interactions.
In addition, it lacks an explicit mechanism to either represent or explain how grouping decisions remain stable when pedestrians exhibit short-term motion fluctuations, such as temporary spatial proximity or abrupt changes in motion state.
To address these issues, we propose \NEWMODEL, a human-inspired extension that introduces interpretable \ANCHOR~anchors and leverages an extended grouping window to interpret agent-specific and context-adaptive social boundaries with temporal coherence to anchor heterogeneous groups.

In summary, our contributions are listed as follows:
\begin{enumerate}
    \item We propose a \ANCHOR~anchors-controlled grouping kernel that simultaneously considers both agents' historical observations and their short-term future trajectory previews to learn to interpret agent-specific, context-adaptive social boundaries with temporal coherence, thus obtaining the group priors. 
    \item We propose the \NEWMODEL~trajectory prediction model that inherits the perception mechanism to explicitly takes the group priors interpreted from \NEWKERNEL~kernel into account to simulate group-wise social interactions modulated by the newly proposed \ANCHOR~anchors in a human-inspired way.
    \item We conduct experiments on standard benchmarks to demonstrate the performance gain of \NEWMODEL, and further provide qualitative analyses together with statistical studies of anchor distributions to verify the interpretability and stability of the proposed \ANCHOR~anchors.
\end{enumerate}

\section{Related Works}
\label{sec_relatedWorks}

\subsection{Trajectory Prediction and Social Interactions}
Trajectory prediction aims at predicting agents' future movements based on their historical states and potential interactions \cite{alahi2016social}.
Early trajectory forecasting and crowd navigation studies largely relied on hand-crafted rules, where social effects were explicitly encoded as physics-inspired constraints between agents.
The Social Force model \cite{helbing1995social} formulates pedestrian motion as the superposition of attractive and repulsive forces, yielding emergent collision avoidance and flow patterns.
In parallel, velocity-obstacle formulations \cite{fiorini1998motion,van2008reciprocal} cast interaction as geometric feasibility in velocity space, enforcing collision avoidance through reciprocal kinematic constraints.
With the development of deep learning methods, researchers gradually treat pedestrian motion as a temporal sequence generation task.
These works predict agents' future positions via recurrent architectures combined with hand-crafted rules.
Trajectory prediction models based on RNNs \cite{kim2017probabilistic,sun2020recursive} and LSTMs \cite{alahi2016social,zhang2019sr,zhang2020social} were subsequently introduced to model temporal dependencies in trajectories while implicitly accounting for inter-agent dependencies.
Among them, Social-LSTM \cite{alahi2016social} first introduced the social pooling mechanism, which aggregates neighboring hidden states to capture local social context around the target agent.
This paradigm has since been widely adopted and extended in many follow-up works \cite{deo2018convolutional,pei2019human,liu2023stagp}, serving as a foundation for learning interaction-aware representations.

More recently, graph-based and Transformer-based architectures have become mainstream for modeling interactions, enabling more flexible message passing and context aggregation across agents and scenes \cite{vemula2018social,fernando2018soft,yuan2021agentformer,zhao2020tnt,ivanovic2019trajectron,cao2020spectral,mohamed2020social}.
Recent studies have also moved towards more interpretable, human-inspired formulations of social interaction.
Drawing inspiration from biological echolocation, Wong \ETAL \cite{wong2023socialcircle,wong2024socialcircle+} propose an angle-based representation that characterizes surrounding agents through relative distance, velocity, and direction, yielding an explicit and structured interaction description.
Bae \ETAL \cite{bae2025social,bae2024can} leverage the knowledge embedded in large language models by introducing specially designed numeric tokens, enabling the model to reason about spatial relations and expose interaction cues from an alternative, language-driven perspective.
Moreover, motivated by resonance phenomena, \REMODEL~\cite{wong2024resonance} interprets social interaction as spectrum-level co-vibration among agents, providing a decomposition that helps explain the stochasticity arising in multi-agent dynamics.

\subsection{Group Modeling Methods in Trajectory Prediction}
Recent trajectory prediction studies increasingly incorporate group modeling to capture higher-level social relations beyond pairwise interactions.
Most of these group-modeling methods are graph-based.
Agents are represented as nodes, and group relations are inferred via message passing on interaction graphs, where edges encode spatial proximity or other latent relational dependencies.
Graph neural networks (GNNs) and their variants have been widely adopted to aggregate neighbor information and to form group-aware representations for forecasting \cite{bae2022learning}.
To better reflect collective structures, some methods employ multi-scale graphs or hypergraph-style formulations \cite{xu2022groupnet}, enabling relation reasoning at both individual and group levels, and thus modeling collective motion patterns in a unified architecture.

Despite their effectiveness, existing graph-based group modeling methods still face several limitations when viewed from the perspective of social boundary establishment.
On the one hand, their grouping inference is typically implicit and tightly coupled with representation learning, where group membership and grouping rules are often hidden in node features, edge weights, or message-passing processes rather than being explicitly exposed as agent-wise social boundaries.
On the other hand, since group relations are inferred from local observations or instantaneous interaction graphs, the resulting group structures may be unstable over time and sensitive to short-term motion fluctuations.
These limitations motivate alternative designs that explicitly establish lightweight, interpretable, and temporally coherent social boundaries, thereby providing more reliable group priors for subsequent interaction modeling.

\subsection{Human-inspired Methods in Trajectory Prediction}
Pedestrian trajectory prediction is a human-centric task \cite{jiang2025survey}, as it aims to forecast human motion in shared environments.
Accordingly, beyond purely data-driven forecasting, a line of works explicitly draws inspiration from human perception and cognition to build more interpretable and human-like prediction mechanisms.
Some approaches mimic how humans selectively attend to relevant neighbors and regions to filter and weight social cues, thereby making interaction reasoning more structured and interpretable \cite{wong2023socialcircle,sadeghian2019sophie}.

Moreover, some human-inspired trajectory prediction methods are motivated by how humans use their sensors to perceive the world.
Consequently, instead of treating all surrounding agents as equally observable, a number of methods incorporate view-based mechanisms to approximate how humans perceive social cues.
For example, Hasan \ETAL \cite{hasan2018seeing} leverage the visual frustum of attention to emphasize that head orientation can serve as an explicit prior for deciding which neighbors are likely to influence the future motion.
Similarly, Liao \ETAL \cite{liao2024human} introduce an adaptive visual sector that dynamically allocates attention over angular regions according to motion states, enabling the model to filter and weight social information in a more human-like way.
In this human-inspired spirit, the perception mechanism in \MODEL~also adopts an ego-centric, observation-driven design to modulate which social cues are emphasized when forming interaction representations, aligning grouping and interaction reasoning with the target agent's local perceptual context.
Moreover, pedestrian motion is largely shaped by intention and goal-directed planning \cite{rehder2015goal}.
This motivates goal-driven forecasting, which explicitly models destinations as structured variables and conditions when forecasting trajectories \cite{guo2024goal,wang2022stepwise,chiara2022goal,mangalam2020s}.

~\\
In summary, existing group-modeling methods have shown the importance of groupness in trajectory prediction.
However, most of them infer group relations implicitly through graph construction or message passing, where group membership is usually hidden in learned representations rather than explicitly established as agent-wise social boundaries.
Therefore, they may provide less transparent characterization of agent-specific and context-dependent grouping preferences, especially when distinguishing in-group companions from nearby but out-group pedestrians.
Meanwhile, human-inspired studies have introduced selective perception and goal-directed planning into trajectory prediction, but how to explicitly establish temporally coherent social boundaries from human-centric grouping cues remains to be addressed.
Built upon the perception mechanism in \MODEL~Model, our enhanced \NEWMODEL~Model takes a step further by introducing learnable \ANCHOR~anchors to capture diverse grouping preferences and establish explicit group-bounded social boundaries.
This design yields lightweight, interpretable, and stabilizable group priors, leading to more effective group-aware interaction reasoning and improved trajectory forecasting.

\begin{figure}[t]
\centering
\includegraphics[width=1.0\textwidth]{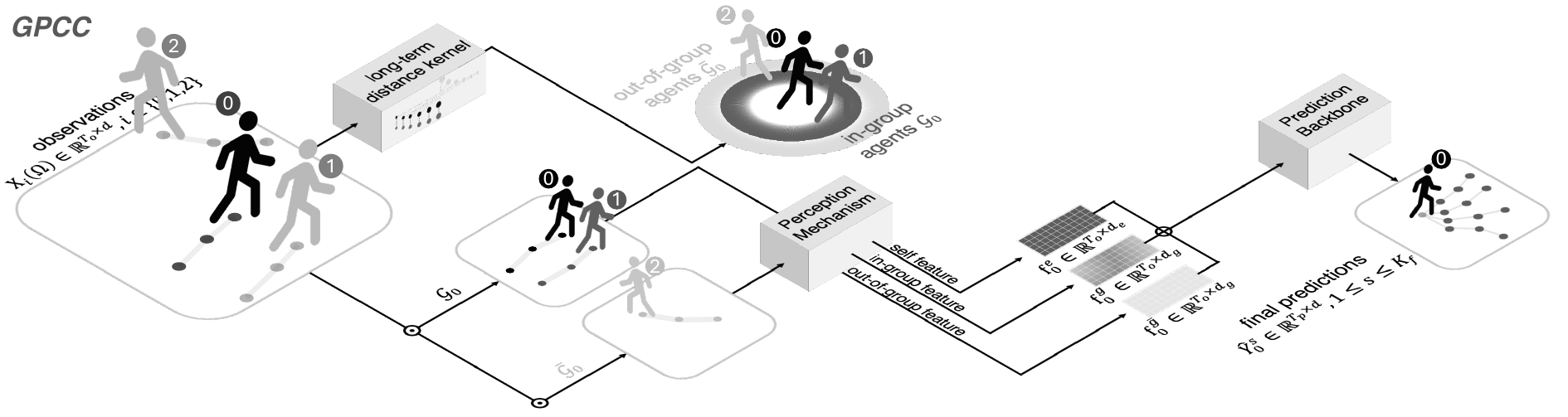}
\caption{ 
    \MODEL~method illustration.
    \MODEL~uses the long-term distance kernel and the perception mechanism to model group-wise social interactions.
        }\label{fig_method_GPCC}
\end{figure}

\section{Method}
\label{sec_method}
\begin{figure}[h]
\centering
\includegraphics[width=1.0\textwidth]{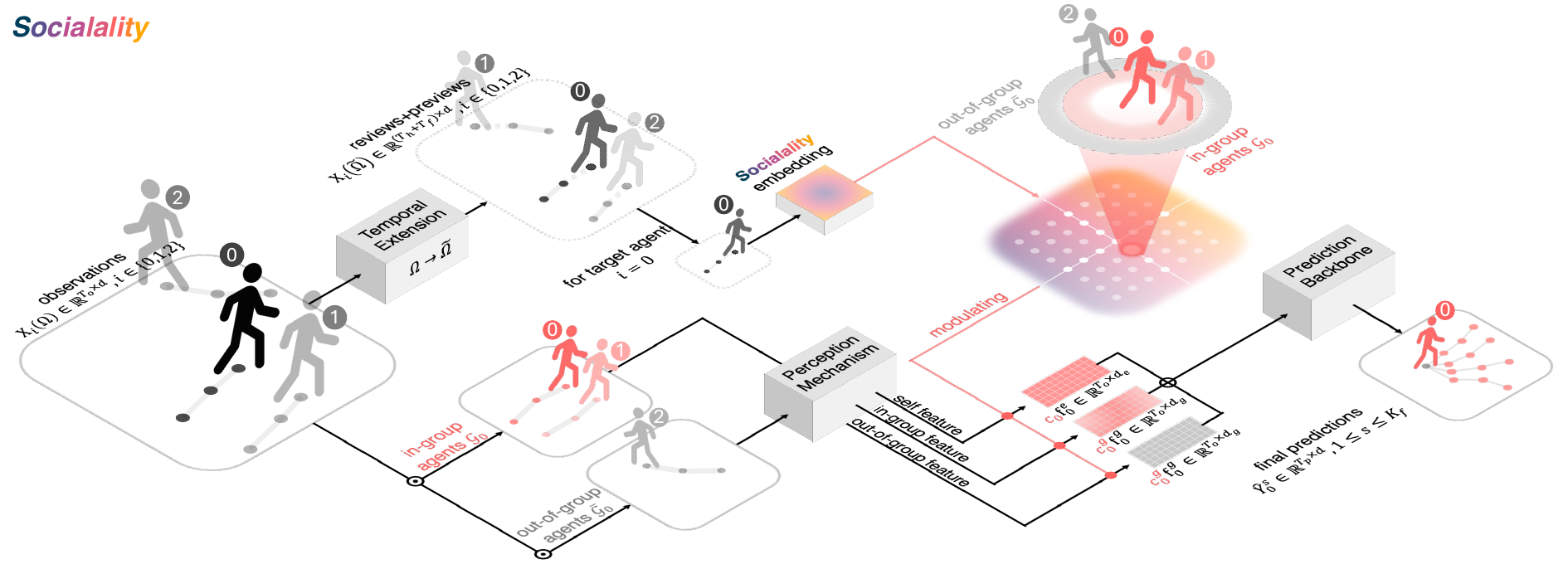}
\caption{ 
    \NEWMODEL~method illustration.
    We use a forecasting scene with three walking agents to illustrate the overall modeling.
    A social group of agent $0$ and agent $1$ can be observed in this case.
    }\label{fig_method}
\end{figure}

In our conference paper \cite{zou2024who}, the proposed \MODEL~Model (short for \MODEL) aims at forecasting trajectories conditioned by social interactions with structured group priors.
In this manuscript, we propose the enhanced \NEWMODEL~Model (short for \NEWMODEL) to extend the original frozen groupings in \MODEL.
It jointly integrates \emph{retentions} (historical reviews) and \emph{protentions} (short-term future previews) corresponding to each target agent through two trainable and complementary social anchors, therefore better understanding and forecasting \emph{agent-specific}, \emph{temporally coherent}, and \emph{context-adaptive} grouping rules.
As illustrated in \FIG{fig_method_GPCC} and \FIG{fig_method}, both \MODEL~and \NEWMODEL~Models share almost the same computational pipelines, including a \textbf{grouping kernel} and a \textbf{perception mechanism}, achieving the goal of learning to group neighboring agents, as well as sensing group-wise differences in interaction preferences, or even human motions. 
This section first contrasts two grouping kernels, the long-term distance kernel and the enhanced \NEWKERNEL~kernel, then introduces the perception mechanism and the enhanced \NEWMODEL~trajectory prediction model that builds upon these kernels.

\subsection{Problem Formulation}
We denote the 2D coordinate of any agent $i$ ($i \in \{1, 2, \dots, N_e\}$) at time step $t$ as $\mathbf{p}_i^t = (x_i^t, y_i^t)\in \mathbb{R}^2$ (the current observation step is set to $t=0$ in this manuscript).
Our considered trajectory is uniformly sampled at a fixed temporal interval, and we use $t_o$ discrete observing time steps to forecast trajectories in the following $t_p$ steps.
For a clear presentation, we denote any trajectory (agent $i$) with $t_2$ steps and ends at $t=t_1$ using the symbol $\mathbf{x}_i(t_2, t_1)=\left( \mathbf{p}_i^{t_1-t_2+1}, \dots, \mathbf{p}_i^{t_1-1}, \mathbf{p}_i^{t_1}\right)^\top \in \mathbb{R}^{t_2 \times 2}$.
Correspondingly, the observed trajectory (ends at $t=0$, with $t_o$ steps) is represented as $\mathbf{X}_i \left(t_o, 0\right) = \mathbf{x}_i(t_o, 0) \in \mathbb{R}^{t_o \times 2}$.
In this manuscript, we focus on designing a network $\mathcal{N}$ to forecast possible future trajectory(ies) $\hat{\mathbf{Y}}_i \left(t_p, t_p\right) = \hat{\mathbf{x}}_i(t_p, t_p) \in \mathbb{R}^{t_p \times 2}$ (starts at $t=1$ and ends at $t=t_p$, with $t_p$ steps) for any agent $i$, using all $N_e$ agents' observed trajectories $\mathcal{X}=\left\{ \mathbf{X}_i \left(t_o, 0\right) | i = 1, 2, \dots, N_e \right\}$, \IE, $\hat{\mathbf{Y}}_i \left(t_p, t_p\right) = \mathcal{N}\left( \mathcal{X}, i \right)$.
Summaries of other symbols and network structures are listed in \TABLE{tab_both_symbol}.

\begin{table}
    \footnotesize
    \caption{
        Symbols and network structures of \MODEL~Model and the enhanced \NEWMODEL~Model
    }
    \label{tab_both_symbol}
    \centering
    \begin{tabular}{@{}lll@{}}
        \toprule
        Sym. & Descriptions & Shapes \\

        \midrule
        
        $\epsilon$         &  An infinitesimal quantity  & - \\
        $\mathbf{d}_j$ & Neighboring agent $j$'s heading direction & $(2)$ \\
        $K_f$ & Multi-style trajectory generation number in \NEWMODEL~Model& - \\
        $K_g$ & Trajectory generation number in short-term prediction network ($K_g < K_f$)& - \\
        $\hatbf{Y}_i(t_p, t_p)$ & Agent $i$'s multi-styled predicted trajectories & $(K_f, t_p, 2)$\\
        $\mathbf{Y}_i(t_p, t_p)$ & Agent $i$'s ground-truth future trajectory & $(t_p, 2)$\\
        $\hat{\mathbf{Y}}_i^l$ & Agent $i$'s linearly fitted trajectory in prediction period & $(t_p, 2)$ \\

        \midrule

        $d$         & Feature dimensions in most networks $(d=32)$  & - \\
        $l(\cdot)$         &  $\to \mathrm{fc}(2, \mathrm{ReLU}) \to \mathrm{fc}(d, \mathrm{tanh})$  & - \\
        $m(\cdot)$         &  $\to \mathrm{fc}(d, \mathrm{ReLU}) \to \mathrm{fc}(2d, \mathrm{ReLU})*2 \to \mathrm{fc}(2d * t_o, \mathrm{ReLU}) \to \mathrm{fc}(2, \mathrm{tanh})$  & - \\
        $e(\cdot)$         &  $\to \mathrm{fc}(2, \mathrm{ReLU}) \to \mathrm{fc}(d, \mathrm{tanh})$  & - \\
        $h(\cdot)$         &  $\to \mathrm{fc}(3, \mathrm{ReLU}) \to \mathrm{fc}(d, \mathrm{ReLU}) \to \mathrm{fc}(2d, \mathrm{ReLU}) \to \mathrm{fc}(d, \mathrm{tanh})$  & - \\
        $n(\cdot)$         &  $\to \mathrm{fc}(5d, \mathrm{ReLU}) \to \mathrm{fc}(4d, \mathrm{tanh})$  & - \\

        \botrule
    \end{tabular}

\end{table}

\subsection{Grouping Kernels}

For a specific agent $i$ and all its neighboring agents $\mathcal{N}_i$ ($\mathcal{N}_i \subseteq \{1, 2, \dots, N_e\}$), \IE, group candidates, grouping kernels justify whether each agent $j \in \mathcal{N}_i$ belongs to the same group as the target agent.
These kernels are the foundations of both the original \MODEL~and the proposed \NEWMODEL~Models, providing direct grouping priors for understanding and forecasting social interactions when predicting trajectories, especially those latent group behaviors.
In detail, the original \MODEL~adopts a static grouping kernel, named long-term distance kernel.
However, real-world grouping relations may vary with different agents, keep continuous between adjacent periods, and significantly differ over time.
The \NEWKERNEL~kernel is then proposed to address these limitations.
We first introduce the original long-term distance kernel, then describe how the new \NEWKERNEL~kernel will be structured.

\subsubsection{The Long-Term Distance Kernel}

Considering all time steps during an observation window $\Omega$, the original long-term distance kernel is a manual-rule controlled kernel to justify any target agent's group members.
It uses a threshold $\Gamma$ to compare the distance summation during the observation window $\Omega$ between the target agent $i$ and any agent $j$ of its neighboring agents $\in \mathcal{N}_i$.
Here, we assume that agent $j$ will be considered to be a group member of agent $i$ only if they are consistently close enough during the whole observation window $\Omega$.
This distance summation is calculated through the long-term distance function $D_i(j|\Omega)$, represented as:
\begin{equation}
    D_i(j|\Omega)
        =
        \sum_{t \in \Omega}
            \left\|
                \mathbf{p}_j^{t} - \mathbf{p}_i^{t}
            \right\|_2.
\end{equation}
Every time step is taken into account through the distance summation.
Then the manually controlled threshold $\Gamma \ge 0 $ will be statically applied on all agents and time steps to evaluate whether their distance summation exceeds the threshold $\Gamma$.
The long-term distance grouping kernel is then formulated as
\begin{equation}
    \mathcal{K}_i (j|\Omega, \Gamma)
        =
        \mathbb{I}\left[ D_i(j|\Omega) \le \Gamma \right].
\end{equation}
Here, $\mathbb{I}[\cdot]$ denotes an indicator. 
It equals to $1$ if the condition is satisfied (where the long-term distance is smaller than the given threshold $\Gamma$), and to $0$ otherwise.
The original long-term distance kernel regards spatial distance as explicit manifestation of social relations, \IE, pedestrians who remain spatially close over time are assumed to have certain social affiliation.
It helps subsequent interaction-modeling networks to evaluate the grouping effectiveness during whole observation, rather than few outlier moments or periods caused by certain social interactions such as temporary collision avoidances.
Compared to former works that utilize graph-based methods \cite{xu2022groupnet,bae2022learning} to model group relations, the long-term distance kernel not only straightforwardly conducts a lightweight computation (the distance summation), but also provides an efficient grouping rule for the subsequent grouping stage.

Specifically, in the original \MODEL~Model, the grouping stage only concerns the observation window $\mathbf{\Omega}_0=\{-t_o+1, \dots, -1, 0\}$ for grouping agents, where $t=0$ represents current time step.
For our considered target agent $i$ and the observation period $\mathbf{\Omega}_0$, $i$'s group $\mathcal{G}_i$, \IE, the set of its in-group agents selected from all its candidate neighbors $\mathcal{N}_i$, is computed as
\begin{equation}
    \mathcal{G}_i
    =
        \left\{
        j \in \mathcal{N}_i
        \,|\,
        \mathcal{K}_i (j|\,\mathbf{\Omega}_0, \Gamma)=1
        \right\}.
\end{equation}
Correspondingly, we have its out-of-group agents' set $\overline{\mathcal{G}}_i$:
\begin{equation}
    \overline{\mathcal{G}}_i
    =
        \left\{
        j \in \mathcal{N}_i
        \,|\,
        \mathcal{K}_i (j|\,\mathbf{\Omega}_0, \Gamma)=0
        \right\}.
\end{equation}
Naturally, we have $\mathcal{G}_i \cup \overline{\mathcal{G}}_i = \mathcal{N}_i$.


\subsubsection{The \NEWKERNEL~Kernel}
\label{sec_new_kernel}

The long-term distance kernel performs grouping using a manually-set fixed-distance threshold $\Gamma$.
Notably, this static threshold $\Gamma$ is used on all agents when grouping, meaning that the same grouping rule will be applied across agents and scenes.
However, grouping rules are intuitively not determined by one single variable.
On the one hand, the acceptable social distance between the same group's members often varies with time, and the group-wise acceptable social distance may also be obviously differentiated for different groups.
On the other hand, a single threshold implicitly assumes that all group decisions can be governed by the same rule, thus overlooking a wide range of grouping cases.
For example, for different target agents belonging to distinct groups, their group-averaged speeds may also be substantially adjusted for greater in-group interaction flexibility \cite{helbing1995social}.

Corresponding to the diverse \emph{personalities} and \emph{culturalities} of human beings, we propose the \NEWKERNEL~kernel to learn different target agents' \emph{agent-specific}, \emph{temporally coherent}, and \emph{context-adaptive} grouping rules, thus emulating their \emph{socialalities} for sensing and forming their own groups.
Here, \emph{Agent-specific} indicates that grouping rules will change for different target agents.
Specifically, the \NEWKERNEL~kernel explicitly learns target agent's own grouping anchors to model such agent-specific grouping rules.
\emph{Temporally coherent} indicates that grouping is perceived by target agent not only based on past observations, but also over a temporal neighborhood around the current time, thus enforcing the original grouping stage to be extended.
\emph{Context-adaptive} indicates that the grouping rules are not single-context-guided, but structured with multiple grouping anchors. 
Accordingly, the proposed \NEWKERNEL~kernel uses multiple grouping anchors on an extended grouping stage to learn the \emph{agent-specific}, \emph{temporally coherent}, and \emph{context-adaptive} grouping rules, therefore conducting robust grouping assignments.

\NEWHALFLINE\noindent\textbf{Grouping Window $\Omega$.}
To learn the temporally coherent grouping rules, the proposed \NEWKERNEL~kernel takes a grouping window shifted from observation-only window $\Omega$ to an extended temporal neighborhood window around current time $\tilde{\Omega}$ ($\Omega \mapsto  \tilde{\Omega}$) into account.
An extended grouping window $\tilde{\Omega}$ consists of an observation time window $\Omega$ and an anticipation time window $\hat{\Omega}$, \IE, $\tilde{\Omega} = \Omega \cup \hat{\Omega}$.
Trajectories during observed time window $\Omega$ are easy to obtain, as the short-term previews (future trajectories) require a simple prediction network.
Such previews aim to provide only temporally coherent grouping cues around the current time for the \NEWKERNEL~kernel, rather than extra supervision in any subsequent modeling process.
This short-term prediction network is trained by trajectories only within observation time window $\mathbf{\Omega}_0$.
Specifically, the observation time window $\mathbf{\Omega}_0$ is divided into two parts, $\mathbf{\Omega}_1=\{-(t_h+t_f)+1, \dots, -t_f\}$ and $\mathbf{\Omega}_2=\{-t_f+1, \dots, -1, 0 \}$ ($t_o = t_h + t_f$).
During the training stage, the network uses all considered agents' trajectories during $\mathbf{\Omega}_1$ as the input, and their trajectories during $\mathbf{\Omega}_2$ as the groundtruth.
We define $\mathbf{x}(|\Omega|, \max\Omega)=\mathbf{x}(\Omega)$ (ends at $t=\max\Omega$, with $|\Omega|$ steps) for better illustration.
The \emph{best-of-$K_g$} \cite{gupta2018social} $\ell_2$ loss function (short for preview loss) is used to supervise this training process:
\begin{equation}
    L_p  = 
        \min_{1 \leq s \leq K_g} \frac{1}{t_f|\mathcal{N}_i|} \sum_{j\in\mathcal{N}_i} \left\|
            \hatbf{X}^s_j(\mathbf{\Omega}_2) - \mathbf{X}_j(\mathbf{\Omega}_2)
        \right\|_2.
\end{equation}
The forecasted trajectories $\left\{\mathbf{X}_j(\hat{\Omega}|j\in\mathcal{N}_i)\right\}$ of all considered agents are then predicted by this short-term prediction network\footnote{See discussions of backbone choices of this short-term prediction network in \SECTION{sec_ablations}.}.
Correspondingly, the extended trajectory of any agent $i$ is denoted as:
\begin{equation}
    \mathbf{X}_i(\tilde{\Omega})= \mathrm{Concat}(\mathbf{X}_i(\Omega), \mathbf{X}_i(\hat{\Omega}))
\end{equation}

\NEWHALFLINE\noindent\textbf{Socialality Anchors $\tau_a$ and $\tau_b$.}
For all directly perceivable grouping-rule-modifiers, in-group social interactions are commonly characterized by both spatial distance and walking speed \cite{hall1973hidden,helbing1995social}, reflecting both the target agent $i$'s individual preference and the collective dynamics of its group.
This observation suggests that grouping rules are not governed by a single fixed threshold, but are modulated along two dimensions.
To enable the model to reflect such grouping rules observed in human social behavior, the proposed \NEWKERNEL~kernel introduces two learnable \ANCHOR~anchors, which serve as parametric containers for the grouping-rule modifiers.
Specifically, the anchors $\tau^a$ and $\tau^b$ respectively regulate the acceptable social distance and walking speed, thereby providing basis for adaptively modulating grouping rules in a human-inspired manner.
These \ANCHOR~anchors are also applied to calculate three modulation coefficients in the whole trajectory prediction model, which we will introduce later.

Given the target agent $i$'s trajectory during an extended grouping window $\tilde{\Omega}$, we use the network $m(\cdot)$ to compute its \emph{socialality} anchors:
\begin{equation}
    \left[\tau^a_i(\tilde{\Omega}), \tau^b_i(\tilde{\Omega})\right] = m \left(
        \mathbf{X}_i(\tilde{\Omega}) 
    \right) = \boldsymbol{\tau}_i(\tilde{\Omega})
    \label{eq_compute_socialality}
\end{equation}
See \TABLE{tab_both_symbol} for detailed structures of the embedding network $m(\cdot)$.

The anchor $\tau^a_i(\tilde{\Omega})$ is aimed to modulate the acceptable social distance range between agent $i$ and its group members.
Considering target agent $i$ and any neighbor $j$ during an extended grouping stage $\tilde{\Omega}$, the \NEWKERNEL~kernel assumes that the step-wise distance $d_i^t(j)=\left\|\mathbf{p}_j^{t}-\mathbf{p}_i^{t}\right\|_2, t \in \tilde{\Omega} $ between $j$ and $i$ should remain within $\tau^a_i(\tilde{\Omega})$-modulated agent $i$'s displacement $p_i(\tilde{\Omega})=\left\|\mathbf{p}_i^{\max \tilde{\Omega}}-\mathbf{p}_i^{\min \tilde{\Omega}}\right\|_2$.
Specifically, these two \emph{socialality anchors}'s values are constrained to $(-1, 1)$ by tanh activation, and thus we use $1+\tau^a_i(\tilde{\Omega})$ to set reasonable modulation range $(0,2)$.
Formally,
\begin{equation}
    \mathcal{S}^a_i(j, \tau_i^a| \tilde{\Omega}) = \mathbb{I}\left[
        \max_{t \in \tilde{\Omega}} \frac{d_i^t(j) }{p_i(\tilde{\Omega})}
        <
        1+\tau_i^a(\tilde{\Omega})
    \right].
    \label{dist_condition}
\end{equation}
Here, we use the maximum over $t$ to enforce a step-wise constraint: a neighbor is considered in-group only if it does not go beyond a certain distance range on each step.

Similar to $\tau^a_i(\tilde{\Omega})$, $\tau^b_i(\tilde{\Omega})$ directly modulates the acceptable walking speed difference between the target agent $i$ and its neighboring agent $j\in\mathcal{N}_i$.
Given an extended grouping stage $\tilde{\Omega}$, their walking speed difference is represented by the ratio $\rho_i(j|\tilde{\Omega})$ of their displacements: $\rho_i(j|\tilde{\mathbf{\Omega}}) = p_j(\tilde{\Omega}) / (p_i(\tilde{\Omega})+\epsilon)$, where $\epsilon$ is an infinitesimal quantity.
Formally, this $\tau^b_i(\tilde{\Omega})$-modulated condition is formulated as:
\begin{equation}
    \mathcal{S}^b_i(j, \tau_i^b | \tilde{\Omega})
        =
        \mathbb{I}\left[
        \left|
        \rho_i(j | \tilde{\Omega}) - 1
        \right|
        <
        \left|\tau_i^b(\tilde{\Omega})\right|
        \right].
\end{equation}
Here, $|\tau_i^b(\tilde{\Omega})|$ controls the tolerance of speed mismatch, where larger values indicate more permissive grouping.

Accordingly, the \NEWKERNEL~kernel is finally defined as
\begin{equation}
    \mathcal{K}_i(j, {\boldsymbol{\tau}_i}|\tilde{\Omega})
    =  \mathcal{S}^a_i(j, \tau_i^a| \tilde{\Omega})
    \cdot
    \mathcal{S}^b_i(j, \tau_i^b | \tilde{\Omega}) .
\end{equation}

In the proposed \NEWMODEL~Model, the subsequent concerns the extended time window $\tilde{\mathbf{\Omega}}$ for group priors.
Similarly, for our considered target agent $i$ and extended time window $\tilde{\mathbf{\Omega}}$, $i$'s group $\mathcal{G}_i$ and out-of-group agents' set $\overline{\mathcal{G}}_i$ are computed as
\begin{equation}
    \mathcal{G}_i =\{j\in\mathcal{N}_i|\mathcal{K}_i(j, {\boldsymbol{\tau}_i}|\tilde{\mathbf{\Omega}})=1\},
    \label{eq_group_assignment}
\end{equation}
\begin{equation}
    \overline{\mathcal{G}}_i =\{j\in\mathcal{N}_i|\mathcal{K}_i(j, {\boldsymbol{\tau}_i}|\tilde{\mathbf{\Omega}})=0\}.
\end{equation}
Accordingly, sets $\mathcal{G}_i$ and $\overline{\mathcal{G}}_i$ are visible to trajectory prediction models, acting as group priors for the further modeling of group-wise social interactions.

Compared to the long-term distance and the \NEWKERNEL~kernel, the improved \NEWKERNEL~kernel uses two \emph{socialality anchors} to adaptively learn target agent $i$'s grouping rules during the extended grouping window.
It enables the presentation of agents' diverse personalities and culturalities, thus modeling the \emph{agent-specific}, \emph{temporally coherent}, and \emph{context-adaptive} grouping rules.
These grouping rules yield more stable group priors around the current step, which in turn improves group-wise social interaction modeling in the subsequent perception mechanism.

\subsection{Perception Mechanism}
Both the original \MODEL~and the proposed \NEWMODEL~Models share a similar perception mechanism.
In the proposed \NEWMODEL~Model, the perception mechanism utilizes the grouping priors obtained from the \NEWKERNEL~kernel to model target agent $i$'s social interactions of both in-group and out-of-group agents.
To learn different interactions between in-group agents and out-of-group agents, we use two different strategies to imitate how humans perceive such group-wise social interactions.
Given the set of target agent $i$'s in-group agents $\mathcal{G}_i$ and their trajectories during the extended grouping stage $\tilde{\mathbf{\Omega}}$, the perception mechanism uses a self-trajectory encoder $l(\cdot)$ and a group-trajectory encoder $e(\cdot)$ to get $i$'s self representation and group representation, represented as
\begin{align}
    \mathbf{f}_i^e &= l \left(
        \mathbf{X}_i(\tilde{\mathbf{\Omega}})
    \right), \\
    \mathbf{f}_i^g &= e\left(
        \left\{
            \mathbf{X}_j(\tilde{\mathbf{\Omega}}) | j \in \mathcal{G}_i
        \right\}
    \right).
\end{align}
See \TABLE{tab_both_symbol} for detailed structure of encoder $e(\cdot)$.
This structurally ensures that the differences between how humans perceive in-group and out-of-group agents are learned from perception level intuitively.

The other strategy is applied on the out-of-group agents.
Inspired by human perception of surrounding agents, we design a strategy that aggregates sensory cues to form the target agent $i$'s out-of-group agents' representations.
In this manuscript, we focus on modeling social interactions between the target agent and its neighboring agents, and thus do not explicitly consider static obstacles or other environmental elements.
Among human sensory modalities, vision plays a central role in perceiving surrounding agents and planning future movements \cite{patla1991visual}.
Accordingly, pedestrians tend to be more responsive to neighboring agents that fall within their \emph{field of view} (FOV), i.e., the angular region of space visible to an agent $i$ given its current heading direction $\mathbf{d}_i$, as such neighboring agents provide more salient social cues~\cite{helbing1995social}.
The angle of FOV is denoted as $\theta_\mathrm{fov}$.

The target agent primarily perceives neighboring agents' social interactions with itself through observations within its field of view (FOV).
In contrast, agents outside the FOV provide only limited perceptual cues, but their presence can still be sensed through peripheral awareness to some extent.
It has been observed that humans are still aware of their neighboring agents even though they fall outside of the FOV, which is supported by sensitivity to luminance changes, and non-visual cues such as the sound of footsteps \cite{wolfe1989guided}. 

Here we define the unit vector $\mathbf{d}_i^{\mathbf{p}}$, which represents any direction of relative position $\mathbf{p}$ to the target agent $i$'s current position $\mathbf{p}_i^{0}$, as $\mathbf{d}_i^{\mathbf{p}}=\left(\mathbf{p} - \mathbf{p}_i^{0}\right)/\left(\left\| \mathbf{p} - \mathbf{p}_i^{0} \right\|_2\right),\mathbf{p} \ne \mathbf{p}_i^{0}$.
The target agent $i$'s heading direction $\mathbf{d}_i$ is denoted as its moving direction between last two adjacent time steps $t\in\{-1, 0\}$:
\begin{equation}
    \label{heading_dir}
    \mathbf{d}_i =\frac{\mathbf{p}_i^{0} - \mathbf{p}_i^{-1}}
{\left\| \mathbf{p}_i^{0} - \mathbf{p}_i^{-1} \right\|_2 + \epsilon}.
\end{equation}
Here, $\epsilon$ also indicates an infinitesimal quantity.
During the observation window $\mathbf{\Omega}_0=\{-t_o+1,\dots,-1,0\}$, the $\mathbf{FOV}_i$ of agent $i$ is then defined as the symmetric angular region centered around its heading direction $\mathbf{d}_i$: 
\begin{equation}
    \mathbf{FOV}_i
    =
    \left\{
    \mathbf{p}\in\mathbb{R}^2
    \;\middle|\;
    \left|\angle\left(\mathbf{d}_i^{\mathbf{p}}, \mathbf{d}_i\right)\right|
    \le
    \frac{\theta_{\mathrm{fov}}}{2}
\right\},
\end{equation} 
Here $\angle(\cdot)$ indicates signed angle, where positive and negative values correspond to counterclockwise (left) and clockwise (right) rotations, respectively.

For the target agent $i$ at current time step $t=0=\max \mathbf{\Omega}_0$, the set of its in-$\mathbf{FOV}$ neighboring agents is defined by
\begin{equation}
    \label{eq_fov}
\mathcal{F}_i =
\left\{j \mid
    j \in \overline{\mathcal{G}}_i ,
    \mathbf{p}_j^{\max \mathbf{\Omega}_0} \in \mathbf{FOV}_i
\right\}.
\end{equation}

Knowing only whether a neighboring agent lies within the target agent $i$'s $\mathbf{FOV}_i$ is insufficient to model fine-grained social interaction.
For instance, a neighboring agent may appear within the target agent's $\mathbf{FOV}_i$ and move toward it.
If they stick to their original heading direction, a collision may happen in this case.
However, they usually make an instant decision of adjusting their heading directions to avoid the incoming intersection, which has been a subconscious process for pedestrians \cite{cutting1995we}.
Inspired by the angle-based method proposed in SocialCircle \cite{wong2023socialcircle}, we regard such instant decision-making process needs a fast left/right choice, which corresponds to which side of $\mathbf{FOV}$ this incoming pedestrian falls in.
If it falls within left side of $\mathbf{FOV}$ (left-$\mathbf{FOV}$), the target agent tends to steer slightly to the right and vice versa.

Accordingly \EQUA{eq_fov}, we further split $\mathcal{F}_i$ into two subsets of neighbors according to left or right direction turning between $\mathbf{d}_i^{\mathbf{p}}$ and $\mathbf{d}_i$. 
As for the outside part of the $\mathbf{FOV}_i$, we mark it as the rear region.
Formally, we denote sets of the target agent $i$'s out-of-group agents positioned in the left-$\mathbf{FOV}_i$, right-$\mathbf{FOV}_i$ and rear region as
\begin{equation}
    \mathcal{C}_i^s \subseteq \overline{\mathcal{G}}_i, \;
    s \in \{\mathrm{right},\mathrm{left},\mathrm{rear}\}
\end{equation}
respectively. 

To model perceptual asymmetry of neighboring agents among these regions, which specifically indicates how the target agent perceives in-FOV, and out-of-FOV agents, we adopt an information selection strategy over neighboring agents' motion cues.
For all out-of-group agents $\overline{\mathcal{G}}_i$, we assume that the target agent $i$ mainly relies on three intuitive cues to perceive how they move in the scene in a SocialCircle-like way \cite{wong2023socialcircle}: their distance to $i$, their relative moving direction\footnote{Original SocialCircle adopts three cues, distance, direction, and speed.
Here, the perception mechanism tends to not consider the exact directions of out-group agents, thus replacing the direction cue with relative moving direction cue.} to $i$ and their walking speed.
For neighbors in the rear region, all motion cues except distance are set to zero, since agents outside the $\mathbf{FOV}_i$ (\IE, $\overline{\mathcal{G}}_i \setminus \mathcal{F}_i=\mathcal{C}_i^{\mathrm{rear}}$) provide much weaker directional and velocity information, while distance can still be perceived through peripheral awareness.
This aligns with the fact mentioned before that pedestrians tend to rely more on motion cues from agents within their $\mathbf{FOV}$, while paying less attention to agents outside them.

Here, we compute three region-level aggregated cues by averaging over neighboring agents in each region:
\begin{equation}
    \phi_q(\mathcal{C}_i^s)
    =
    \frac{1}{\left|\mathcal{C}_i^s\right|}
    \sum_{j \in \mathcal{C}_i^s} g^q_i(j), \;
    q \in \{\mathrm{dis}, \mathrm{dir}, \mathrm{vel}\},
    \label{eq_region_avg}
\end{equation}
where $g^q_i(j)$ provides an in-region measurement between the target agent $i$ and its each neighbor $j$ in the corresponding region:
\begin{align}
    g^{\mathrm{dis}}_i(j)
    &= \left\Vert \mathbf{p}_j^{0} - \mathbf{p}_i^{0} \right\Vert_2,
    \\
    g^{\mathrm{dir}}_i(j)
    &= \angle\left(\mathbf{d}_i, \mathbf{d}_j\right),
    \\
    g^{\mathrm{vel}}_i(j)
    &= \left\Vert \mathbf{p}_j^{0} - \mathbf{p}_j^{-t_o+1} \right\Vert_2.
\end{align}
According to \EQUA{heading_dir}, $\mathbf{d}_j$ also denotes agent $j$'s heading direction here.
We concatenate the motion cues from the right, left, and rear regions to form the perception vector $\mathbf{r}_i$.
An embedding layer $h(\cdot)$ then maps $\mathbf{r}_i$ into a high-dimensional feature:
\begin{equation}
    \mathbf{f}_i^{\overline{g}} = h\!\left(\mathbf{r}_i\right)
\in \mathbb{R}^{2d}.
\end{equation}
See \TABLE{tab_both_symbol} for detailed structure of embedding layer $h(\cdot)$.

In summary, the perception mechanism extracts a region-aware representation that encodes perceptual asymmetry in a human-inspired manner, providing structured social cues $\mathbf{f}_i^e$, $\mathbf{f}_i^g$ and $\mathbf{f}_i^{\overline{g}}$ for subsequent overall modeling.

\subsection{The \NEWMODEL~Trajectory Prediction Model}

Building upon the \NEWKERNEL~kernel and the perception mechanism introduced above, we now formulate the complete \NEWMODEL~trajectory prediction model.
The perception mechanism models target agent $i$'s group-wise interactions based on the group priors provided by \NEWKERNEL~kernel.
Similar to original \MODEL~Model in our conference paper \cite{zou2024who}, we fuse the social cues, \IE, $\mathbf{f}_i^g$, $\mathbf{f}_i^{\overline{g}}$, and the self representation $\mathbf{f}_i^e$ to get final feature $\mathbf{f}_i$ as the input to backbone prediction models.
In the \MODEL~Model, the three representations are simply concatenated, implicitly assuming equal contributions from each branch.
However, such an assumption may bias the optimization toward a dominant representation.
To address this issue, we introduce three learnable modulation coefficients $c_1$, $c_2$, and $c_3$, which adaptively rescale the representations prior before concatenation.
These modulation coefficients $c_1$, $c_2$, and $c_3$ are calculated from \emph{socialality anchors} $\tau^a_i(\tilde{\Omega})$ and $\tau^b_i(\tilde{\Omega})$.
Specifically, larger $\tau_i^a(\tilde{\Omega})$ relaxes distance tolerance, so we down-weight distance-sensitive group cues $\mathbf{f}_i^g$ accordingly, while larger $\tau_i^b(\tilde{\Omega})$ indicates more flexible speed choice, corresponding to more consideration of agent's self-cues $\mathbf{f}_i^e$.
Moreover, larger $\tau_i^a(\tilde{\Omega})$ and $\tau_i^b(\tilde{\Omega})$ both diminish the impact of out-group social cues $\mathbf{f}_i^{\overline{g}}$.
\begin{equation}
    c_1=1+\tau_i^b(\tilde{\Omega}), \; c_2=\left(1+\tau_i^a(\tilde{\Omega})\right)^{-1}, \; c_3=\frac{c_2}{c_1}.
\end{equation}
This structurally ensures that subsequent fusion stage is further intuitively regularized.

The fusion of social cues, \IE, $\mathbf{f}_i^g$, $\mathbf{f}_i^{\overline{g}}$, and $\mathbf{f}_i^e$, to form final representation $\mathbf{f}_i$ is formulated as
\begin{equation}
    \mathbf{f}_i = n \left(
        \mathrm{Concat\left(
            c_1\mathbf{f}_i^e ,
            c_2\mathbf{f}_i^g,
            c_3\mathbf{f}_i^{\overline{g}}
        \right)}
    \right).
\end{equation}
See \TABLE{tab_both_symbol} for detailed structure of encoder $n(\cdot)$.

We take Transformer~\cite{vaswani2017attention} and multi-style trajectory generation module in MSN~\cite{wong2021msn} as backbone trajectory prediction model.
The Transformer encoder captures high-dimensional representations of $\mathbf{f}_i$ by modeling temporal dependencies through self-attention over
$t_o$ observation steps.
Conditioned on the encoded features, the Transformer decoder attends to the encoder outputs while taking the linearly fitted trajectory $\hat{\mathbf{Y}}_i^{l}$ of the target agent as the value input.
The resulting feature embedding is then fed into the trajectory generation module~\cite{wong2021msn} to produce future trajectories.
Formally,
\begin{equation}
    \hat{\mathbf{Y}}_i^s = B_{\rm{prediction}} \left(
    \mathbf{f}_i, \hat{\mathbf{Y}}_i^l
    \right), 1 \leq s \leq K_f.
\end{equation}

\subsection{Training}
We adopt an end-to-end training strategy to train the whole \NEWMODEL~Model, including the short-term future prediction network mentioned in \SECTION{sec_new_kernel}.
Similar to the original \MODEL~Model, basic $\ell_2$ loss is also used, which is calculated as
\begin{equation}
        L_o = \min_{1 \leq s \leq K_f} \frac{1}{t_p} \left\|
            \hat{\mathbf{Y}}_i^s(t_p, t_p) - \mathbf{Y}_i(t_p, t_p)
        \right\|_2.
\end{equation}
Accordingly, the proposed \NEWMODEL~trajectory prediction model is trained with the weighted sum of the basic $\ell_2$ loss $L_o$ and the preview loss $L_p$:
\begin{equation}
    L = L_o + \beta L_p.
\end{equation}

\section{Experiments}
\label{sec_experiments}

\subsection{Experimental Settings}
\label{sec_settings}

\subsubsection{Datasets}
\NEWHALFLINE\noindent\textbf{ETH-UCY} \cite{pellegrini2009youll,lerner2007crowds} includes several videos captured in pedestrian walking scenes.
We follow the \emph{leave-one-out} \cite{alahi2016social} strategy to train and validate the proposed \NEWMODEL~Model with $\left\{t_o, t_p\right\}=\left\{8, 12\right\}$ and a sampling interval of $T = 0.4$s.
In the short-term future prediction subnetwork, we set $\left\{t_h = 4, t_f = 4\right\}$.

\NEWHALFLINE\noindent\textbf{Stanford Drone Dataset} (SDD) \cite{robicquet2016learning} comprises 60 video recordings obtained from an aerial perspective of the campus.
The agents, which fall into different categories (e.g., pedestrian, bicyclist, skateboarder, cart, car, and bus), have been labeled in pixels.
Following the same settings as previous researchers \cite{liang2020simaug}, the 60\% of videos were designated for training, the 20\% for validation, and the 20\% for testing.
We also set $\{t_o, t_p, T\} = \{8, 12, 0.4\mathrm{s}\}$ and $\{t_h, t_f\} = \{4, 4\}$.

\subsubsection{Metrics}
Following previous researchers, we also use the minimum Average/Final Displacement Error over $K$ generated trajectories \cite{alahi2016social,pellegrini2009youll,gupta2018social} to evaluate models' performance, known as ``minADE$_{K_f}$/minFDE$_{K_f}$'', where ${K_f}$ is usually set to 20 when evaluating.
For agent $i$, using subscript $_k$ to denote the $k$th prediction, we have
\begin{align}
    \mbox{minADE}_{20}(i) &= \min_{k \in \{1, 2, ..., K_f\}} \frac{1}{t_p} \sum_{t = 1}^{t_p} \left\|
        \hatbf{p}_i^{t_k} - \mathbf{p}_i^t
    \right\|_2, \\
    \mbox{minFDE}_{20}(i) &= \min_{k \in \{1, 2, ..., K_f\}} \left\|
        \hatbf{p}_i^{{t_p}_k} - \mathbf{p}_i^{t_p}
    \right\|_2.
\end{align}
The final metrics are averaged over every agent in each set, and we will refer to them as ``ADE/FDE'' for clarity.

\subsubsection{Implementation Details}
Both original \MODEL~and the proposed \NEWMODEL~Models are trained on one NVIDIA GeForce RTX 3090. 
Feature dimension of most subnetworks is set to $d = 32$.
The preview loss ratio is set to $\beta=0.4$.
The FOV angle of the perception mechanism is set to be $180^{\circ}$(discussion of choosing the FOV angle in our conference paper \cite{zou2024who}).
Following \cite{zhang2020social}, trajectories are preprocessed by moving to (0, 0).
We use the Adam optimizer with a learning rate 0.0002 to train our models, and the batch size is 1000 for up to 200 epochs.

\begin{table}[t]
\footnotesize
\caption{
    Comparisons to the state-of-the-art methods on ETH-UCY dataset
    }\label{tab_sota_comparisons}%
\begin{tabular}{@{}lllllll@{}}
\toprule
Models & eth  & hotel & univ & zara1 & zara2 & Avg.\\
\midrule

GP-Graph-STG\cite{bae2022learning}~(2022) & 0.48/0.77 & 0.24/0.40 & 0.29/0.47 & 0.24/0.42 & 0.23/0.40 & 0.29/0.49 \\
GP-Graph-PEC\cite{bae2022learning}~(2022) & 0.56/0.82 & 0.18/0.26 & 0.31/0.46 & 0.23/0.40 & 0.17/0.27 & 0.29/0.44 \\
GroupNet+CVAE\cite{xu2022groupnet}~(2022) & 0.46/0.73 & 0.15/0.25 & 0.26/0.49 & 0.21/0.39 & 0.17/0.33 & 0.25/0.44 \\
GroupNet+T++\cite{xu2022groupnet}~(2022) & 0.38/0.74 & 0.11/0.20 & \C{0.19}/\C{0.40} & \C{0.14}/0.32 & \C{0.11}/0.25 & 0.19/0.38 \\
\YNET\YNETCITE~(2021) & 0.28/\C{0.33} & \C{0.10}/\C{0.14} & 0.24/\C{0.41} & \C{0.17}/\C{0.27} &\C{0.13}/\C{0.22} & 0.18/\C{0.27} \\
\SEEM\SEEMCITE~(2023) & 0.62/1.20 & 0.61/1.21 & 0.50/1.04 & 0.31/0.61 & 0.36/0.68 & 0.48/0.95 \\
\EQMOTION\EQMOTIONCITE~(2023) & 0.40/0.61 & 0.12/0.18 & 0.23/0.43 & 0.18/0.32 & \C{0.13}/0.23 & 0.21/0.35 \\
\MSN\MSNCITE~(2023) & 0.27/0.41 & 0.11/0.17 & 0.28/0.48 & 0.22/0.36 & 0.18/0.29 & 0.21/0.34 \\
LMTraj-SUP~\cite{bae2024can}~(2024) & 0.65/1.04 & 0.26/0.46 & 0.57/1.16 & 0.51/1.01 & 0.38/0.74 & 0.48/0.88 \\
ET+HG\cite{kim2024higher}~(2024) & 0.33/0.56 & 0.13/0.21 & \C{0.23}/0.47 & 0.19/0.33 & 0.15/0.25 & 0.21/0.36 \\
\LGTRAJ\LGTRAJCITE~(2024) & 0.38/0.56 & 0.11/0.17 & \C{0.23}/0.42 & 0.18/0.33 & 0.14/0.25 & 0.20/0.34 \\
\SCMODEL\SCCITE~(2024) & 0.25/0.38 & 0.12/\C{0.14} & \C{0.23}/0.42 & 0.18/0.29 & \C{0.13}/\C{0.22} & 0.18/\C{0.29} \\
\SCPMODEL\SCPCITE~(2024)            & 0.25/0.39     & \C{0.10}/\C{0.15}     & 0.24/0.42         & 0.18/\C{0.28}     & \C{0.13}/\C{0.22}     & 0.18/\C{0.29} \\
\UPDD\UPDDCITE~(2024) & \C{0.22}/0.42 & 0.17/0.30 & \C{0.14}/0.28 & \C{0.16}/0.30 & 0.14/0.31 & \C{0.17}/0.32 \\
\EVMODEL\EVCITE~(2025)              & 0.25/0.38     & 0.11/0.16     & \C{0.23}/0.42     & 0.19/0.30         & \C{0.13}/0.24     & 0.18/0.30 \\
\REMODEL\RECITE~(2025)              & \C{0.23}/\C{0.35}     & \C{0.10}/\C{0.15}     & 0.24/\C{0.41}     & \C{0.17}/0.29     & \C{0.13}/\C{0.22}     & \C{0.17}/\C{0.28} \\ 

\midrule
\MODEL~\textbf{(Ours)}  & 0.25/0.38 &\C{0.10}/\C{0.15}&0.25/0.44&\C{0.17}/\C{0.28}&\C{0.13}/\C{0.22}&  0.18/\C{0.29}  \\
\NEWMODEL~\textbf{(Ours)} & \C{0.23}/\C{0.37} &\C{0.10}/0.16&0.24/0.43&\C{0.17}/0.29&\C{0.13}/\C{0.22}& \C{0.17}/\C{0.29}   \\
\botrule

\end{tabular}
\footnotetext{
    Metrics are reported in the form of ``ADE/FDE'' (\emph{best-of-20}) in meters.
    Lower metrics indicate better prediction performance.
    \C{Blue markers} denote the best 3 results on each set.
}
\end{table}

\begin{table}[t]
\caption{
    Comparisons to the state-of-the-art methods on SDD dataset
    }\label{tab_sdd_comparisons}%
\begin{tabular}{@{}ll@{\hspace{2em}}ll@{}}
\toprule
Models & ADE/FDE & Models & ADE/FDE \\
\midrule

\SPECTGNN\SPECTGNNCITE~(2021) & 8.21/12.41 & \IMP\IMPCITE~(2023) & 8.98/15.54 \\
\YNET\YNETCITE~(2021) & 7.85/11.85 & LMTraj-SUP~\cite{bae2024can}~(2024) & 17.5/34.5 \\
GP-Graph-STGCNN\cite{bae2022learning}~(2022) & 10.6/20.5 & \RAN\RANCITE~(2024) & 10.97/19.95 \\
GP-Graph-PECNet\cite{bae2022learning}~(2022) & 9.1/13.8 & \LGTRAJ\LGTRAJCITE~(2024) & 7.80/12.79 \\
GroupNet+PECNet\cite{xu2022groupnet}~(2022) & 9.65/15.34 & \UPDD\UPDDCITE~(2024) & 6.59/13.90 \\
GroupNet+CVAE\cite{xu2022groupnet}~(2022) & 9.31/16.11 & \SCMODEL\SCCITE~(2024) & 6.54/10.36 \\
\NSPSFM\NSPSFMCITE~(2022) & 6.52/10.61 & \SCPMODEL\SCPCITE~(2024) & 6.44/10.22 \\
\MUSEVAE\MUSEVAECITE~(2022) & \C{6.36}/11.10 & \EVMODEL\EVCITE~(2025) & 6.57/10.49 \\
\FLOWCHAIN\FLOWCHAINCITE~(2023) & 9.93/17.17 & \REMODEL\RECITE~(2025) & \C{6.27}/\C{10.02} \\

\midrule
\MODEL~\textbf{(Ours)}  &   6.39/\C{10.17} & \NEWMODEL~\textbf{(Ours)} &  \C{6.28}/\C{10.12} \\
\botrule

\end{tabular}
\footnotetext{
    Metrics are reported in the form of ``ADE/FDE'' (\emph{best-of-20}) in pixels rather than meters compared to ETH-UCY.
    Lower metrics indicate better prediction performance.
    \C{Blue markers} denote the best 3 results on each set.
}
\end{table}

\subsection{Comparisons to State-of-the-Art Methods}
\label{sec_comparisons}
In this section, we compare \NEWMODEL~Model with recent state-of-the-art trajectory forecasting methods on the ETH-UCY and SDD benchmarks. 
Results are summarized in \TABLE{tab_sota_comparisons} and \TABLE{tab_sdd_comparisons}.

On ETH-UCY, \NEWMODEL~Model consistently outperforms our previous GPCC model across most scenes. 
In GPCC, grouping is determined by a static long-term distance threshold, which cannot adapt to different social configurations. 
By introducing the agent-specific \NEWKERNEL~kernel, \NEWMODEL~Model produces more stable grouping decisions, leading to lower ADE and FDE. 
The improvement is more evident in crowded scenes such as univ and eth, where group formations frequently change and agents exhibit diverse motion patterns. 
Compared with representative graph-based group-aware methods such as GP-Graph-STGCNN (0.29/0.49) and GP-Graph-PECNet (0.29/0.44), \NEWMODEL~Model reduces the average ADE by approximately 41\%.  
It also outperforms recent interaction-based models such as LGTRAJ (0.20/0.34) and EigenTrajectory+HighGraph (0.21/0.36).

SDD contains larger spatial layouts and more diverse motion scales.
\NEWMODEL~Model's performance reaches 6.28/10.12, outperforming the original \MODEL (6.39/10.17).  
Compared with earlier graph-based approaches, the improvement is more substantial.  
For instance, relative to GroupNet+PECNet (9.65/15.34) and GP-Graph-STGCNN (10.6/20.5), the ADE reduction exceeds 34\% and 40\%, respectively.  
Even against strong recent models such as \SCMODEL~(6.54/10.36) and NSPSFM (6.52/10.61), \NEWMODEL~Model achieves lower prediction errors.
Since SDD scenes often include agents with different motion intentions and spatial tolerances, a fixed threshold-based grouping strategy becomes less reliable. 
The proposed anchor-based kernel adjusts grouping rules according to each agent's motion pattern, resulting in more accurate interaction modeling.

Notably, the proposed \NEWMODEL~Model's backbone architecture remains identical to the original \MODEL.  
The main structural modification lies in replacing the static long-term distance kernel with the proposed \NEWKERNEL~kernel.  
The consistent improvements on both datasets indicate that agent-specific, temporally coherent, and context-adaptive grouping rules leads to better trajectory forecasting performance.

\subsection{Ablation Analyses}
\label{sec_ablations}

In this section, we provide quantitative evaluations to further validate the proposed \NEWMODEL~Model.  
First, we conduct separate ablations to verify whether grouping priors actually contribute to the prediction model.
Then we conduct several further ablations to validate the contribution of each key component that modifies the \NEWKERNEL~kernel's grouping rules, including the \ANCHOR~anchors and the extended grouping window.
Detailed ablation settings and results are reported in \TABLE{tab_grouping_ablations} and \TABLE{tab_rules_ablations}.

\begin{table}[t]
\footnotesize
\caption{
    Ablation studies on validating the grouping priors on ETH-UCY and SDD datasets
    }\label{tab_grouping_ablations}%
\setlength{\tabcolsep}{3pt}
\begin{tabular}{@{}l|l|lllll|l@{}}
\toprule
ID & Grouping Kernel & eth & hotel& univ& zara1& zara2& SDD \\
\midrule
a1&-&\R{0.510}/\R{1.052}&\R{0.117}/\R{0.163}&\R{0.304}/\R{0.575}&\R{0.188}/\R{0.302}&\R{0.147}/\R{0.231}&\R{6.462}/\R{10.379}\\
a2&Long-Term Distance& 0.254/0.383 &0.103/\C{0.156}&0.256/0.448&0.175/\C{0.284}&0.134/0.225&6.394/10.173\\

\midrule
a0 & \NEWKERNEL &\C{0.233}/\C{0.369}&\C{0.102}/\C{0.156}&\C{0.240}/\C{0.428}&\C{0.168}/0.294&\C{0.130}/\C{0.223}& \C{6.283}/\C{10.121}\\
\botrule

\end{tabular}
\footnotetext{
    ``-'' denotes that no grouping kernel is enabled in the corresponding variation model.
    The variation that uses the Long-Term Distance Kernel is the same as our original \MODEL~Model.
    \C{Blue markers} denote the best results on each set.
    \R{Red markers} denote the worst results on each set.
}
\end{table}

\NEWHALFLINE\noindent\textbf{Grouping Kernels.}
The main difference among the three variations in \TABLE{tab_grouping_ablations} is that whether the groups have been formed or how the groups are formed before forecasting.
Variation a1 serves as the baseline for comparisons, with no grouping priors introduced and no group-aware social interactions considered.
In contrast, grouping kernels are introduced in both variations a2 and a0 to provide direct grouping priors for the subsequent interaction modeling.
The difference between a2 and a0 is that a2 uses the original long-term distance kernel, while a0 adopts our proposed \NEWKERNEL~kernel.

Based on such setup and the results in \TABLE{tab_grouping_ablations}, we find that both variations a2 and a0 achieve 50.2\%/63.6\% significantly better ADE/FDE on eth, compared with the base variation a1.
Compared with the \emph{unbounded} variation a1, a0 finally achieves up to 40\% ADE gains across these datasets, though these variations share the same backbone prediction model with exactly the same network capacity. 
This means that the explicit modeling of social interactions with specific grouping priors has brought considerable performance improvements in the \emph{group-bounded} variations a0 and a2 over the \emph{unbounded} a1, regardless of which grouping kernel has been used.

Moreover, the full \NEWMODEL~Model (a0) further obtains up to 4.1\% better ADE on eth and 4.5\% FDE on univ than variation a2 (\MODEL, which uses our original long-term distance kernel).
Here, the most significant difference between such kernels is that the recommendations of group members rely on multiple adaptively collaborated anchors in the proposed \NEWKERNEL~kernel, rather than a single fixed threshold in the long-term distance kernel.
We can infer from such performance differences that the real world groupings are more likely to be determined dynamically and collaboratively.
See our further discussions in \SECTION{sec_discussions_groupings}.
These results indicate that stable performance gains could be obtained through such explicit grouping priors, while the proposed \NEWKERNEL~kernel still has the ability to further expand network capability by refining our original grouping rules in \MODEL, especially in scenarios with more complex interactions and higher uncertainty like univ and eth.

\begin{table}[t]
\footnotesize
\caption{
    Ablation studies on validating the proposed grouping rules on ETH-UCY and SDD datasets
    }\label{tab_rules_ablations}%
\setlength{\tabcolsep}{4pt}
\begin{tabular}{@{}l|c|c|lllll|l@{}}
\toprule
ID & $\tau^a$ $\tau^b$ &R P C& eth & hotel& univ& zara1& zara2& SDD \\
\midrule
a1&-~\checkmark&\checkmark \checkmark \checkmark&\R{0.252}/\R{0.398}&0.115/0.180&\R{0.251}/0.451&\R{0.185}/0.317&\R{0.141}/\R{0.241}&\R{6.431}/\R{10.452}\\
a2&\checkmark~-&\checkmark \checkmark \checkmark&0.240/0.384&0.110/0.175&0.244/0.440&0.183/0.313&0.139/0.235&6.343/10.278\\
a3 &\checkmark~0&\checkmark \checkmark \checkmark&0.232/0.364&0.104/0.158&0.241/0.434&0.180/0.305&0.134/0.225&6.305/10.189\\
a4 &0~\checkmark &\checkmark \checkmark \checkmark&0.237/0.375&0.107/0.169&0.243/0.435&0.180/0.304&0.137/0.233&6.331/10.218\\
a5 &0~0&\checkmark \checkmark \checkmark&0.234/0.370&0.106/0.166&0.242/0.434&0.180/0.308&0.132/\C{0.222}&6.313/10.262\\

\midrule
a6 &\checkmark~\checkmark&- - \checkmark&0.238/0.378&\R{0.116}/\R{0.187}&0.248/\R{0.460}&0.184/\R{0.325}&0.137/0.238&6.356/10.315\\
a7 &\checkmark~\checkmark&\checkmark - \checkmark&\C{0.231}/\C{0.363}&0.113/0.179&0.247/0.441&0.183/0.317&0.135/0.230&\C{6.260}/\C{10.020}\\ 
a8 &\checkmark~\checkmark&- \checkmark -&0.234/0.367&0.110/0.168&0.244/0.436&0.182/0.308&0.135/0.229&6.303/10.194\\

\midrule
a0 & \checkmark~\checkmark&\checkmark \checkmark \checkmark&0.233/0.369&\C{0.102}/\C{0.156}&\C{0.240}/\C{0.428}&\C{0.168}/\C{0.294}&\C{0.130}/0.223&6.283/10.121\\
\botrule

\end{tabular}
\footnotetext{
    $\tau^a$ and $\tau^b$ denote whether the corresponding \ANCHOR~anchor is activated during training.
    ``R'', ``P'', and ``C'' denote whether reviews, previews, or current time are included in the grouping window respectively.
    ``$0$'' denotes the corresponding \ANCHOR~anchor is set to $0$ during training.
    ``-'' denotes the corresponding \ANCHOR~anchor is completely removed.
    Both anchors are activated in the proposed full \NEWMODEL~Model.
    \C{Blue markers} denote the best results on each set.
    \R{Red markers} denote the worst results on each set.
}
\end{table}

\NEWHALFLINE\noindent\textbf{\emph{Socialality}~Anchors.}
We next validate the grouping rules in the proposed \NEWMODEL in \TABLE{tab_rules_ablations}.
The grouping rules are mainly shaped by two components, i.e., the proposed \ANCHOR~anchors and the extended grouping window.
The \ANCHOR~anchors are designed to encode context-adaptive and agent-specific grouping preferences.
In particular, they modulate the acceptable social distance and speed flexibility for grouping, \emph{directing} different agents to follow different grouping rules.
The extended grouping window is designed to provide temporally extended grouping cues around the current time step.
Rather than making grouping decisions from a single instantaneous moment or only observed period, the extended grouping window allows the \NEWMODEL~to incorporate grouping evidence from both reviews (observed window) and previews (short-term forecasted window).

\emph{(i) Overall Verification:}
Among all variations, the full \NEWMODEL~(a0) achieves the best or near-best performance on almost all subsets.
This result indicates that the complete grouping rules, jointly shaped by the two \ANCHOR~anchors and the extended grouping window, are the most effective for trajectory forecasting.
For example, on hotel, the full model enhances ADE/FDE from 0.115/0.180 in a1 to 0.102/0.156, corresponding to improvements of 11.3\% and 13.3\%, respectively.
On univ, compared with the current-only window variant a6, the full model further reduces ADE/FDE from 0.248/0.460 to 0.240/0.428, yielding improvements of 3.2\% and 7.0\%.
These observations suggest that both the agent-specific anchor design and the temporally extended grouping cues are important to the final prediction performance.

\emph{(ii) Anchor Verification:}
We evaluate the contribution of the proposed \ANCHOR~anchors from two aspects, including the whether each anchor exists and the respective roles of distance anchor $\tau^a$ and speed anchor $\tau^b$.

We first compare the variations where an anchor is set to $0$ with those where the corresponding anchor is completely removed from the whole model.
These two degradations are distinctively different.
When an anchor is removed, it is discarded from the entire model pipeline.
Accordingly, it is not involved in learning agent-specific grouping preferences in the grouping stage, nor can it contribute to the subsequent modulation of leveled interactions.
In contrast, the anchor set to $0$ loses its capability to learn specific grouping preferences and applies the same fixed anchor value to all neighboring agents instead.
However, its associated constraint and modulation effect are still partially preserved in the model.
In summary, setting an anchor to $0$ mainly weakens the learning of agent-specific grouping rules, while removing it disables the corresponding constraint completely.

When both anchors are set to $0$ (a5), the model still keeps competitive performance on most scenes, such as 0.234/0.370 on eth and 0.242/0.434 on univ, which are only slightly worse than the full \NEWMODEL~model by 0.4\%/0.3\% and 0.8\%/1.4\% in ADE/FDE.
In contrast, when one anchor is completely removed, the performance drops more clearly than merely setting it to $0$.
This difference can be seen from paired comparisons where only one variable is changed at a time.
When $\tau^a$ is removed while $\tau^b$ is retained, a1 performs worse than a4, where $\tau^a$ is instead fixed to $0$.
For example, on eth, the ADE/FDE gets worse from 0.237/0.375 in a4 to 0.252/0.398 in a1, corresponding to degradations of 6.3\% and 6.1\%.
Similarly, on hotel, the performance drops with degradations of 7.5\% and 6.5\% in a1.
A similar trend can also be observed for $\tau^b$ by comparing a2 and a3.
When $\tau^b$ is removed, a2 is consistently worse than a3, where $\tau^b$ is fixed to $0$.
These observations suggest that retaining an anchor in the model with a fixed value is different from removing it completely, and the two degradations lead to systematically different performance.

We further verify the respective roles of $\tau^a$ and $\tau^b$.
By comparing a3 and a4, we can observe that a3 is also consistently better than enabling only $\tau^b$ while setting $\tau^a$ to $0$ (a4) on several subsets.
For example, on hotel, a3 improves over a4 from 0.107/0.169 to 0.104/0.158, corresponding to relative gains of 2.8\% and 6.5\% in ADE/FDE.
On zara2, a3 further improves over a4 from 0.137/0.233 to 0.134/0.225, yielding gains of 2.2\% and 3.4\%, respectively.
A similar trend can also be observed when one anchor is completely removed.
For example, on hotel, a2 improves over a1 with the gains of 4.3\% and 2.8\%.
On univ, the corresponding gains are 2.8\% and 2.4\%, with the performance improving from 0.251/0.451 to 0.244/0.440.
On zara2, a2 further improves over a1 1.4\% and 2.5\%, on ADE/FDE respectively.
We can infer from these observations that $\tau^a$ plays a more basic role in grouping, while $\tau^b$ mainly acts as a complementary refinement.

\emph{(iii) Grouping Window Verification:}
We further validate the role of the extended grouping window.
Compared with the current-only variant a6, introducing reviews or previews consistently improves the prediction performance.
When reviews are added, a7 improves over a6 2.6\% and 4.3\% in ADE/FDE on hotel.
On univ, the corresponding gains are 0.4\% and 4.1\%, and on zara1 they are 0.5\% and 2.5\%.
Similarly, when previews are introduced, a8 improves over a6 on hotel, univ, and zara1 by 5.2\%/10.2\%, 1.6\%/5.2\%, and 1.1\%/5.2\% in ADE/FDE, respectively.

Moreover, using only previews (a8) even performs slightly better than using reviews with the current time step (a7) on most subsets, including hotel (2.7\%/6.1\%), univ (1.2\%/1.1\%), zara1 (0.5\%/2.8\%), and zara2 (0.0\%/0.4\%) in ADE/FDE, respectively.
This observation suggests that grouping evidence may be direction-sensitive.
In many scenes, group relations may be more clearly reflected by whether agents are likely to continue moving coherently in the near future, rather than only by whether they remained close in the observed past.

When reviews, previews, and the current time step are all included together, the full model a0 achieves the best overall performance among all grouping-window variants.
Compared with a7, the full model's ADE/FDE further improves 9.7\% and 12.8\% on hotel.
Compared with a8, the corresponding gains on zara1 are 7.7\% and 4.5\%.
Overall, these results suggest that the extended grouping window improves the prediction performance.


\NEWHALFLINE\noindent\textbf{Additional Verifications.}
We additionally evaluate the design of the short-term preview network.  
Results are summarized in \TABLE{tab_beta_ablations} and \TABLE{tab_backbone_ablations}.

\begin{table}[t]
\caption{
    Ablation studies on validating the preview loss ratio $\beta$ on ETH-UCY and SDD datasets
    }\label{tab_beta_ablations}%
\begin{tabular}{@{}llllllll@{}}
\toprule
ID   &  $\beta$& eth & hotel& univ& zara1& zara2& SDD \\

\midrule
a1  & 0&0.235/0.374&0.111/0.176&0.248/0.440&0.183/0.311&0.135/0.230&6.419/10.421\\
a2  & 0.2&0.236/0.377&0.109/0.168&\C{0.239}/\C{0.424}&0.177/0.295&0.137/0.231&6.274/10.148\\
a3  & 0.6&\C{0.232}/\C{0.362}&0.108/0.166&0.242/0.432&0.185/0.318&0.136/0.232&6.296/10.155\\
a4  & 0.8&0.234/0.365&0.104/0.157&0.242/0.431&0.176/0.302&0.132/0.233&6.344/10.241\\
a5  & 1.0&0.235/0.365&0.110/0.171&0.245/0.437&0.180/0.305&0.138/0.234&6.369/10.340\\
\midrule
a0  & 0.4&0.233/0.369&\C{0.102}/\C{0.156}&0.240/0.428&\C{0.168}/\C{0.294}&\C{0.130}/\C{0.223}&\C{6.283}/\C{10.121}\\
\botrule

\end{tabular}
\footnotetext{
    The preview loss ratio $\beta$ is set with different values from $0$ to $1.0$.
    The default setting is $\beta = 0.4$.
    \C{Blue markers} denote the best results on each set.
}
\end{table}

\emph{(i) Preview Loss Ratio Verification:}
We first analyze the effect of the preview loss ratio $\beta$.
Overall, different values of $\beta$ lead to relatively moderate performance variations across scenes, while a moderate setting gives the best overall balance.
When $\beta=0$ (a1), i.e., without preview supervision, the performance degrades on several subsets.
For example, on univ, the ADE/FDE is worse than a0 of 3.3\%/2.8\%.
When $\beta$ is set to 0.2 (a2), the performance on univ becomes the best among all settings.
When $\beta$ is increased to 0.6 (a3), eth reaches the best result of 0.232/0.362, but zara1 worsens 10.1\% and 8.2\% than a0.
Further increasing $\beta$ to 1.0 (a5) leads to another drop on multiple subsets, such as hotel, where the ADE/FDE worsens 7.8\%/9.6\%.
We can infer from the above observations that preview supervision is beneficial, but its weight should remain within a moderate range.

\begin{table}[t]
\footnotesize
\caption{
    Ablation studies on validating the short-term prediction networks on ETH-UCY and SDD datasets
    }\label{tab_backbone_ablations}%
\setlength{\tabcolsep}{5pt}
\begin{tabular}{@{}llllllll@{}}
\toprule
ID  & Backbone &   eth & hotel& univ& zara1& zara2&   SDD \\
\midrule
a1 &Linear&0.236/0.372&0.111/0.172&0.242/0.435&0.182/0.307&0.137/0.233&6.375/10.194\\ 
a2 &MLP&0.237/0.373&0.110/0.183&0.245/0.437&0.183/0.312&0.137/0.231&6.438/10.422\\

\midrule
a0 & Tran & \C{0.233}/\C{0.369}&\C{0.102}/\C{0.156}&\C{0.240}/\C{0.428}&\C{0.168}/\C{0.294}&\C{0.130}/\C{0.223}& \C{6.283}/\C{10.121}\\
\botrule

\end{tabular}
\footnotetext{
    Backbone indicates different backbones used in short-term prediction networks.
    ``Linear'' means simple linear fit is applied to generate short-term previews.
    ``MLP'' is used in another variation.
    The default setting in the full \NEWMODEL~Model uses ``Tran (Transformer)'' as the prediction backbone.
    \C{Blue markers} denote the best results on each set.
}
\end{table}

\emph{(ii) Backbone Verification:}
We then validate the role of the short-term prediction network with different prediction backbones.
Although the default \NEWMODEL~uses a Transformer predictor, replacing it with simpler alternatives still yields reasonably competitive performance across scenes.
For example, using a Linear predictor (a1) gives 0.236/0.372 on eth, which is only 1.3\% and 0.8\% worse than the full model, while using an MLP predictor (a2) leads to similarly small degradations of 1.7\% and 1.1\%.
It should be noted that the main purpose of this short-term prediction network is not to rely on a specific predictor architecture.
Instead, the key point is to introduce a short-term prediction network itself, so that the model can obtain preview cues for grouping.

\subsection{Discussions}
\label{sec_discussions}
In this section, we further analyze and discuss the proposed \NEWMODEL~Model through multiple validations.
As defined in \SECTION{sec_method}, grouping kernels in the proposed trajectory prediction models determine whether each neighbor belongs to a target agent's group.
Specifically, for the same target agent, we focus on evaluating different grouping decisions of the proposed \NEWKERNEL~kernel compared with the original long-term distance kernel.
Moreover, the core of the \NEWKERNEL~kernel is a pair of grouping anchors, \IE, \ANCHOR~anchors, modulating acceptable social distances and walking speeds within each target agent's group.
We visualize and analyze how such agent-specific \ANCHOR~anchors distribute in a 2D space at different scenes.

\subsubsection{Overall Model Predictions}

\begin{figure}[!htbp]
\centering
\includegraphics[width=1.0\textwidth]{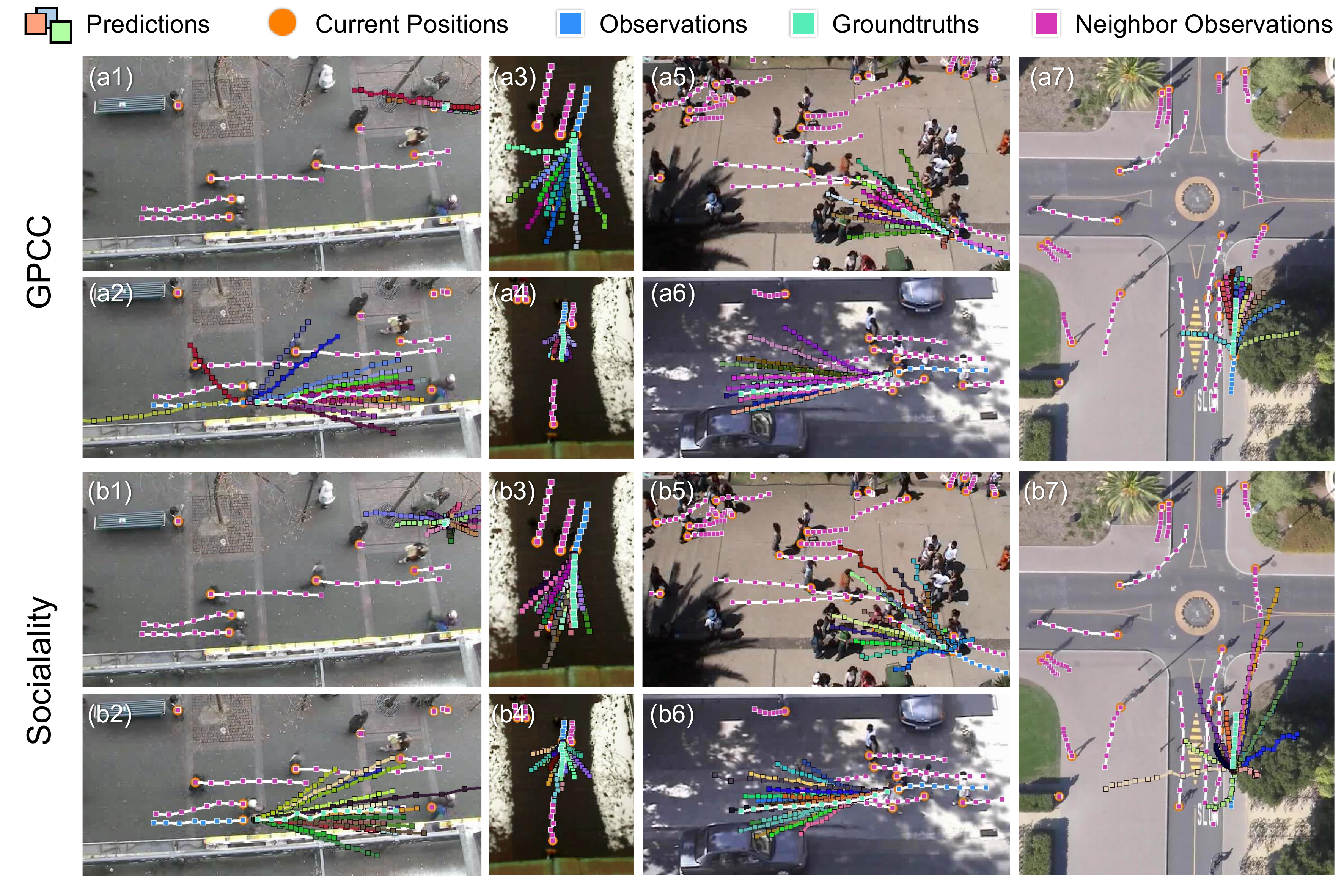}
\caption{Visualized predictions of the original \MODEL~Model and the proposed \NEWMODEL~Model in different scenes.}\label{fig_overall_visualization}
\end{figure}

Here, we visualize trajectories predicted by the proposed \NEWMODEL~Model and the original \MODEL~Model under representative scenes from ETH-UCY and SDD, as shown in \FIG{fig_overall_visualization}. 
Overall, we could observe that the \NEWMODEL~Model produces more behaviorally plausible and scene-consistent future trajectories than \MODEL~Model. 
In (a1) and (b1), the original \MODEL~Model tends to generate less stable predictions for agents with weak motion cues, \EG, mostly generating sudden movements for nearly static agents. 
By contrast, the proposed \NEWMODEL~Model yields more comprehensive yet still reasonable predictions, better reflecting the fact that such agents are more likely to remain locally stable instead of moving abruptly. 
In the SDD cases shown in (a7) and (b7), \NEWMODEL~Model produces noticeably more multi-style predictions than the original \MODEL~Model. 
Such diversity better matches the highly heterogeneous motion patterns in SDD, where agents are often influenced by open spaces, loose interactions, and multiple navigable choices. 
In summary, the proposed model not only improves the rationality of prediction outcomes, but also enhances the expressiveness of trajectory modes under different social environments.

\subsubsection{Discussions of Groupings}
\label{sec_discussions_groupings}

\begin{figure}[!htbp]
\centering
\includegraphics[width=1.0\textwidth]{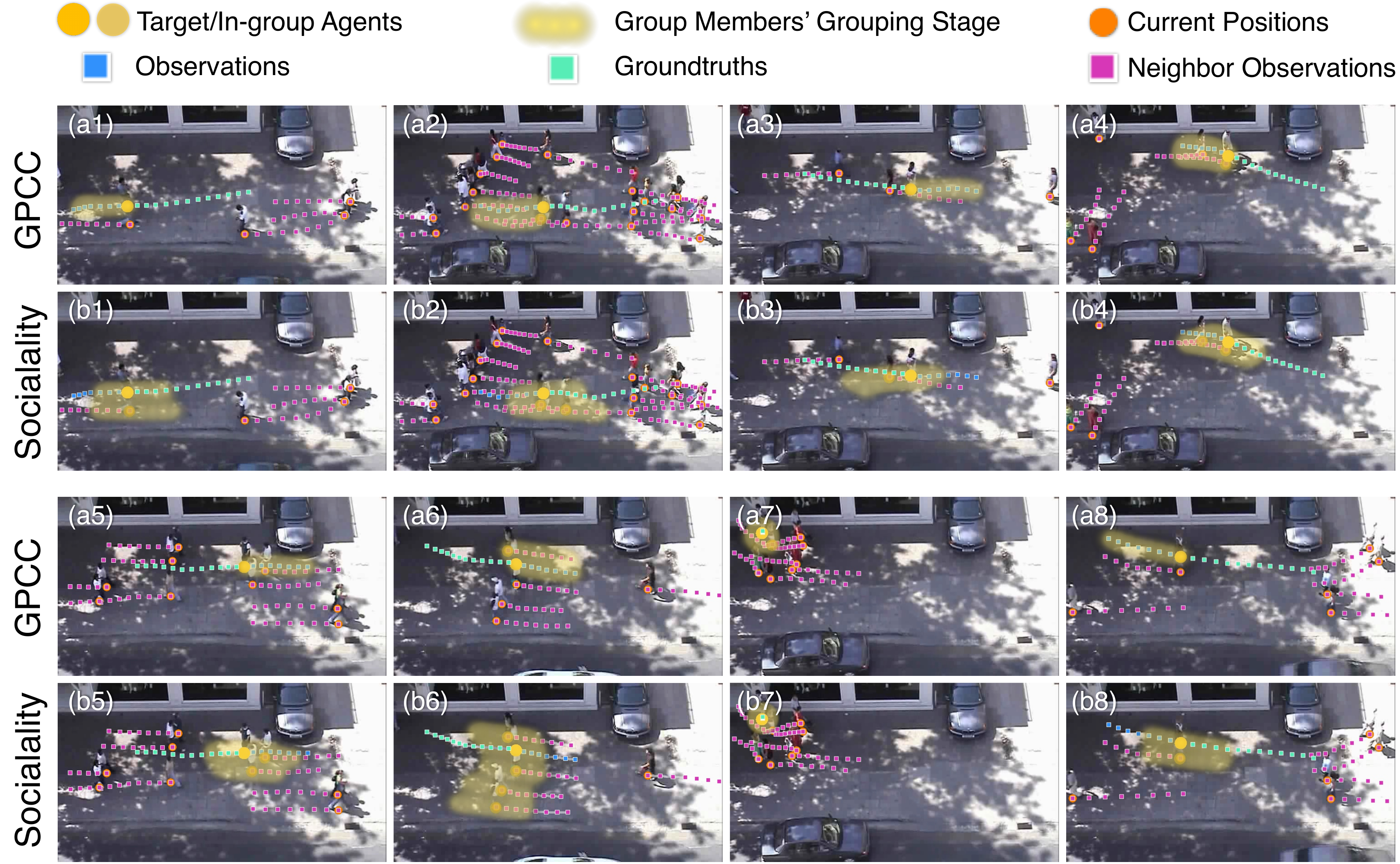}
\caption{
    Visualized grouping results of the original \MODEL~Model and the proposed \NEWMODEL~Model in different scenes.
    \MODEL~uses the original observation-only grouping window, whereas \NEWMODEL~uses the proposed extended grouping window.
    }\label{fig_grouping_validations}
\end{figure}

\NEWHALFLINE\noindent\textbf{Overall Validations.}
To qualitatively evaluate the evolving grouping capability of the proposed \NEWKERNEL~kernel, we visualize how \NEWKERNEL~kernel handles cases where the long-term distance kernel fails to process in \FIG{fig_grouping_validations}.
Here, we treat agents walking together, having social interactions (talking to each other, eye contact, and so on), and share similar destinations as likely group members.

It can be seen from the above figures that grouping decisions vary substantially across different prediction models even for the same target agent.
By comparing (a1), (a2), (a3) and (b1), (b2), (b3) in \FIG{fig_grouping_validations}, we can observe that one neighboring agent is considered to be within the group by \NEWMODEL~Model, while the original \MODEL~Model regards them as out-of-group agents due to the static-threshold-controlled grouping kernel.
The static nature of this threshold determines that the choice of this threshold can only be a trade-off\footnote{The threshold value is discussed in our conference paper \cite{zou2024who}.}.
On the one hand, we could set a relatively larger threshold value to let the long-term distance kernel cover more neighboring agents when grouping, but large threshold also means including more agents that are not the target agent's group members.
On the other hand, a relatively smaller threshold could be applied to strictly constrain the group ``membership'', thus overlooking in-group agents that are slightly away from the target agents.
With the newly proposed \NEWKERNEL~kernel, we could observe that a precise group assignment is conducted even in crowded scenes like (b2) and (b5).

Previous studies have shown that relatively large groups (with more than three members) tend to move in a ``V''-shaped formation to ensure in-group social interactions \cite{moussaid2010walking}.
Here, (a6) and (b6) represent a large-group scene, where we could observe five group members (including the target agent itself).
In (a5), the long-term distance kernel simply assigns the nearest neighbor as a group member and fails to recognize the rest of the group members.
In such large-group cases, the original long-term distance kernel could only conduct incomplete group assignments based on the distance threshold alone, which neglects the phenomenon that the acceptable social distance between any group member and the target agent varies a lot due to ``V''-shaped formation.
Notably, every in-group agent is included by the \NEWKERNEL~kernel, as shown in (b6).

The capability of including more agents for larger groups does not necessarily mean the \NEWKERNEL~kernel tends to include an out-of-group agent more easily.
In (a7) and (b7), the target agent moves in a small range and almost stays still in front of the zara store.
It seems that this target agent is waiting for someone by itself.
In (a7), the long-term distance kernel justifies a neighboring agent walking out of the store as a group member of the target agent.
However, the \NEWKERNEL~kernel succeeds in excluding this closely passing-by agent, as shown in (b7).
Visualizations in \FIG{fig_grouping_validations} demonstrate the \NEWKERNEL~kernel's effectiveness of capturing the agent-specific and context-adaptive grouping rules.

\begin{figure}[!htbp]
\centering
\includegraphics[width=1.0\textwidth]{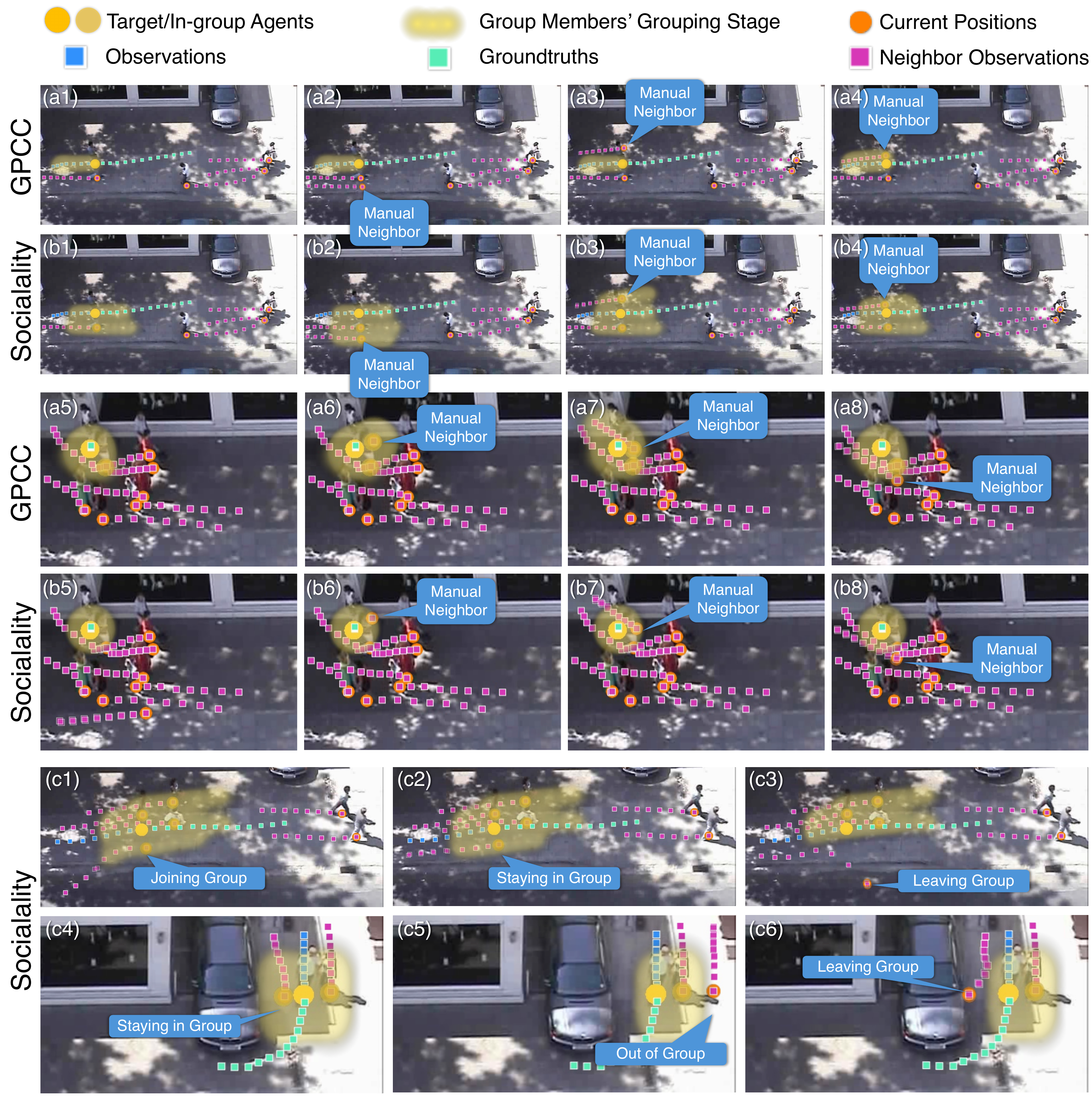}
\caption{
    Examples of grouping decisions of \MODEL~and \NEWMODEL~with manually added neighbors.
    Subfigures (a1)-(a8) and (b1)-(b8) visualize how \MODEL~and \NEWMODEL~respond to manual neighbors placed at different relative positions under the same scenes and target agents.
    Subfigures (c1)-(c6) present representative near-boundary grouping cases, including joining group, staying in group, leaving group, and out of group.
}\label{fig_manual_neighbor}
\end{figure}

\begin{figure}[!htbp]
\centering
\includegraphics[width=1.0\textwidth]{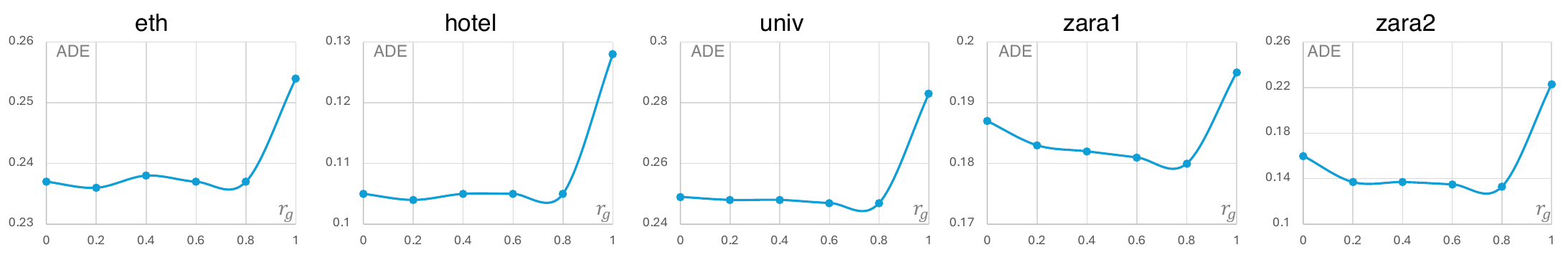}
\caption{
    ADE under different random grouping ratio $r_g$ setting on ETH-UCY dataset.
    }\label{fig_ade_ratio}
\end{figure}

\begin{figure}[!htbp]
\centering
\includegraphics[width=1.0\textwidth]{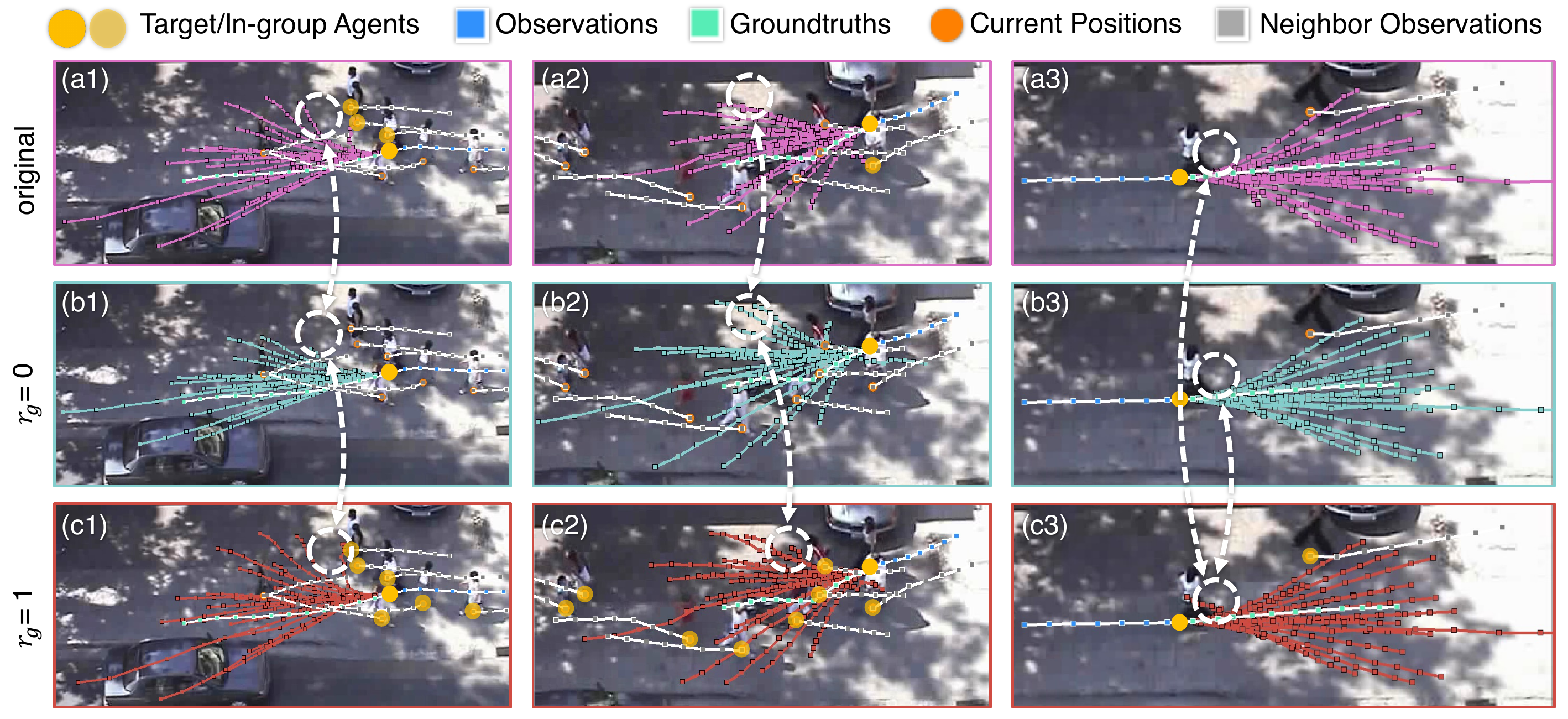}
\caption{
    Visualizations of \NEWMODEL~final predictions under different counterfactual settings.
    }\label{fig_conditioned_predictions}
\end{figure}

\NEWHALFLINE\noindent\textbf{Counterfactual Validations.}
We further conduct a series of counterfactual analyses to verify whether the grouping rules are learned and whether they indeed affect the final prediction results.
First, we manually add neighboring agents \cite{wong2024socialcircle+} to examine how the grouping kernel reacts to different social relations in \FIG{fig_grouping_validations}.
Then, we randomly disturb the grouping structure by assigning different ratios of agents into the same group, and evaluate how the prediction performance changes under these counterfactual grouping settings.
The performance results are reported in \TABLE{tab_random_grouping} and visualized in \FIG{fig_ade_ratio}.
Finally, as shown in \FIG{fig_conditioned_predictions}, we visualize the conditioned predictions to examine whether changes in grouping relations lead to corresponding changes in the predicted trajectories.

\textbf{\emph{(i) Manual-neighbor Intervention:}}
We first conduct manual-neighbor interventions \cite{wong2024socialcircle+} to examine whether the group assignment responds to controlled local social contexts.
Specifically, we manually add a neighboring agent around the target agent and change its relative position and motion pattern.
As illustrated in \EQUA{eq_group_assignment}, the proposed \NEWKERNEL~conducts group decisions $\mathcal{G}_i$ according to each neighboring agent's motion patterns within the extended grouping window $\tilde{\mathbf{\Omega}}$.
If the manually inserted neighbor under different social context settings is reasonably classified, it indicates that the grouping rule is learned.

We visualize several controlled local social contexts in \FIG{fig_manual_neighbor}.
\FIG{fig_manual_neighbor} (a1)-(a8) and (b1)-(b8) show cases where the manual neighbor is placed under different co-walking settings with the target agent.
Additional examples in \FIG{fig_manual_neighbor} (c1)-(c6) illustrate more subtle changes of group assignments under varying motion patterns.

By observing \FIG{fig_manual_neighbor} (a1)-(a4), we find that \MODEL~may apply an overly strict criterion when determining whether a neighbor should be classified as an in-group agent.
In (a2) and (a3), the manual neighbor is already placed relatively close to the target agent and exhibits a plausible co-walking pattern, but it is still excluded from the group assignment.
Only when the closeness reaches the extreme case in (a4) does \MODEL~include it in the target agent's group.
This suggests that the fixed-rule kernel in \MODEL~relies on a strict grouping criterion, which is inconsistent with many realistic social scenarios.
In contrast, \NEWMODEL~produces group assignments that better match human intuition in \FIG{fig_manual_neighbor} (b1)-(b4), indicating that the learned grouping rule can adapt its acceptance boundary according to the local social context rather than relying on a universal fixed threshold.

A complementary failure case of \MODEL~is shown in \FIG{fig_manual_neighbor} (a5)-(a8).
Here, the manual neighbor passes by the target agent within a short period and then diverges, \IE, they do not maintain a sustained co-walking relationship after the brief encounter.
Nevertheless, \MODEL~may mis-classify such a short-time passing neighbor as an in-group agent.
By comparing \FIG{fig_manual_neighbor} (b5)-(b8), we observe that \NEWMODEL~largely avoids this kind of incorrect group assignment.
Even when the manual neighbor comes close for a short time, \NEWMODEL~is less likely to classify it as an in-group agent if the subsequent motion is inconsistent with that of the target agent.
This is consistent with the design of \NEWKERNEL, where group assignment is determined not only by spatial closeness but also by motion consistency within the extended grouping window, yielding more temporally coherent grouping outcomes.

We further present cases in \FIG{fig_manual_neighbor} (c1)-(c6) to reveal how \NEWMODEL~adjusts its group assignment under subtle changes of the manual neighbor.
In (c1), a neighbor that is initially slightly far away becomes an in-group agent after moving in parallel with the target agent for a period (\emph{Joining Group}), while (c2) shows a stable \emph{Staying in Group} status when such co-walking persists.
In contrast, (c3) shows an opposite transition.
A neighbor that initially appears to walk together changes direction and leaves the group, and \NEWMODEL~correspondingly excludes it from the target agent's group assignment (\emph{Leaving Group}).
Finally, \FIG{fig_manual_neighbor} (c4)-(c6) highlights an even more subtle phenomenon.
When the target agent moves relatively slowly, its preference to in-group social distance decreases noticeably.
As a result, the manual neighbor in (c5) is already classified as an out-of-group agent, and this tendency is further emphasized in (c6).
This behavior is also consistent with real-world intuition that for slow pedestrians, being together often indicates a tighter spatial positions.
Overall, these manual-neighbor interventions provide basic evidence that the proposed \NEWKERNEL~learns better grouping rules than the long-term distance kernel in \MODEL~, which relies only on simple spatial proximity.

\textbf{\emph{(ii) Random-grouping Intervention:}}
We further conduct random-grouping interventions to examine whether the final prediction is sensitive to the correctness of group assignments.
Different from the manual-neighbor intervention, which modifies the local social context around a specific target agent, this intervention directly alters the grouping structure at the scene level.
Specifically, we introduce a random grouping ratio $r_g$ to control the proportion of agents that are randomly assigned into the target agent's group in a specific scene.
We sample $r_g$ from $0$ to $1$ with a step size of $0.2$, as reported in \TABLE{tab_random_grouping}.
It should be noted that $r_g=0$ does not denote the original learned grouping strategy.
Instead, it removes the grouping strategy by treating no neighboring agent as an in-group member of the target agent.
In contrast, $r_g=1$ denotes an edge intervention where all neighboring agents are forced into the same group, regardless of their actual social relations.

\begin{table}[t]
\caption{
    Prediction performance under different random grouping ratio $r_g$ setting on ETH-UCY dataset
    }\label{tab_random_grouping}%
\begin{tabular}{@{}lllllll@{}}
\toprule
ID & $r_g$ & eth & hotel& univ& zara1& zara2 \\
\midrule
a1&0&0.237/0.375&0.105/0.160&0.249/0.445&0.187/0.324&0.160/0.283\\
a2&0.2&0.236/0.375&0.104/0.159&0.248/0.441&0.183/0.315&0.137/0.225\\
a3&0.4&0.238/0.376&0.105/0.160&0.248/0.441&0.182/0.313&0.137/0.226\\
a4&0.6&0.237/0.374&0.105/0.162&0.247/0.440&0.181/0.31&0.135/0.223\\
a5&0.8&0.237/0.374&0.105/0.162&0.247/0.441&0.180/0.305&0.133/0.223\\
a6&1&\R{0.254}/\R{0.414}&\R{0.128}/\R{0.221}&\R{0.283}/\R{0.499}&\R{0.195}/\R{0.344}&\R{0.223}/\R{0.372}\\

\midrule
a0 & - &\C{0.233}/\C{0.369}&\C{0.102}/\C{0.156}&\C{0.240}/\C{0.428}&\C{0.168}/\C{0.294}&\C{0.130}/\C{0.223}  \\
\botrule

\end{tabular}
\footnotetext{
    ``-'' denotes that no random grouping ratio is used in the corresponding full a0 model.
    \C{Blue markers} denote the best results on each set.
    \R{Red markers} denote the worst results on each set.
}
\end{table}

By observing \TABLE{tab_random_grouping}, we can see that manually intervening the grouping results noticeably affects the final prediction performance.
The full model a0, which uses the original learned grouping strategy without random intervention, achieves the best performance on all five subsets.
This indicates that the learned grouping rule provides more reliable group relations than randomly assigned grouping structures.
Compared with a0, the $r_g=0$ setting (a1) prediction performance is worse about 7.4\% and 8.0\%.
This suggests that removing group-aware modeling weakens the final prediction, since the target agent can no longer explicitly perceive its in-group members through the proposed group-wise perception mechanism.

The most severe degradation appears when $r_g=1$.
In this case, all neighboring agents are forced into one group, leading to the worst results on all subsets.
Compared with $r_g=0$, the average ADE/FDE further worsens 15.5\% and 16.6\%.
Compared with the full model a0, the performance drop becomes even larger, with the average ADE/FDE worsening 24.1\% and 25.8\%, respectively.
This result indicates that simply considering all neighboring agents as one group is harmful, probably because it largely removes the distinction between in-group and out-of-group interactions.
As a result, the target agent may lose important perception patterns toward different neighboring agents.

For better illustration, we visualize the ADE changes under different $r_g$ settings in \FIG{fig_ade_ratio}.
Across different subsets, we observe a common pattern that the performance first slightly improves when $r_g$ increases from 0 to intermediate values, but then drops sharply when $r_g$ reaches 1.
A possible explanation is that, when no grouping is used ($r_g=0$), the model ignores in-group interactions between the target agent and its true companions.
As $r_g$ gradually increases, some random group assignments may accidentally cover part of the true group relations, which slightly improves the prediction performance.
However, these random assignments are still not as reliable as the learned grouping strategy in a0.
When $r_g=1$, the grouping structure collapses into one group, introducing strong counterfactual priors into the group-wise perception process and leading to clear performance degradation.

Overall, the random-grouping intervention provides quantitative counterfactual evidence that the grouping relation is an effective condition for final trajectory prediction.
Removing grouping or randomly forcing agents into groups both weakens the prediction performance, showing that \NEWMODEL~does not benefit from arbitrary group assignments, but from structured grouping relations.

\textbf{\emph{(iii) Conditioned-prediction Intervention:}}
We finally conduct conditioned-prediction interventions to directly examine how different grouping priors affect the final predicted trajectories.
Different from the above random-grouping intervention, which provides quantitative evidence through ADE/FDE comparisons, here we visualize the prediction results under different grouping settings for a more intuitive analysis.
The corresponding group members are marked by yellow circles in \FIG{fig_conditioned_predictions}.

Specifically, we present prediction results under three settings, including the original grouping strategy without intervention, the $r_g=0$ setting, and the $r_g=1$ setting.
By comparing \FIG{fig_conditioned_predictions} (a1), (b1), and (c1), we observe that the same scene can produce substantially different final predictions under different grouping priors.
In this four-agent group case, when the other in-group agents are no longer treated as group members under $r_g=0$ setting, the predicted trajectories become more concentrated.
A similar phenomenon can also be observed from \FIG{fig_conditioned_predictions} (a2), (b2), and (c2), although the effect appears in an opposite form.
Here, the predictions under $r_g=0$ become more diverse.
This result further indicates that grouping priors do not impose a fixed change on the prediction distribution, but instead act as conditions that reshape the prediction according to the scene context.
A possible explanation is that, when the target agent belongs to a group, its future motion is conditioned not only by its own motion pattern but also by the preference of the whole group.
Therefore, the presence or absence of group members may lead to visibly different prediction distributions.

A particularly interesting case is shown in \FIG{fig_conditioned_predictions} (a3), (b3), and (c3).
In this scene, the target agent is originally walking alone.
Accordingly, when setting $r_g=0$, the final predicted trajectories remain almost unchanged, which is consistent with our expectation since no effective group prior is removed.
However, under $r_g=1$, when another neighboring agent is forcibly treated as belonging to the same group, an additional prediction branch appears toward that neighbor.
Notably, this new trajectory mode implies that the target agent may even make an almost reversed turn to approach the manually assigned group member.

Overall, these conditioned-prediction visualizations provide direct qualitative evidence that grouping priors serve as effective conditions for the \NEWMODEL~Model.

\begin{figure}[!htbp]
\centering
\includegraphics[width=1.0\textwidth]{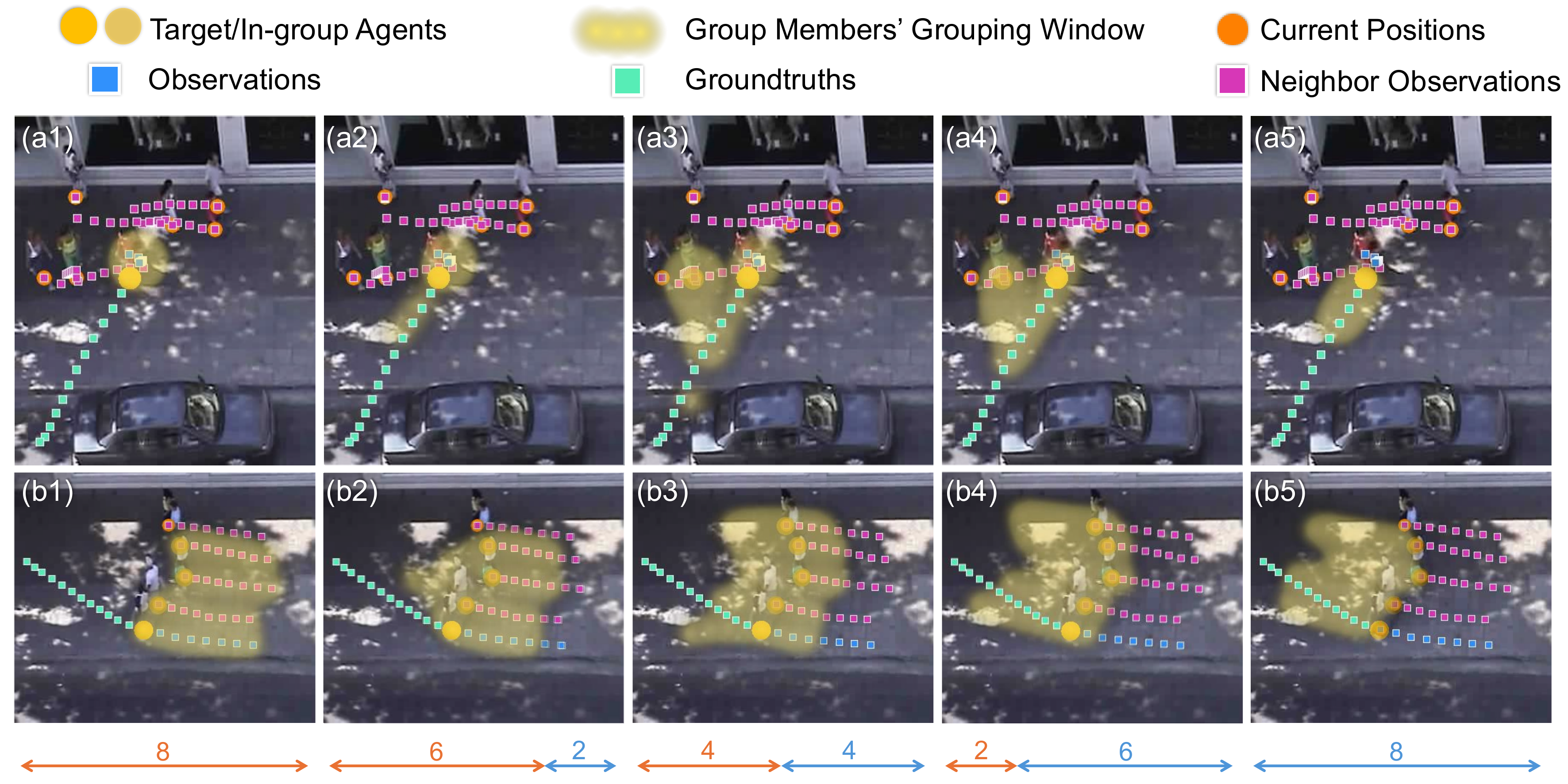}
\caption{
    Visualizations of different grouping results under different grouping windows.
    Subfigures (a1)--(a5) and (b1)--(b5) show two representative cases, where the grouping windows are constructed with different observation-prediction compositions, including $\{8,0\}, \{6,2\}, \{4,4\}, \{2,6\}, \{0,8\}$ observed steps and anticipated steps.
    Yellow shaded regions visualize the grouping windows of current in-group agents.
}\label{fig_grouping_stage}
\end{figure}

\NEWHALFLINE\noindent\textbf{Grouping Window.}
There are two main components within the proposed \NEWKERNEL~kernel, \IE, \ANCHOR~anchors and the extended grouping window.
After validating the capability of learning agent-specific and context adaptive grouping rules of the \ANCHOR~anchors, we further analyze how time-coherent grouping decisions are made when relying on different grouping windows.
As mentioned in \SECTION{sec_method}, \NEWMODEL~jointly integrates \emph{retentions} (historical reviews) and \emph{protentions} (short-term future previews) corresponding to each target as grouping window.
\FIG{fig_grouping_stage} illustrates five temporal segments: the retention-only window, three mixed windows, and the protention-only window.
These segments represent different stages at which group membership may emerge, stabilize, or dissolve.

For the two-agent case in \FIG{fig_grouping_stage} (a1)-(a5), using retention alone (a1) and using limited retentions (a2) fail to include the companion of the target agent.
Using protention alone (a5) also fails, because short-term predictions remain uncertain and do not yet reveal stable joint motion.
Only when both temporal cues are combined (a3) and (a4) does the model correctly group the two agents.
This indicates that group membership is not fully observable from either historical evidence or short-term anticipation alone.

A similar phenomenon appears in the five-agent V-shaped group shown in \FIG{fig_grouping_stage} (b1)-(b5).
Retention alone (b1) mainly captures nearby agents due to distance variation within the formation.
Protention alone (b5) provides insufficient stability to recover the full structure.
When retention and protention are integrated (b2)-(b4), both motion consistency and directional coherence are captured, enabling correct grouping of all members.

These observations also present a limitation of the original long-term distance kernel.
Since grouping is determined only from past spatial proximity with a fixed threshold, the kernel tends to miss group members when historical distances are not consistently small, as shown in \FIG{fig_grouping_stage} (a1) and (b1).
Conversely, relying on short-term future prediction only  is also insufficient, because predicted trajectories alone are not stable enough to represent full group structures, as seen in \FIG{fig_grouping_stage} (a5) and (b5).

The proposed \NEWKERNEL~kernel resolves this issue by jointly using observations and anticipations.
Historical trajectories provide reliable evidence of sustained interaction, while predicted trajectories reveal whether agents will continue moving together.
Combining both cues allows the kernel to include true group members that are slightly separated in the past, while avoiding agents that are only temporarily close.
Therefore, the advantage of \NEWKERNEL~kernel is not merely extending the observation window, but enabling grouping decisions to depend on both past proximity and future motion consistency.
Such grouping results in \FIG{fig_grouping_stage} validate the \NEWKERNEL~kernel's capability of learning temporally coherent grouping rules.

\subsubsection{Discussions of Socialality Anchors}
It should be noted that the learned \ANCHOR~anchors are not designed to directly represent different existing groups.
Instead, they represent agent-specific grouping preferences, \IE, how each target agent adjusts its tolerance to social distance and motion inconsistency when determining group membership.
Therefore, the following analyses focus on whether the learned anchor space is structurally meaningful, and their statistical properties.

The original \MODEL~Model relies on a single fixed threshold $\Gamma$ in the long-term distance kernel, implicitly assuming that a \emph{universal} grouping rule can be applied to all agents and scenes.
However, as discussed in \SECTION{sec_new_kernel}, grouping decisions are not governed by one variable only.
They are jointly shaped by acceptable social distance and moving speed \cite{hall1973hidden,helbing1995social}.
Motivated by this, \NEWMODEL~replaces the fixed $\Gamma$ with two learnable \ANCHOR~anchors, $\tau_i^a$ and $\tau_i^b$.
These anchors serve as parametric containers for agent-specific grouping-rule modifiers.
Therefore, each target agent $i$ can be represented as a point $(\tau_i^a,\tau_i^b)$ in a 2D space.

\begin{figure}[!htbp]
\centering
\includegraphics[width=1.0\textwidth]{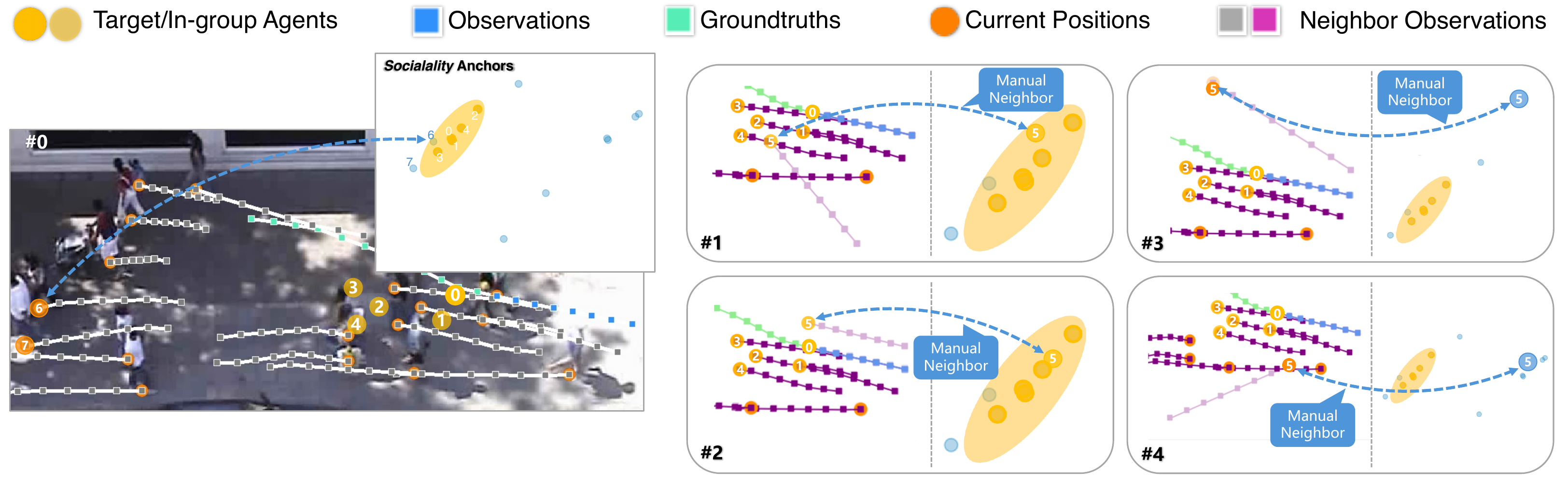}
\caption{
    Visualizations of grouping decisions and anchors.
    Counterfactual interventions of grouping decisions with manually added neighbors are also presented.
    }\label{fig_group_anchor}
\end{figure}

\NEWHALFLINE\noindent\textbf{Grouping Rules of Anchors.}
We first visualize a representative scene from the zara1 subset in \FIG{fig_group_anchor}.
After the group assignment of the proposed \NEWKERNEL~kernel, agents 0--4 are highlighted in the scene, where agent 0 denotes the target agent.

As mentioned in \SECTION{sec_method} and \EQUA{eq_compute_socialality}, we compute the \ANCHOR~anchors of all agents in this scene through the network $m(\cdot)$, and visualize their anchor coordinates in a 2D scatter plot.
We can observe that the anchor coordinates $(\tau_j^a,\tau_j^b), j \in \{0,1,2,3,4\}$, corresponding to the group members assigned by \NEWKERNEL~kernel, are also close to each other in the anchor space.
This indicates that the learned anchors indeed capture the grouping preference of agents, and agents within the same group tend to share similar grouping preferences.
For clearer illustration, we roughly use a yellow mask centered at the anchor coordinate of the target agent, \IE, agent 0, to cover the anchors of agents with the same group membership.

Through this visualization, we can also observe that one blue point close to the target agent in the anchor space does not correspond to an actual group member in the original zara1 scene.
However, by checking their observed trajectories, we find that this agent still shows a motion pattern similar to the target agent, and may therefore share a similar grouping preference.
This observation suggests that the relationship between anchors and group membership is asymmetric.
Agents in the same group are likely to have nearby anchors, but nearby anchors do not necessarily imply that the corresponding agents belong to the same group.
In other words, the anchor space represents grouping-rule preferences rather than direct group labels.

We further visualize the grouping results after adding differentiated manual neighbors, together with the corresponding anchor of the manual neighbor and the original anchors.
Similar to \FIG{fig_manual_neighbor}, we adopt four strategies to add manual neighbors, corresponding to typical group dynamics, including \emph{joining a group}, \emph{staying in a group}, \emph{leaving a group}, and \emph{remaining out of a group}.
By comparing interventions \#1 and \#2 in \FIG{fig_group_anchor}, we can observe that the manual neighbor in \#2, which consistently stays within the group of target agent 0, has an anchor coordinate closer to agent 0 than the manual neighbor in \#1.

Intervention \#3 visualizes a manual neighbor that gradually moves away from the target agent.
This manual neighbor may be grouped with the target agent during an earlier period, but then goes separate ways after a certain point.
Correspondingly, its anchor is no longer located inside the yellow masked region, but still remains relatively closer to the target agent than a completely irrelevant neighbor.
Intervention \#4 visualizes a manual neighbor that is clearly unrelated to target agent 0, with distinct differences in moving direction, speed, and spatial relation.
By comparing the anchor coordinates of the manual neighbors in \#3 and \#4, we can observe that the manual neighbor in \#4 is located in a clearly different region of the anchor space, far from the target agent.
This further indicates that the anchor distance reflects the similarity of grouping preferences among difference agents.

\begin{figure}[!htbp]
\centering
\includegraphics[width=1.0\textwidth]{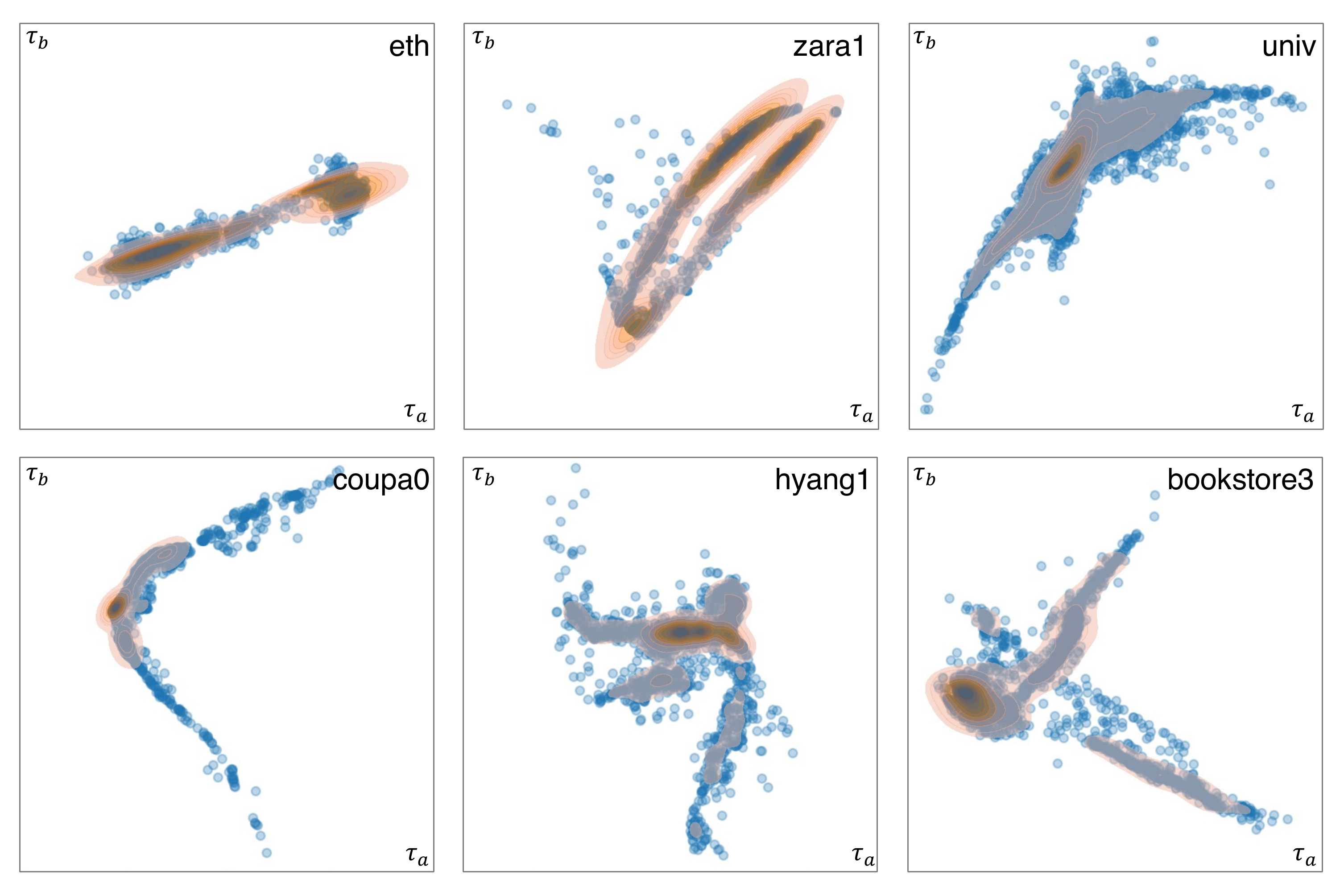}
\caption{
    Visualization of \ANCHOR~anchors in a 2D space across different scenes.
    Each subfigure shows the distribution of \ANCHOR~anchors parameterized by $(\tau^{a}, \tau^{b})$ for one scene.
    Each blue dot denotes one agent's anchor sample, while the orange region indicate their density distribution.
}\label{fig_vis_anchors}
\end{figure}

\NEWHALFLINE\noindent\textbf{Distribution of Anchors.}
\FIG{fig_vis_anchors} visualizes the anchor distributions on multiple subsets of ETH-UCY and SDD.
A first notable phenomenon is that anchors do \emph{not} spread uniformly.
Instead, they concentrate along several narrow ``wings'' (thin-manifold shape), and the patterns could be observed to be varying with different datasets.
For example, eth exhibits a compact diagonal wing where $\tau^a$ and $\tau^b$ increase coherently, suggesting a correlation between distance tolerance and speed flexibility across agents.
In contrast, zara1 forms two separated wings, indicating that the model might prefer two regimes of grouping rules, which might share common tendency since two separated wings are spatially close and almost parallel.
Within SDD, the structures become more diverse.
For example, coupa0 shows a curved wing (a ``C''-shape in 2D space), hyang1 displays multiple wings with a noticeable downward tail, and bookstore3 presents a multi-wing pattern where wings seem to grow from a dense core.
Such wing-shaped concentrations imply that the anchors are not arbitrary free parameters; rather, they converge according to specific grouping-rule to shape different patterns.

Second, the number of wings vary substantially across scenes.
ETH-UCY subsets tend to present one or two dominant wings (e.g., the single compact wing in eth and the two-wing structure in zara1), whereas SDD subsets often exhibit richer multi-wing structures (e.g., hyang1 and bookstore3).
This difference suggests that the learned anchors capture scene-specific mixtures of motion modes.
Specifically, fewer wings indicate more similar grouping behaviors, while multiple wings indicate that agents in the scene may follow multiple distinct movement and interaction styles, which demand different distance-speed anchor combination to yield stable grouping.

Overall, the anchor distribution provides an interpretable illustration of how \NEWMODEL~learns \emph{agent-specific} and \emph{context-adaptive} grouping rules, suggesting that human grouping preferences in social scenes can be parameterized by the two anchors that jointly regulate distance tolerance and speed flexibility in a 2D space.

\begin{figure}[!htbp]
\centering
\includegraphics[width=1.0\textwidth]{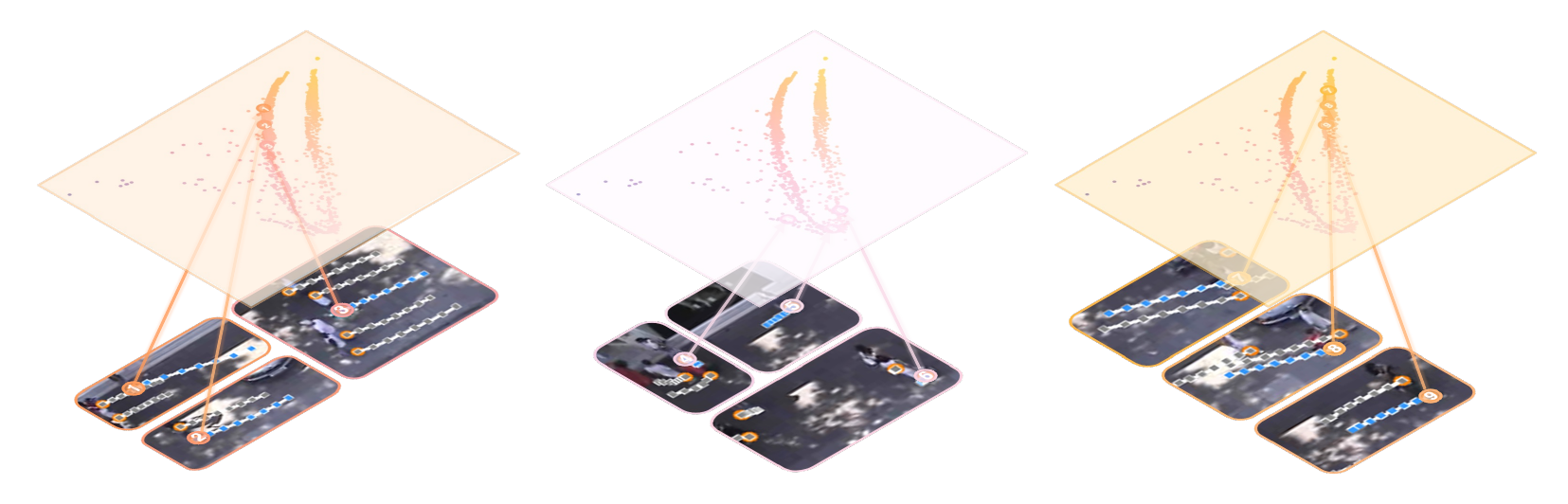}
\caption{
    Structural visualization of \ANCHOR~anchors distributions evaluated on ETH-UCY zara1.
    Different colors indicate different regions in the same distribution figure.
    We choose three representative examples from each region here.
    Best viewed in color.
    }\label{fig_anchor_structure}
\end{figure}

\NEWHALFLINE\noindent\textbf{Structural Analyses of Anchors.}
To connect the anchor distribution to physical behaviors, we map representative anchor locations back to the scene in \FIG{fig_anchor_structure}.
Here, we take \ANCHOR~anchor distribution in zara1 as example.
The anchor distribution forms a clear V-shape with two wings.
We select several targets agent corresponding to the anchor coordinates on different parts of the V-shape pattern and visualize these target agents and their neighbors' observed trajectories.

In \FIG{fig_anchor_structure}, we can observe that targets (7, 8, and 9 in the right subfigure) consistently move to the right, while targets (1, 2, and 3 in the left subfigure) consistently move to the left.
In contrast, samples (4, 5, and 6 in the middle subfigure) exhibit a slow-motion pattern.
Both target agents (4) and (6) remain nearly stationary during the observation window.
This separation emerges without any motion-mode supervision, since the anchors only affect grouping through the \NEWKERNEL~kernel.

Within each wing of this V-shape pattern, we observe an ordering related to walking speed.
For the right-moving-agent wing, the target agents (7) and (8) move faster than agent (9).
For the left-moving-agent wing, the target agents (1) and (2) move faster than that of agent (3).
In both cases, faster examples, \IE, (1), (2), (7), and (8), lie closer to the wing boundary, while slower examples, \IE, (3) and (9), lie closer to the base of the wing.
For near-stationary targets (4), (5), and (6) in the middle subfigure, the displacement is very small.
Accordingly, these target agents' anchor positions occupy the bottom region of the V-shape.
A plausible explanation is that higher speed increases neighborhood variability and short-lived encounters, requiring a different balance between distance tolerance ($\tau_i^a$) and speed-flexibility tolerance ($\tau_i^b$), which shifts the anchor position along the wing.

\begin{figure}[!htbp]
\centering
\includegraphics[width=1.0\textwidth]{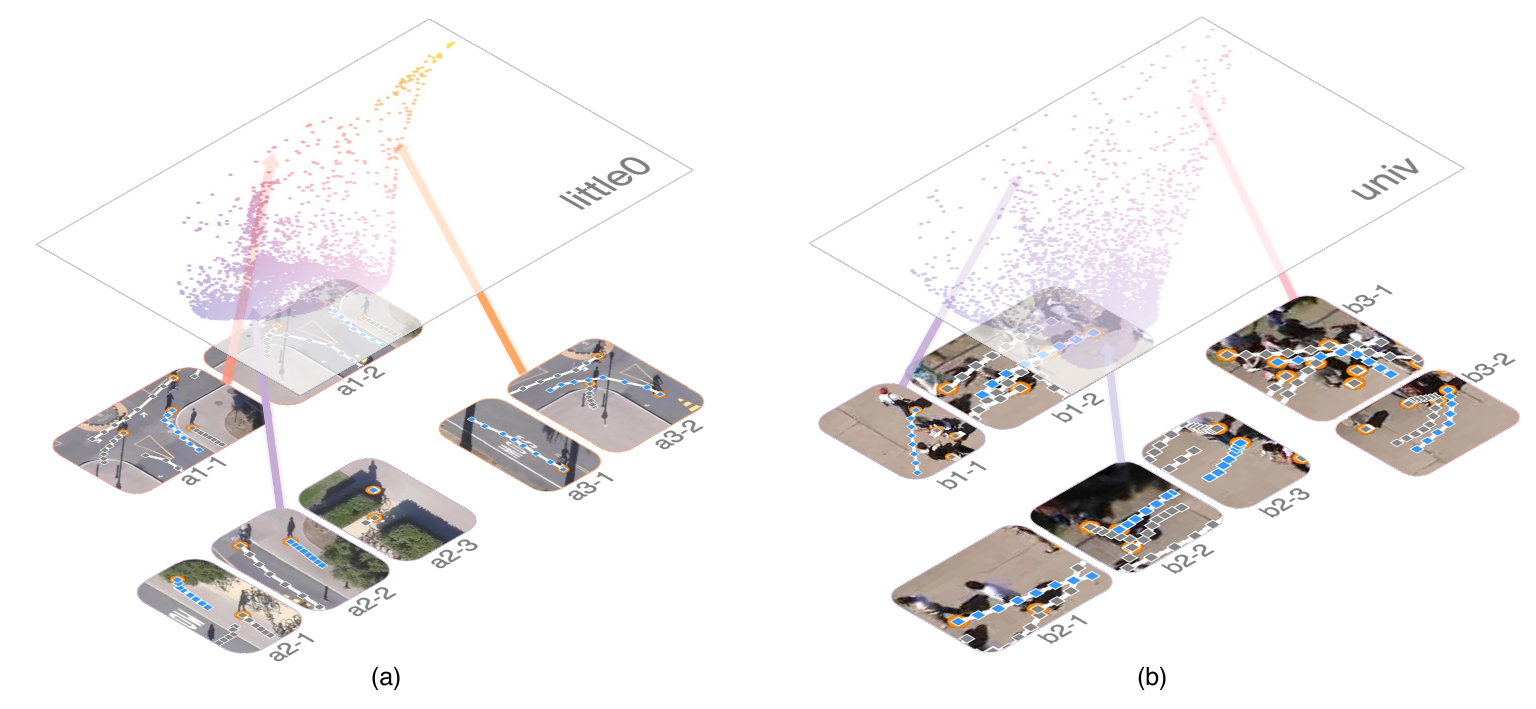}
\caption{
    Structural visualization of \ANCHOR~anchors distributions evaluated on SDD little0 (a) and ETH-UCY univ (b).
    Different colors indicate different regions in distribution figures.
    We choose several representative examples from each region here.
    Best viewed in color.
    }\label{fig_anchor_structure_other}
\end{figure}

The anchor distributions in the other two subsets in \FIG{fig_anchor_structure_other} reveal that such structural relations with motion modes is not unique to zara1.
In univ (b), the anchor distribution is more dispersed and continuous, suggesting a more diverse mixture of local interaction regimes.
Nevertheless, nearby anchor regions still correspond to target agents with similar motion tendencies and neighborhood configurations, indicating that local smoothness in anchor space is preserved.
In SDD subset little0 (a), the anchor distribution exhibits a multi-wing structure rather than a simple two-wing pattern, implying that the learned anchors adapt to richer motion modes in that scene.
Even in this more complex case, different wings remain associated with visually distinguishable motion behaviors as demonstrated in \FIG{fig_anchor_structure_other}.

Overall, \FIG{fig_anchor_structure} and \FIG{fig_anchor_structure_other} provides direct structural evidence that the learned anchor space is behaviorally meaningful.
This suggests that the two-anchor design offers a compact yet flexible representation for agent-specific and context-adaptive grouping rules in real scenes.

\NEWHALFLINE\noindent\textbf{Statistical Analyses of Anchors.}
\label{sec_anchor_statistics}
In this section, we analyze the statistical and geometric structure of the learned \ANCHOR~anchors.
First, we compute the mean, variance, covariance, and correlation coefficient.
Given $N_b$ samples $\{(\tau^a_i,\tau^b_i)\}_{i=1}^{N_b}$, the covariance is defined as
\begin{equation}
\mathrm{Cov}(\tau^a,\tau^b)
=
\frac{1}{N_b}
\sum_{i=1}^{N_b}
\left(\tau^a_i - \mu_a\right)
\left(\tau^b_i - \mu_b\right),
\end{equation}
where $\mu_a$ and $\mu_b$ denote the means of $\tau^a$ and $\tau^b$, respectively.
The corresponding correlation $\rho_{ab}$ is calculated as
\begin{equation}
\rho_{ab}
=
\frac{\mathrm{Cov}(\tau^a,\tau^b)}
{\sqrt{\mathrm{Var}(\tau^a)\,\mathrm{Var}(\tau^b)}}.
\end{equation}

\begin{figure}[!htbp]
\centering
\includegraphics[width=1.0\textwidth]{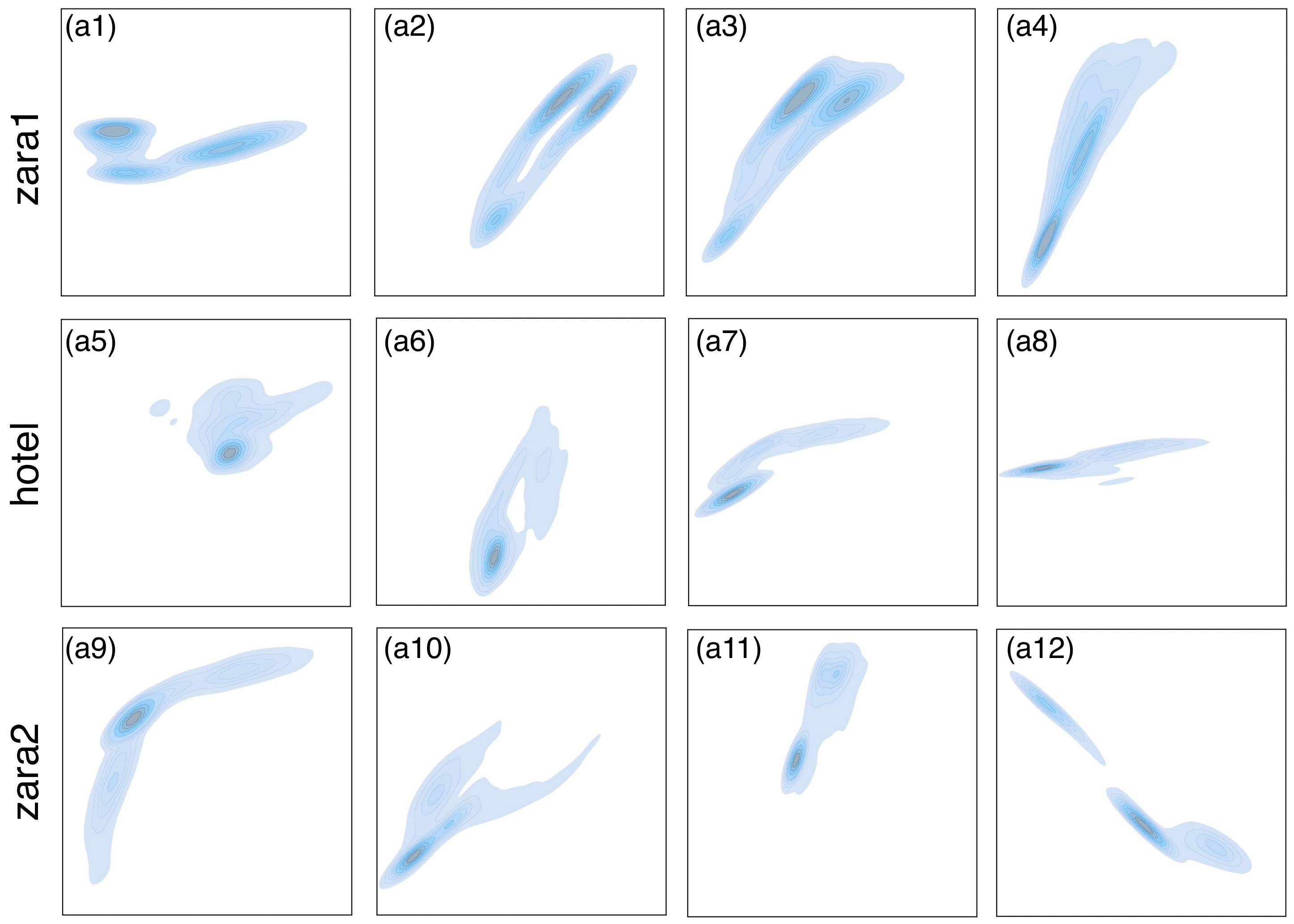}
\caption{
    Anchor distributions learned from different independent runs.
    Each row corresponds to one scene (zara1, hotel, and zara2), while each subfigure shows the anchor distribution density.
}\label{fig_different_runs_anchors}
\end{figure}

\TABLE{correlation} lists correlation $\rho_{ab}$ across different clips and training runs.
Non-independence implies that the correlation coefficient is not zero, while non-perfect correlation implies that the correlation coefficient is not equal to $1$ or $-1$.
From \TABLE{correlation}, we can infer that the experimental results consistently bounded away from $0$ and $1$, which indicates that the two different anchors are neither completely independent nor perfectly correlated.
From another perspective, this also supports our conclusion in the ablation study in \TABLE{tab_rules_ablations} that a fixed anchor (distance in \MODEL) is insufficient to describe the space where grouping rules reside.
The butterfly-like multi-wing shape in \FIG{fig_different_runs_anchors} and \FIG{fig_vis_anchors}, where the manifold is neither a linear straight line nor randomly occupies the entire 2D space, further illustrates this phenomenon, \IE, the distribution lies on a specific butterfly-shaped manifold.
This suggests that these two anchors are already capable of representing grouping rules for different agents, rather than requiring three or more anchors.
In other words, additional anchors would be redundant.
Introducing a third anchor would increase parameterization without introducing new statistically supported degrees of freedom.

\begin{table}[t]
\caption{The correlation $\rho_{ab}$ across different subsets and training runs.}\label{correlation}%
\begin{tabular}{@{}lccccccc}
\toprule
clip & hyang0 & hyang12 & nexus11 & nexus5 & nexus10 \\
\midrule
$\rho_{ab}$ (train\#1)&0.268 & 0.169 & 0.288 & 0.839 & 0.168 \\
$\rho_{ab}$ (train\#2)&0.435 & -0.810 & 0.862 & 0.608 & 0.118 \\
$\rho_{ab}$ (train\#3)&0.518 & 0.465 & 0.356 & 0.839 & 0.309 \\
\midrule
clip & coupa3 & bookstore0 & gates0 & deathCircle0 & little0 \\
\midrule
$\rho_{ab}$ (train\#1)&0.277 & 0.527 & 0.243 & 0.637 & 0.703 \\
$\rho_{ab}$ (train\#2)&-0.566 & -0.489 & 0.642 & 0.548 & -0.259 \\
$\rho_{ab}$ (train\#3)&-0.030 & 0.273 & 0.2436 & 0.638 & 0.703 \\
\botrule
\end{tabular}
\end{table}

To further understand the intrinsic dimensionality of the anchor space, we consider the covariance matrix
\begin{equation}
\Sigma
=
\begin{pmatrix}
\mathrm{Var}(\tau^a) & \mathrm{Cov}(\tau^a,\tau^b) \\
\mathrm{Cov}(\tau^a,\tau^b) & \mathrm{Var}(\tau^b)
\end{pmatrix}.
\end{equation}

This suggests that the anchor distribution lies close to a low-dimensional manifold embedded in $\mathbb{R}^2$.

Furthermore, from the polar coordinates perspective, in \FIG{fig_vis_anchors}, we observe that the angular variable $\theta$ concentrates within a small angular range, while the radial magnitude $r$ varies more freely.
This separation between direction and magnitude indicates that the system primarily modulates along a dominant direction, with limited orthogonal variation.

The above findings indicate that the learned socialality representation occupies an intermediate regime:
it is neither purely one-dimensional nor genuinely two-dimensional.
For intuitive illustration, we refer to this as ``1.5D''.
It should be noted that this terminology is descriptive rather than formal, and does not imply a strict fractal or Hausdorff dimension \cite{falconer2013fractal}.
These phenomena further indicate that human spatial social activities might be closely related to the two factors corresponding to these two \ANCHOR~anchors, distance and speed, and may also provide new attempts for studying human behaviors or inspiring other human-inspired research.

After establishing the statistical structure of the learned anchors, we further examine whether their geometric organization is stable across independent training runs.
Here, we repeatedly train \NEWMODEL~Model on the same dataset with random initializations, and visualize the resulting anchor distributions in \FIG{fig_different_runs_anchors}.

We observe that although the exact patterns vary from run to run, the learned anchor distributions are not arbitrary.
Such variations mainly affect the fine-grained appearance of the distributions, rather than their underlying structural information.
Each dataset tends to repeatedly exhibit a consistent geometric pattern.
For example, on zara1, the anchors consistently form two-wing-like structures like a letter ``V''; on hotel, they remain concentrated within a relatively compact region with limited directional spread; and on zara2, they repeatedly organize into bent shapes, similar to zara1.
Notably, the \ANCHOR~anchors are learned without explicit geometric supervision.
These results support the stabilizability of the proposed \NEWKERNEL~kernel.
The anchor distributions are not fragile artifacts caused by a particular random seed, but reproducible structures that repeatedly emerge under the same dataset.
This indicates that \NEWKERNEL~kernel can consistently capture intrinsic socialality patterns from data, thereby providing a stable basis for the \NEWMODEL~Model.

Overall, these analyses suggest that the learned \ANCHOR~anchors provide a compact and interpretable parameterization of agent-specific and context-adaptive grouping preferences.
The anchor space captures how different agents modulate acceptable social distance and in-group common speed under different social contexts.
The structured distributions and patterns across independent training runs together support the interpretability and stability of the proposed \ANCHOR~anchor design.


\section{Conclusion and Limitations}
In this manuscript, we propose the \NEWKERNEL~kernel and the corresponding \NEWMODEL~trajectory prediction model to further expand \MODEL~\cite{zou2024who} by explicitly obtaining different agents' \emph{agent-specific}, \emph{temporally coherent}, and \emph{context-adaptive} social boundaries for grouping when forecasting trajectories.
The \NEWKERNEL~kernel uses two learnable coefficients $(\tau_i^a,\tau_i^b)$, \IE, \ANCHOR~anchors, to jointly modulate acceptable social distance and speed flexibility within each group over an extended grouping window.
The boundaries anchored by the \ANCHOR~anchors determines the group affiliation, which serves as the group priors conditioned to interaction modeling in the subsequent perception mechanism.
We validate the \NEWMODEL~Model on standard trajectory prediction benchmarks and observe consistent improvements over strong baselines, indicating that the proposed grouping mechanism yields measurable performance gains.
We further assess the proposed \ANCHOR~anchors from an interpretability and robustness perspective.
Through qualitative visualizations, the learned anchors exhibit coherent and scene-consistent structures that align with recognizable different motion modes.
These analyses jointly validate the interpretability, stability, and practical effectiveness of the learned \ANCHOR~anchors. 
The analyses of two \ANCHOR~anchors indicate that human spatial social activities might be related to the two factors corresponding to these two \ANCHOR~anchors, distance and speed, and may also provide new attempts for studying human behaviors or inspiring other human-inspired research.

Despite these promising results, our framework has certain technical limitations. 
Based on the current two-stage network structure, the proposed \NEWMODEL~aims to obtain the group priors first before interaction modeling.
This sequential paradigm of grouping before modeling effectively leverages the obtained group priors in pedestrian-dominated scenes.
However, the impact of such group priors might be less evident in scenes where the primary agents are vehicles.
This suggests that the proposed method may exhibit certain limitations when generalized to vehicle trajectory prediction datasets.
Accordingly, exploring and designing a more universal prior to accommodate complex scenes in a human-inspired manner will be one of our future considerations.

\section*{Declarations}

\begin{itemize}
\item Conflict of interest statement: The authors declare that they have no known competing financial interests or personal relationships that could have appeared to influence the work reported in this manuscript.
\item Funding Information: This work is supported by the National Natural Science Foundation of China (Grant 62172177).
\item Data availability: There is no additional data associated with this manuscript. Our code is available at \url{https://github.com/LivepoolQ/Socialality}.
\end{itemize}

\FloatBarrier

\bibliography{ref.bib}

\end{document}


\begin{appendices}


\documentclass[../paper.tex]{subfiles}

\begin{document}

\section{Nomenclature Table}

To assist readers in better understanding our formulations, particularly the temporal slicing operations required in our framework, we provide a nomenclature table here.

\begin{table}[h]
    \centering
    \caption{Nomenclature of the key variables in the proposed methodology}
    \label{tab_nomenclature}
    \begin{tabular}{@{}ll@{}}
        \toprule
        \textbf{Symbol} & \textbf{Description} \\
        \midrule
        \multicolumn{2}{@{}l}{General} \\
        $N_e$ & Total number of agents in the scene \\
        $i, j$ & Indices for the target agent and its neighboring agent \\
        $t$ & Time step (current observation step is set to $t=0$) \\
        $\epsilon$ & An infinitesimal quantity \\
        $K_f, K_g$ & Trajectory generation number for main and short-term prediction networks \\
        \midrule
        \multicolumn{2}{@{}l}{Time Windows} \\
        $t_o, t_p$ & Number of discrete observation and prediction time steps \\
        $\Omega_0$ & Observation time window $\{-t_o+1, \dots, -1, 0\}$ \\
        $\hat{\Omega}$ & Anticipation (short-term future preview) time window \\
        $\tilde{\Omega}$ & Extended temporal grouping window ($\tilde{\Omega} = \Omega \cup \hat{\Omega}$) \\
        \midrule
        \multicolumn{2}{@{}l}{Trajectories} \\
        $\mathbf{p}_i^t \in \mathbb{R}^2$ & 2D coordinate of agent $i$ at time step $t$ \\
        $\mathbf{d}_i$ & Heading direction of agent $i$ \\
        $\mathbf{X}_i, \mathbf{Y}_i$ & Ground-truth observed and future trajectories of agent $i$ \\
        $\hat{\mathbf{Y}}_i$ & Predicted multi-styled future trajectories \\
        $d_i^t(j)$ & Step-wise distance between agent $j$ and $i$ at time $t$ \\
        $p_i(\tilde{\Omega})$ & Displacement of agent $i$ during time window $\tilde{\Omega}$ \\
        $\rho_i(j|\tilde{\Omega})$ & Walking speed difference ratio between agent $j$ and $i$ \\
        \midrule
        \multicolumn{2}{@{}l}{Groups and Perceptual Sets} \\
        $\mathcal{N}_i$ & Set of all neighboring agents (group candidates) for target agent $i$ \\
        $\mathcal{G}_i, \overline{\mathcal{G}}_i$ & Sets of in-group agents and out-of-group agents for agent $i$ \\
        $\mathbf{FOV}_i$ & Field of view angular region for agent $i$ \\
        $\mathcal{F}_i$ & Set of in-FOV neighboring agents for agent $i$ \\
        $\mathcal{C}_i^s$ & Sets of out-of-group agents in region $s \in \{\mathrm{right}, \mathrm{left}, \mathrm{rear}\}$ \\
        \midrule
        \multicolumn{2}{@{}l}{Grouping Kernels} \\
        $\tau^a_i, \tau^b_i$ & Socialality anchors modulating acceptable social distance and speed difference \\
        $\Gamma$ & Static distance threshold in the long-term distance kernel \\
        $\mathcal{K}_i$ & Grouping kernel function \\
        $\mathcal{S}^a_i, \mathcal{S}^b_i$ & Grouping condition indicators modulated by anchors $\tau^a_i$ and $\tau^b_i$ \\
        $c_1, c_2, c_3$ & Learnable modulation coefficients for social cues fusion \\
        \midrule
        \multicolumn{2}{@{}l}{Features and Representations} \\
        $\mathbf{f}_i^e, \mathbf{f}_i^g, \mathbf{f}_i^{\overline{g}}$ & Self representation, in-group representation, and out-of-group representation \\
        $\mathbf{r}_i$ & Concatenated perception vector of region-level motion cues \\
        $\mathbf{f}_i$ & Final fused feature representation for trajectory prediction \\
        \botrule
    \end{tabular}
\end{table}

\section{Discussions of Short-term Prediction Network}

\subsection{Ego Loss Ratio}

\documentclass[../../paper.tex]{subfiles}
\begin{document}

\begin{figure}[!htbp]
\centering
\includegraphics[width=1.0\textwidth]{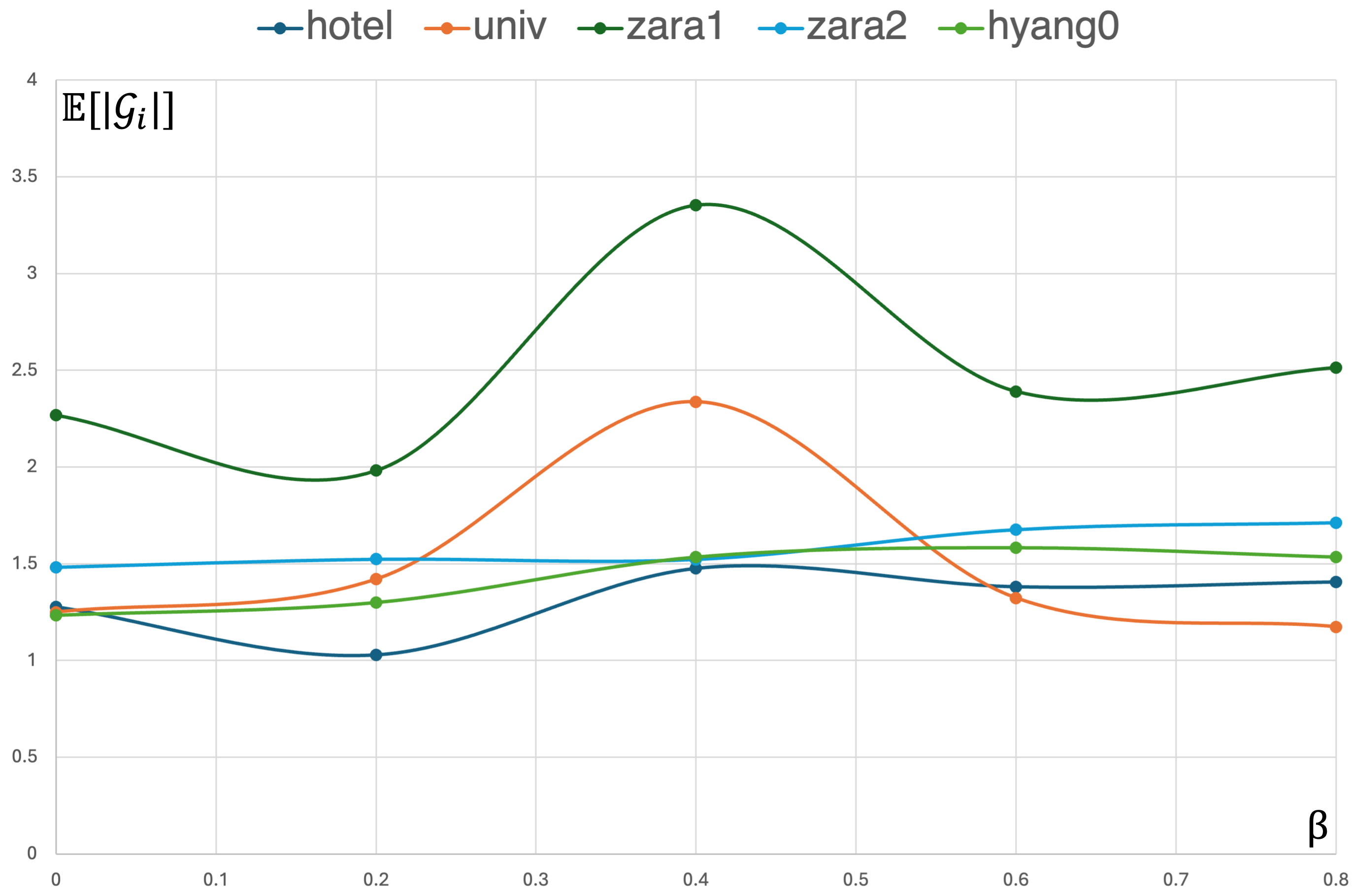}
\caption{
    Effect of the ego loss ratio $\beta$ on the average number of group members across different scenes.
    The x-axis denotes the ego loss ratio, the y-axis denotes the average number of group members, and each curve corresponds to one scene.
}\label{fig_elr_visualization}
\end{figure}

\end{document}

In this section, we further discuss the parameterization of the short-term prediction network\footnote{Following Wong \ETAL \cite{wong2026encore}, we generate the agent-specific \emph{ego rehearsals} as short-term previews for temporary extension.} in \NEWMODEL.
\FIG{fig_elr_visualization} investigates how the average number of inferred group members changes as we vary the ego loss ratio $\beta$ across different ETH-UCY and SDD subsets.
The calculation of average number of inferred group members can be formulated as
\begin{equation}
    \mathbb{E}[|\mathcal{G}_i|]= \sum_{i \in \{1, 2, \dots, N_e\}} \frac{|\mathcal{G}_i|}{N_e}.
\end{equation}
As mentioned in the Method section, the original $\ell_{2}$ loss is fixed with a unit coefficient, the ego loss ratio $\beta$ still substantially affects the learned grouping behaviors.

Grouping decisions depend critically on the stability of the ego-centric relative motion.
When the ego loss ratio is too small, ego predictions are weakly supervised, thus remaining relatively noisy.
As a result, the grouping kernel tends to make conservative and inconsistent membership decisions, preventing the formation of larger and coherent groups.
Increasing the ego loss ratio to a moderate regime (around $0.4$) effectively stabilizes ego predictions, providing a reliable motion reference that enables the grouping kernel to expand groups more confidently and consistently.
This effect is most prominent on interaction-heavy subsets such as univ and zara1, where the grouping structures are more common.

Notably, further increasing the ego loss ratio beyond $0.4$ does not further improve grouping.
Instead, on subsets such as univ, the inferred group size decreases sharply when the ego loss ratio becomes large (e.g., $0.6$-$0.8$).
In this case, grouping becomes more cautious that neighbors are less likely to be assigned into the target agent's group, leading to smaller inferred groups.
Therefore, the peak around $0.4$ can be interpreted as the balance point between concentration on self and social behaviors.

Different subsets also exhibit distinct sensitivities to the ego loss ratio.
Interaction-heavy scenes such as zara1 and univ present clear peaks around $0.4$.
In contrast, subsets such as hotel, zara2, and hyang0 show relatively mild variations, implying that their social structures are less dependent on expanding group membership.

Overall, analyses above further validate the effectiveness of setting the ego loss ratio $\beta=0.4$.
In summary, setting the ego loss ratio to $0.4$ can be viewed as a trade-off across subsets.
It is large enough to learn consistent grouping priors in highly interactive scenes, and meanwhile it does not cause noticeable degradation on subsets with weaker interactions.

\subsection{Short-term Prediction Intervention}

As the short-term predictions are leveraged for grouping, it is necessary to analyze the reliability of the short-term predictions and its impact on the overall results.
To systematically address this, we design a set of intervention experiments.
Following the notation in Sec. 3.1, let $\hat{\mathbf{X}}_j(\hat{\Omega})$ denote the original forecasted short-term trajectory for neighboring agent $j$.
We introduce three types of interventions to generate the intervened previews, which are then concatenated with observed reviews $\mathbf{X}_j(\Omega)$ to form the extended trajectory $\mathbf{X}_j(\tilde{\Omega})$ for the \NEWKERNEL~kernel.
We formulate these interventions using the do-calculus as follows.

\NEWHALFLINE\noindent\textbf{Noise Injection.}
The noise injection intervention aims to evaluate the model's performance change to different degrees of short-term prediction deviations, which simulates the unreliable previews generated from short-term prediction network.
Here, we inject zero-mean Gaussian noise into the short-term previews of agent $j$ to perform this intervention.
Formally, for all neighboring agent $j \in \mathcal{N}_i$,
\begin{equation}
    \begin{split}
        &do(\hat{\mathbf{X}}_j(\hat{\Omega}) = \hat{\mathbf{X}}_j^{\text{noise}}(\hat{\Omega})), \quad \text{where}~ \\
        &\hat{\mathbf{X}}_j^{\text{noise}}(\hat{\Omega}) = \hat{\mathbf{X}}_j(\hat{\Omega}) + \boldsymbol{\epsilon}.
    \end{split}
\end{equation}
Here, $\boldsymbol{\epsilon}$ is a matrix sharing the same shape with $\mathbf{X}_j(\hat{\Omega})$.
Each element in matrix $\boldsymbol{\epsilon}$ is an independently sampled noise, which satisfies $\boldsymbol{\epsilon}_{ij} \overset{\mathrm{i.i.d.}}{\sim} \mathcal{N}(0 , \sigma^2)$.

\NEWHALFLINE\noindent\textbf{Trajectory Swapping.}
The trajectory swapping intervention aims to test the effect of assigning entirely irrelevant previews to agents, which simulates replacing agents' trajectories.
Here, we introduce a shuffle ratio $r_s \in [0, 1]$ to determine the proportion of agents among all neighbors $\mathcal{N}_i$ that are subjected to swapping. 
Specifically, we randomly sample a subset of neighboring agents $\mathcal{N}^\text{swap}_i \subseteq \mathcal{N}_i$. 
We then shuffle the assigned trajectories among the selected agent $j \in \mathcal{N}^\text{swap}_i$. 
$\overline{j}$ indicates the original neighbor index is shuffled from $j$ to $\overline{j} \in \mathcal{N}^\text{swap}_i$.
We perform this intervention as follows:
\begin{equation}
    \begin{split}
        &do(\hat{\mathbf{X}}_j(\hat{\Omega}) = \hat{\mathbf{X}}_j^{\text{swap}}(\hat{\Omega})), \quad \text{where}~ \\
        &\hat{\mathbf{X}}_j^{\text{swap}}(\hat{\Omega}) = 
        \begin{cases}
            \hat{\mathbf{X}}_{\overline{j}}(\hat{\Omega}), & \text{if } j \in \mathcal{N}^\text{swap}_i \\
            \hat{\mathbf{X}}_j(\hat{\Omega}), & \text{otherwise}.
        \end{cases}
    \end{split}
\end{equation}

\NEWHALFLINE\noindent\textbf{Trajectory Flipping.}
The trajectory flipping intervention aims to simulate directional failures of the short-term prediction network output.
Here, we introduce a flip ratio $r_f \in [0, 1]$ to determine the proportion of agents among all neighbors $\mathcal{N}_i$ that are subjected to flipping.
Similarly, we also randomly sample a subset of neighboring agents $\mathcal{N}^\text{flip}_i \subseteq \mathcal{N}_i$.
For any selected agent $j \in \mathcal{N}^\text{flip}_i$, we mirror its predicted future trajectory according to its last observed position $\mathbf{p}_j^0$.
Specifically, for any future step $t \in \hat{\Omega}$, the flipped coordinate is symmetrically computed as $2\mathbf{p}_j^0 - \hat{\mathbf{p}}_j^t$.
We perform this intervention as follows:
\begin{equation}
    \begin{split}
        &do\left(\hat{\mathbf{X}}_j(\hat{\Omega}) = \hat{\mathbf{X}}_j^{\text{flip}}(\hat{\Omega})\right), \quad \text{where}~ \\
        &\hat{\mathbf{X}}_j^{\text{flip}}(\hat{\Omega}) = 
        \begin{cases}
            \big\{ 2\mathbf{p}_j^0 - \hat{\mathbf{p}}_j^t \big\}_{t \in \hat{\Omega}}, & \text{if } j \in \mathcal{N}^\text{flip}_i \\
            \hat{\mathbf{X}}_j(\hat{\Omega}), & \text{otherwise}.
        \end{cases}
    \end{split}
\end{equation}

\begin{figure}[h]
\centering
\includegraphics[width=1.0\textwidth]{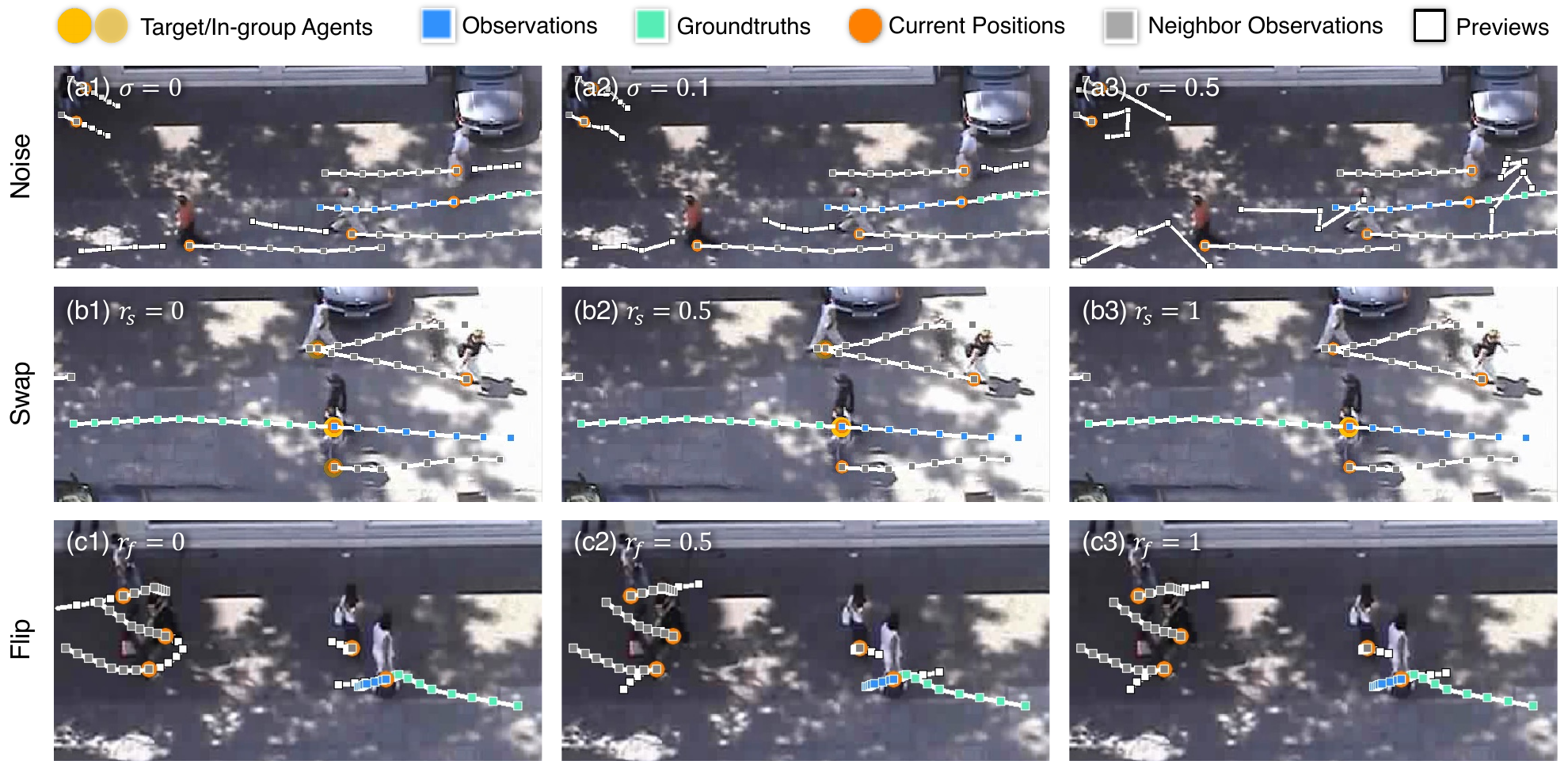}
\caption{
    Intervention examples under different intervention parameters in each corresponding intervention method.
    }\label{fig_intervention_example}
\end{figure}

By conducting the interventions above, we can quantitatively observe how the short-term predictions affect the subsequent grouping and the trajectory prediction performance.
We visualize some examples corresponding to each intervention manner to better illustrate how we intervene the short-term predictions in \FIG{fig_intervention_example}.

The effects of noise injection can be easily observed in \FIG{fig_intervention_example} (a1)-(a3).
We can observe that the previews, \IE, the short-term predictions, in (a2) and (a3) exhibit varying degrees of fluctuation compared to the baseline in (a1).
In (a2), when $\sigma=0.1$, this fluctuation appears relatively moderate, whereas the previews in (a3) already display random coordinate jumps.
The trajectory flipping applied to different proportions of neighbors can also be observed in \FIG{fig_intervention_example} (c1)-(c3).
Specifically, we validate the trajectory swapping intervention by visualizing the target agent's group members.
In (b2), swapping the trajectories of a subset of neighbors reduces the original two in-group agents to one, while swapping all of them in (b3) completely reduces the in-group members to zero.
It should be noted that since a certain proportion of neighbors is randomly selected, the specific group priors shown in (b2) are only one of the possibilities.
Next, we conduct quantitative experiments for each intervention strategy under various parameter settings to evaluate their impact on the overall performance.
The intervention of the short-term predictions would directly change the subsequent grouping process.
To quantify how the intervened variants' group priors differ from the base \NEWMODEL, we propose the group consistency metric to calculate the ratio of the target agents that have the same group priors as the base model.

\begin{table}[h]
\footnotesize
\caption{
    Intervention experiments of the short-term predictions on the zara1 dataset
    }\label{tab__previews_interventions}%
\setlength{\tabcolsep}{4pt}
\begin{tabular}{@{}l|l|c|ccc@{}}
\toprule
Intervention & ID & Param & zara1 (ADE/FDE) & $\Delta\%$ (ADE/FDE)& Group Consistency\\

\midrule
Noise Injection ($\sigma$) & n1 & 0.1 & 0.1710 / 0.2940 & +0.43\% / +0.06\% & $89.98\%$\\
 & n2 & 0.5 & 0.1814 / 0.3069 & +6.29\% / +4.45\% &$60.23\%$\\
 & n3 & 1.0 & 0.2019 / 0.3372 & +18.29\% / +14.74\% &$33.23\%$\\
 & n4 & 2.0 & 0.2219 / 0.3800 & +30.05\% / +29.31\% &$9.19\%$\\
\midrule
Trajectory Swapping ($r_s$) & s1 & 0.2 & 0.1884 / 0.3274 & +10.41\% / +11.42\% &$41.61\%$\\
 & s2 & 0.5 & 0.2136 / 0.3771 & +25.14\% / +28.34\% &$11.04\%$\\
 & s3 & 0.8 & 0.2401 / 0.4282 & +40.67\% / +45.74\% &$1.77\%$\\
 & s4 & 1.0 & 0.2556 / 0.4556 & +49.78\% / +55.06\% &$0.35\%$\\
\midrule
Trajectory Flipping ($r_f$) & f1 & 0.2 & 0.1886 / 0.3250 & +10.51\% / +10.59\% &$69.25\%$\\
 & f2 & 0.5 & 0.2126 / 0.3629 & +24.56\% / +23.49\% &$41.00\%$\\
 & f3 & 0.8 & 0.2392 / 0.4056 & +40.15\% / +38.05\% &$26.13\%$\\
 & f4 & 1.0 & 0.2558 / 0.4316 & +49.92\% / +46.88\% &$21.22\%$\\
 \midrule
No intervention & base & 0 & 0.1686 / 0.2938 & (base) & (base) \\
\botrule
\end{tabular}
\footnotetext{
    $\sigma$ denotes the standard deviation of the injected Gaussian noise. 
    $r_s$ and $r_f$ denote the ratios of agents whose previews are randomly exchanged or symmetrically flipped, respectively. 
    Results are reported over 5 independent runs. 
    The performance degradations $\Delta\%$ and the group consistency of all variants are compared to the base \NEWMODEL~model, which involves no intervention operation.}
\end{table}

As shown in \TABLE{tab__previews_interventions}, we observe a consistent degradation in prediction performance across all three intervention strategies as the intervention intensity increases.
First, under the noise injection intervention, the model exhibits a certain degree of robustness against slight preview fluctuation, showing merely a $0.43\%/0.06\%$ ADE/FDE degradation when $\sigma=0.1$.
However, as the injected noise becomes more severe, the performance drops significantly.
Second, the structural interventions, \IE, trajectory swapping and flipping, inflict much more severe performance drops compared to the noise injection.
Even when a small proportion of neighbors' previews are exchanged or flipped ($r_s=0.2$ or $r_f=0.2$), the ADE and FDE increase by over $10\%$.
When the previews of all neighbors are entirely replaced or mirrored ($r_s=1.0$ or $r_f=1.0$), the performance drops drastically by up to almost $50\%$.
Furthermore, the newly introduced group consistency metric matches this performance deterioration.
When a slight noise ($\sigma=0.1$) is injected, the group consistency remains almost the same at $89.98\%$, which explains the minimal impact on the final predictions.
However, structural interventions severely disrupt the grouping accuracy.
Specifically, swapping just $20\%$ of the neighbors' previews ($r_s=0.2$) drastically decreases the group consistency to $41.61\%$, while completely exchanging them ($r_s=1.0$) leads to a consistency of merely $0.35\%$.
Similarly, trajectory flipping significantly lowers the consistency down to $21.22\%$, confirming that the correct directionality of previews is also essential for grouping decisions.
These comprehensive results demonstrate that our model explicitly relies on the structural and directional correctness of the short-term predictions.
When these previews are unreliable or completely wrong, the subsequent grouping mechanism is heavily disrupted, which naturally leads to a substantial deterioration in the final trajectory prediction performance.

\section{Further Discussions of Perception Mechanism}

\subsection{The Adaptive FOV Variant}

The different FOV setting experiments in \MODEL~\cite{zou2024who} indicate that agents might be able to capture wider vision range under specific scenes such as competitive games or extreme high-pedestrian-density scenes through frequent head movement. 
In real-world scenes, the ego agent also possesses a zooming capability towards different objects in vision range.
This motivates us to regard introducing such zooming mechanism into social interaction modeling as our further considerations.
Here, we conducted some basic modifications to our base model.
Accordingly, we modify the FOV module in our perception mechanism and carry out the prediction experiments over the original ETH-UCY and SDD datasets, as well as the NBA dataset \cite{linou2016nba}.
The settings of ETH-UCY and SDD datasets are strictly equal to the vanilla \NEWMODEL~model.
As for the NBA dataset, the distributions of NBA dataset are completely different from the pedestrian datasets.
Following the methodology proposed by Xu \ETAL~\cite{xu2022groupnet,xu2022remember}, we set the parameters to be $\left\{t_o, t_p, T\right\} = \left\{5, 10, 0.4\mathrm{s}\right\}$ and $\left\{t_h, t_f\right\} = \left\{3, 2\right\}$.
We randomly select approximately 50000 samples, with 65\% allocated for training, 25\% for testing, and 10\% for validation.

Specifically, we introduce a light-weight embedding network $m_\mathrm{fov}(\cdot)$ to encode target player's surrounding players' relative directions into a normalized FOV angle $\hat{\theta}_{\mathrm{FOV}}$.
Such normalized FOV angle $\hat{\theta}_{\mathrm{FOV}}$ corresponds to real FOV angle from $0^\circ$ to $360^\circ$.
For a target agent $i$, the vector $\mathbf{r}_i$ represents each surrounding player's relative position at the last observation step.
Formally, its normalized adaptive FOV angle is calculated through

\begin{equation}
    \hat{\theta}_{\mathrm{FOV}}^i=m_\mathrm{fov}(\mathbf{r}_i) \in (-1, 1).
\end{equation}
The light-weight embedding network $m_\mathrm{fov}(\cdot)$ comprises a two-layer MLP and its output is constrained to $(-1,1)$ through the final Tanh activation.
Then the real FOV angle is gained through 
\begin{equation}
    \theta_{\mathrm{FOV}}^i=\left(1 + \hat{\theta}_{\mathrm{FOV}}^i\right)\pi \in (0, 2\pi).
\end{equation}

\subsection{Evaluation with NBA Dataset}

The prediction performance of the original \MODEL, \NEWMODEL, and the adaptive FOV variant \NEWMODEL+adaptive FOV on ETH-UCY and SDD datasets is reported in \TABLE{tab_fov_variants}.
We make some further modifications and get some interesting results on NBA dataset, which is reported in \TABLE{tab_nba_comparisons}.
We observe in \TABLE{tab_fov_variants} that equipping the model with the full dynamic FOV zooming capability brings only \emph{marginal} performance improvements on the current pedestrian datasets (ETH-UCY and SDD).
Specifically, although the adaptive FOV variant achieves slightly better ADE and FDE on the eth subset, and FDE on the zara2 subset (with a 2\% improvement on the FDE of zara2), compared to the vanilla \NEWMODEL~in the ETH-UCY dataset, its performance on other subsets is even slightly worse than the original model.
On the SDD dataset, it also underperforms the vanilla \NEWMODEL~by 0.74\%/0.69\% in terms of ADE/FDE.
These \emph{marginal} effects validate our initial settings that in standard pedestrian crowds, which share similar motion patterns with indoor scenes like shopping malls and airports, a fixed $180^\circ$ FOV is an optimal and highly acceptable compromise without over-engineering.

\begin{table}[h]
\footnotesize
\caption{
    Prediction performance evaluated on ETH-UCY and SDD datasets
    }\label{tab_fov_variants}%
\begin{tabular}{@{}l|lllll|l@{}}
\toprule
Models & eth & hotel& univ& zara1& zara2& SDD \\
\midrule
\MODEL&0.254/0.383 &0.103/\C{0.156}&0.256/0.448&0.175/\C{0.284}&0.134/0.225&6.394/10.173\\
\NEWMODEL& 0.233/0.369&\C{0.102}/\C{0.156}&\C{0.240}/\C{0.428}&\C{0.168}/0.294&\C{0.130}/0.223& \C{6.283}/\C{10.121}\\
\NEWMODEL+aF &\C{0.232}/\C{0.365}&0.104/\C{0.156}&0.245/0.435&0.171/0.300&0.132/\C{0.218}& 6.330/10.195\\
\botrule

\end{tabular}
\footnotetext{
    \C{Blue markers} denote the best results on each set.
    \NEWMODEL+aF indicates the \NEWMODEL+adaptive FOV variant.
}
\end{table}

These results suggest that in pedestrian crowds environments (e.g., ETH-UCY and SDD), the reliance on an adaptive FOV might be relatively low, where pedestrians mostly focus on their immediate forward direction.
However, the dynamics in NBA games are notably different due to frequent collaboration and intense competition. 
During a basketball game, each player must constantly notice the positions of all teammates and opponents across the entire court, as well as the ball. 
Under these circumstances, restricting the model with a fixed biological FOV angle might largely limit its prediction capability.

Furthermore, the definition of a \emph{group} in the NBA dataset differs significantly from the social groups (e.g., families or friends walking together) observed in street scenes.
In professional sports, groups are inherently determined by team affiliation. 
Compared to instantly judging whether a near neighbor belongs to the target player's same team with motion cues, game players know well who are their teammates.
The audience can also tell this by observing the jersey colors easily.
Accordingly, we extract deterministic team ownership labels directly from the metadata of the raw NBA dataset to construct a binary teammate group mask, where neighbors belonging to the same team as the target player are considered in-group agents, while opponents and the ball are out-of-group agents. 
The results are shown in \TABLE{tab_nba_comparisons}.

\begin{table}[h]
\caption{
    Comparisons to the state-of-the-art methods on NBA dataset
    }\label{tab_nba_comparisons}%
\begin{tabular}{@{}lll@{}}
\toprule
Models &  $t_p=5$ & $t_p=10$ \\
\midrule

\SOCIALLSTM\SOCIALLSTMCITE~(2016) & 0.88/1.53 & 1.79/3.16 \\
\SOCIALGAN\SOCIALGANCITE~(2018) & 0.85/1.36 & 1.62/2.51 \\
\STGCNN\STGCNNCITE~(2020) & 0.75/0.99 & 1.59/2.37 \\
GroupNet+NMMP\cite{xu2022groupnet}~(2022) & 0.69/1.08 & 1.25/1.80 \\
GroupNet+CVAE\cite{xu2022groupnet}~(2022) & 0.62/0.95 & \C{1.13}/1.69 \\
\SCMODEL\SCCITE~(2024) & 0.67/0.90 & 1.18/\C{1.46} \\

\midrule
\MODEL~\textbf{(Ours)}  &  0.62/0.87  & 1.19/1.58 \\
\NEWMODEL~\textbf{(Ours)} & 0.624/0.875 & 1.212/1.641\\
\NEWMODEL+adaptive FOV \textbf{(Ours)}& 0.618/0.862 & 1.198/1.614\\
\NEWMODEL+team \textbf{(Ours)}& \C{0.605}/\C{0.829} & 1.172/1.521\\

\botrule

\end{tabular}
\footnotetext{
    Metrics are reported in the form of ``ADE/FDE'' (\emph{best-of-20}) in meters.
    Lower metrics indicate better prediction performance.
    \C{Blue markers} denote the best 3 results on each set.
    \NEWMODEL+adaptive FOV replaces fixed FOV angle with adaptive FOV angle embedding.
    \NEWMODEL+team uses team ownership as group priors before the perception mechanism.
}
\end{table}

In \TABLE{tab_nba_comparisons}, our four methods (\MODEL, \NEWMODEL, \NEWMODEL+adaptive~FOV, and \NEWMODEL+team) achieve promising results compared to SOTA methods despite the completely different interaction patterns in the NBA dataset.
These results validate the generalization capability of our approach.
Moreover, comparing \NEWMODEL~and \NEWMODEL+adaptive~FOV shows that the adaptive variant performs better.
This indicates that a fixed FOV angle limits the model's prediction performance on the NBA dataset.
Also, \NEWMODEL+team variant showing the best prediction performance coincidentally validates our basic motivation, which is acquiring the group priors before social interaction modeling (\emph{group-bounded} in the title of this manuscript).

\subsection{Distributions of FOV}

\begin{figure}[h]
\centering
\includegraphics[width=1.0\textwidth]{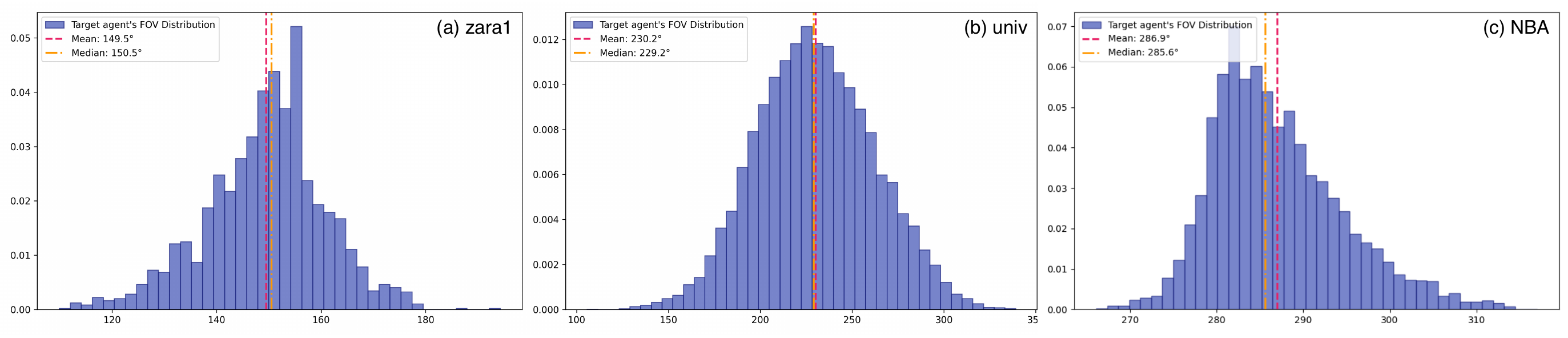}
\caption{
    Visualized distributions of FOV angle $\theta_\mathrm{FOV}$ on different datasets gained through \NEWMODEL+adaptive FOV.
    }\label{fig_nba_fov_distribution}
\end{figure}

We also analyse the adaptive FOV angle $\theta_\mathrm{FOV}$ distributions across different datasets.
The results are visualized in \FIG{fig_nba_fov_distribution}.
For the typical pedestrian scene zara1, the average FOV angle of 149.5$^\circ$ aligns well with the biological human visual field.
Meanwhile, in the univ dataset, the average FOV angle expands to 230.2$^\circ$, suggesting the need for broader perception in denser or more complex environments.
For the highly dynamic NBA scene, the average FOV angle significantly increases to 286.9$^\circ$.
Furthermore, for the NBA dataset, the majority of the samples concentrate within the highly-focused interval of 280$^\circ$ to 295$^\circ$.
It aligns with cognitive patterns in competitive sports that players must maintain wider surrounding perception range.
They might achieve this through more eye movements and more frequent head rotations compared to that of pedestrians.
Furthermore, these experiments above demonstrate that the model adaptively adjusts agents' perception range based on the unique motion distributions of different scenes without any physical hard-coding.

\section{Discussions of Considering Environmental Contexts}

\subsection{Design of \emph{sa}-map Variant}

To investigate the impact of environmental contexts, we follow our previous work SocialCircle+ \cite{wong2024socialcircle+} to bring environmental information into consideration.
Specifically, SocialCircle+ extends the base SocialCircle \cite{wong2023socialcircle} by introducing a conditional branch that extracts a behavior-semantic segmentation map from the scene image.
It treats the pixels in this map labeled from walkable to unwalkable as virtual agents to construct PhysicalCircle meta components, which capture environmental factors such as relative velocity and equivalent distance to obstacles.
These environmental conditions are then adaptively fused with social components to simulate how physical environments affect agents' interactive decisions.
In our proposed \NEWMODEL, the perception mechanism uses group priors as conditions to model the final social interaction representation.
Similarly, we adopt the virtual agent strategy from SocialCircle+ to simultaneously consider the environmental and social interactions.
In SocialCircle+, we conceptually treat the physical environment as a collection of discrete and stationary virtual agents.

To illustrate how the environment comes into play with the grouping mechanisms, we naturally extend our framework by incorporating these virtual agents directly into the \NEWMODEL~framework.
For neighboring agents, \NEWMODEL~divides them into in-group agents and out-of-group agents through \NEWKERNEL, which represents social constraints.
Similarly, we apply the exact same grouping strategy to these virtual agents representing the environmental constraints.
Furthermore, these grouped virtual agents are also jointly processed by the perception mechanism.
The in-group and out-of-group features output by the perception mechanism inherently encode the contextual information of the virtual agents surrounding the target agent.
These features are then fused into a comprehensive representation that simultaneously accounts for social interactions with neighboring agents and environmental interactions with virtual agents.
This fused representation is then fed into the final trajectory prediction backbone to generate future paths.

\begin{figure}[H]
\centering
\includegraphics[width=1.0\textwidth]{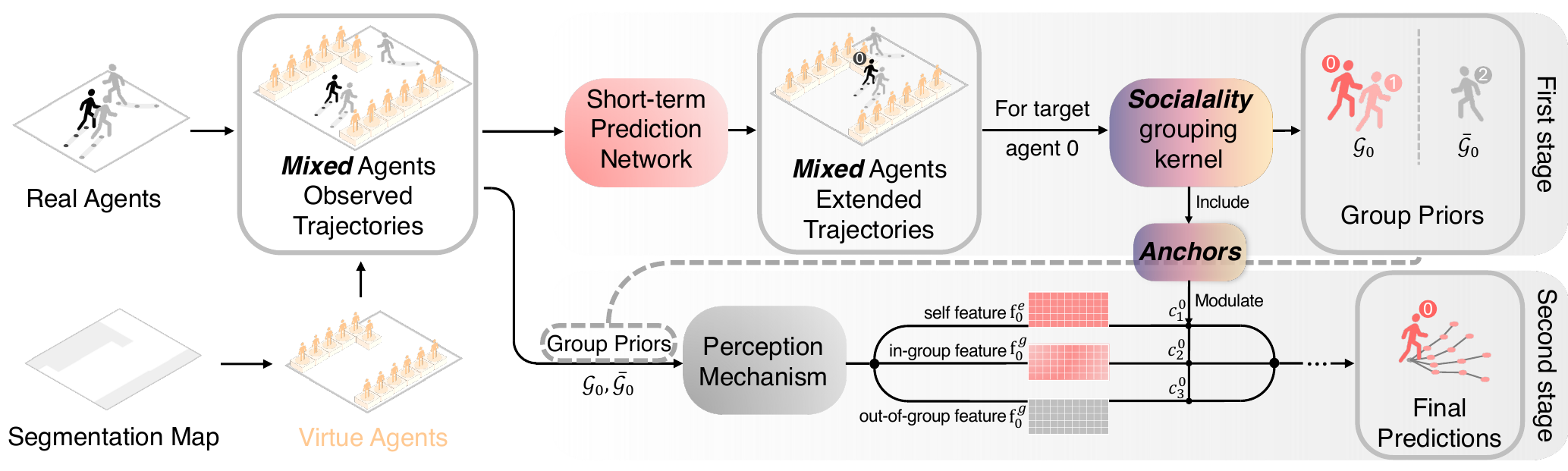}
\caption{
    Method illustration when introducing environmental contexts to the proposed \NEWMODEL~network.
    }\label{fig_segmap_method}
\end{figure}

\begin{figure}[H]
\centering
\includegraphics[width=1.0\textwidth]{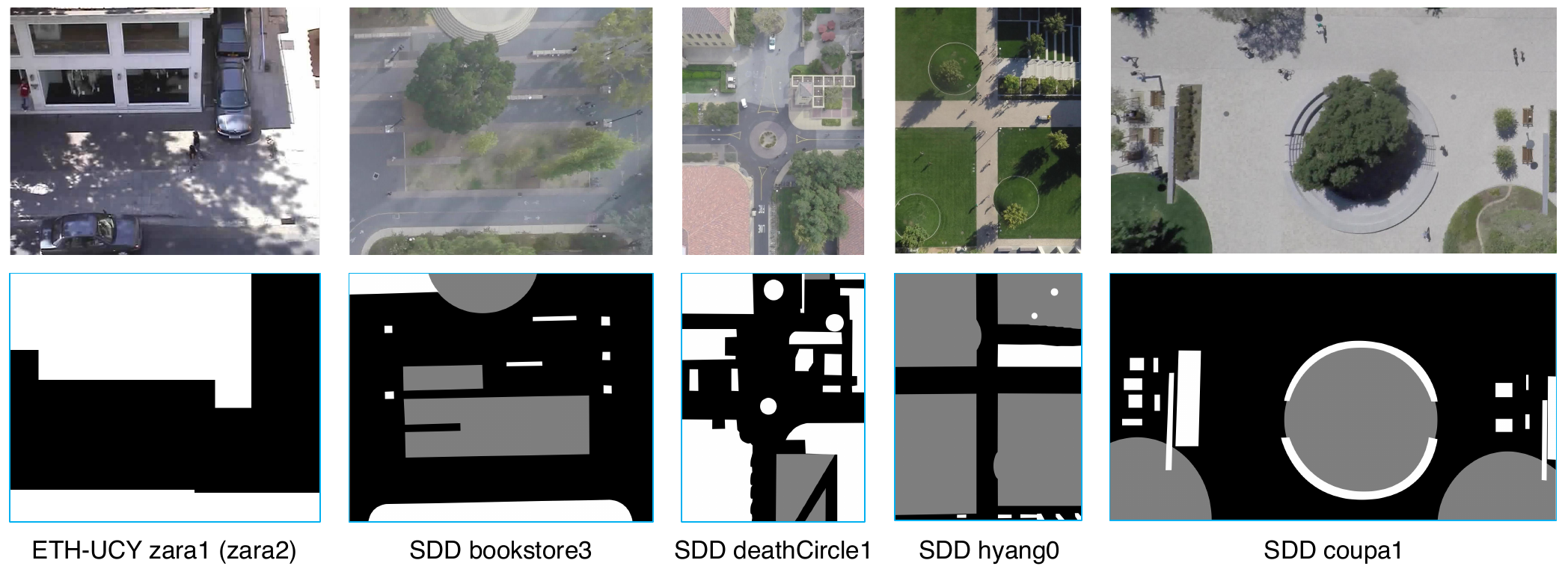}
\caption{
    Some examples of manually labeled segmentation maps on ETH-UCY and SDD datasets.
    }\label{fig_segmap_example}
\end{figure}

Here, we specifically draw a method figure to better illustrate how we introduce the segmentation map and the virtual agents to the \NEWMODEL~in \FIG{fig_segmap_method}.
First, the segmentation map is obtained following the method in SocialCircle+ \cite{wong2024socialcircle+}.
The segmentation map is hand-labeled at the pixel level to indicate the degree of walkability for corresponding regions.
For the ETH-UCY dataset, the scenes are relatively simple and small in scale, allowing us to basically annotate the corresponding regions using just two labels, \IE, walkable or completely non-walkable.
For the SDD dataset, the map range is larger and has more complex environmental conditions.
For example, there are large areas of grass frequently appearing on the campus, where most agents generally will not walk under normal circumstances.
However, through observation of the SDD dataset videos, there are also cases indicating that agents may walk onto the grass to rest, cross the grass to take a shortcut, or briefly step on the grass to avoid other agents.
For regions similar to grass, we make a trade-off and assign them an intermediate value between walkable and non-walkable.
We select several representative segmentation maps that we labeled here in \FIG{fig_segmap_example}.
The complete segmentation maps and related detailed settings can be found in our previous paper SocialCircle+ \cite{wong2024socialcircle+}.

In both the original base \NEWMODEL~model and the current \emph{sa}-map variant, the perception mechanism uses group priors as conditions to perceive interactions with other agents, including real agents and virtual agents.
Specifically, the coordinates of the pixels from the segmentation map are directly treated as the positions of the virtual agents.
Since these virtual agents represent stationary environmental objects, their velocities are set to zero.
Then, the proposed \NEWKERNEL~evaluates both the spatial distance and the relative velocity between the target agent and these virtual agents.
In this way, the grouping mechanism seamlessly integrates the physical environment by treating it as a special type of social interaction.
Comparing \FIG{fig_segmap_method} with the method figure in the Method section, we can observe that the overall framework of the current \emph{sa}-map model variant is the same as \NEWMODEL.
The only difference is the input part.
The \emph{sa}-map variant uses mixed agents observed trajectories as input, while base \NEWMODEL~model receives only trajectories from real agents.
It should be noted that the mixed trajectories are also extended through the short-term prediction network.

\begin{table}[h]
\footnotesize
\caption{
    Prediction performance of \emph{sa}-map model evaluated on ETH-UCY and SDD datasets
    }\label{tab_samap_performance}%
\setlength{\tabcolsep}{5pt}
\begin{tabular}{@{}l|c|lllll|l@{}}
\toprule
Models & $s$ &eth & hotel& univ& zara1& zara2& SDD \\
\midrule
\emph{sa}-map & 1 &0.232/0.361&0.109/0.173&0.248/0.446&0.183/0.309&0.140/0.241&6.354/10.176\\
\emph{sa}-map & 2 &0.238/0.375&\C{0.115}/0.168&0.244/0.439&0.186/0.319&0.139/0.245&6.310/10.106\\
\emph{sa}-map & 5 &\C{0.230}/\C{0.360}&0.104/0.155&\C{0.237}/\C{0.416}&0.170/\C{0.294}&\C{0.130}/\C{0.218}&\C{6.274}/\C{10.074}\\
\emph{sa}-map & 10 &0.234.0.369&0.107/0.160&0.239/0.430&0.181/0.304&0.134/0.233&6.303/10.102\\
\midrule
\NEWMODEL& - &0.233/0.369&\C{0.102}/0.156&0.240/0.428&\C{0.168}/\C{0.294}&\C{0.130}/0.223& 6.283/10.121\\
\botrule

\end{tabular}
\footnotetext{
    Stride value $s$ indicates the pooling stride used when reducing virtual agent density.
    Larger stride means fewer virtual agents.
    The default maximum virtual agents in a scene is set to $100 \times  100$.
    When stride $s$ is set to 5, it means the maximum virtual agents are $20 \times 20$.
    \C{Blue markers} denote the best results on each set.
}
\end{table}

\subsection{Quantitative Analyses}
\begin{figure}[h]
\centering
\includegraphics[width=1.0\textwidth]{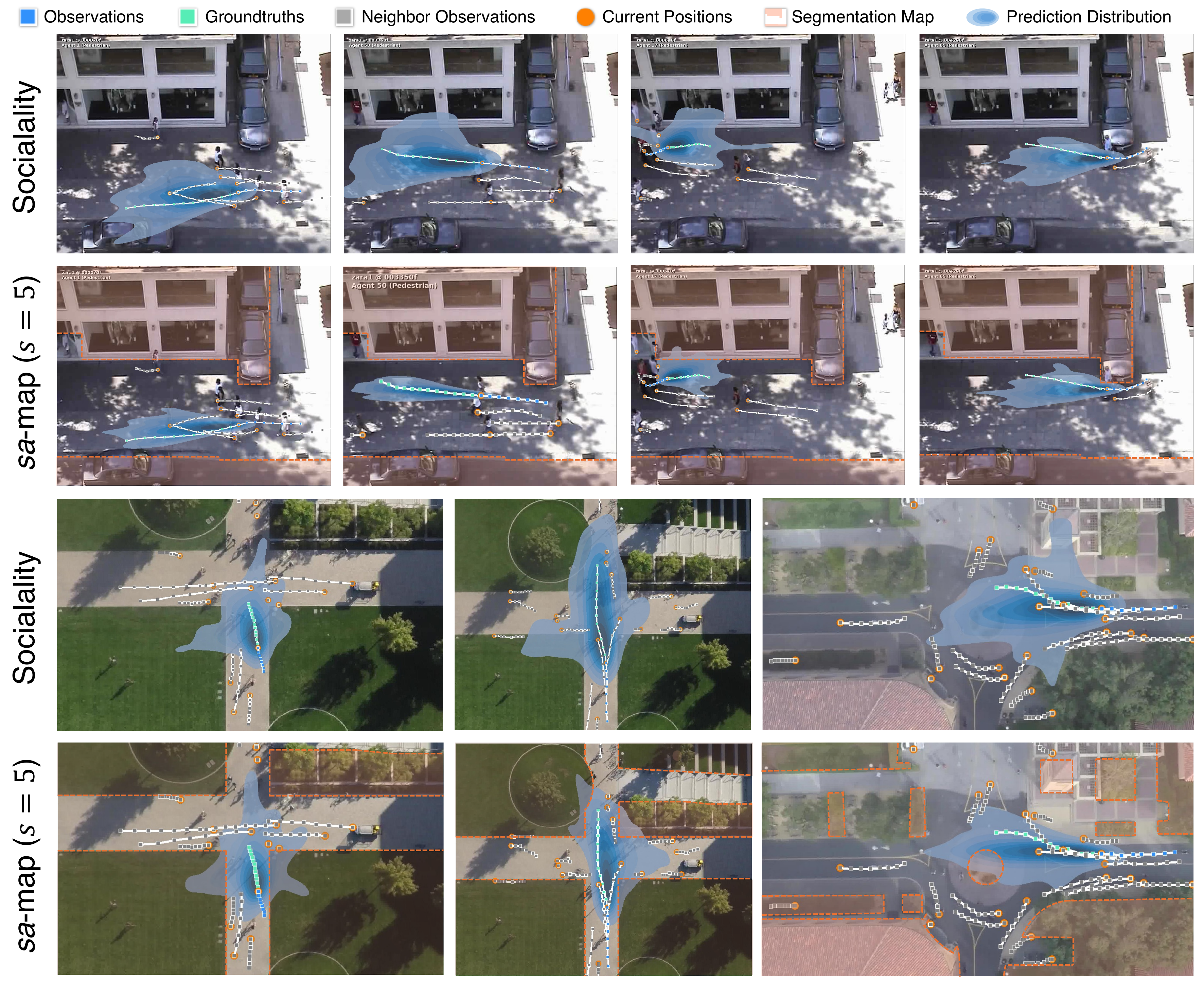}
\caption{
    Distributions of predicted trajectories provided by the base \NEWMODEL~and the \emph{sa}-map variant on the ETH-UCY zara1, SDD hyang0, and deathCircle2 scenes.
    Segmentation regions are noted by semi-transparent orange color on the figure.
    }\label{fig_segmap_predictions}
\end{figure}

We evaluated the prediction performance of the \emph{sa}-map model variant on the ETH-UCY and SDD datasets, and the results are presented in \TABLE{tab_samap_performance}. 
It should be noted that the \emph{sa}-map variant has only 3.2\% more parameters than the original \NEWMODEL.
As shown in \TABLE{tab_samap_performance}, the performance of the \emph{sa}-map variant varies with the pooling stride $s$ which controls the virtual agent density. 
Setting a moderate stride of $s=5$ achieves the overall best balance between environmental cues and social interactions, outperforming the baseline \NEWMODEL~across most datasets. 
This demonstrates that incorporating environmental contexts into our grouping mechanism can effectively improve prediction accuracy.
To further analyse the grouping mechanism of current \emph{sa}-map model variant, we visualize several prediction and grouping examples comparing \NEWMODEL~with the current \emph{sa}-map model variant in \FIG{fig_segmap_predictions} and \FIG{fig_segmap_grouping}.
Here we choose $s=5$ as the stride for \emph{sa}-map variant since this setting leads to best prediction performance.

\subsection{Qualitative Analyses}

\NEWHALFLINE\noindent\textbf{\emph{1) Overall Prediction:}} We can observe in \FIG{fig_segmap_predictions} that the final predictions of the base model \NEWMODEL~and the \emph{sa}-map variant are significantly different.
On the one hand, comparing (a1) and (a2) with (b1) and (b2) reveals that the distributions of the predicted trajectories from the \emph{sa}-map variant are more concentrated, attempting to constrain the trajectories within walkable regions.
Without the constraints of the segmentation map, the predicted trajectories of the \NEWMODEL~clearly exhibit a more diverse distribution.
This also applies to relatively slow-moving agents, as (a3) and (b3) demonstrate different prediction patterns for the future movement trends of a nearly stationary target agent with higher uncertainty.
On the other hand, comparing (a4) and (b4) shows that the \emph{sa}-map variant might not achieve this simply by imposing greater restrictions on the overall prediction distributions of the model, since the prediction distributions in (a4) and (b4) are relatively similar.
It indicates that the model adaptively chooses whether to apply environmental constraints to the current target agent.
The same phenomenon can be observed in the SDD dataset.
For example, in the SDD hyang0 intersection scene shown in (a5), (a6) and (b5), (b6), the \emph{sa}-map variant outputs trajectories that are more aligned with map restrictions compared to those of the base \NEWMODEL.
This is also observed in (a7) and (a8), where the \emph{sa}-map variant predicts that the target agent tends to avoid collisions with the building, while the \NEWMODEL~provides much more diverse predictions.

Accordingly, the quantitative and the qualitative analyses above validate that the \emph{sa}-map variant has successfully learned the impact of environmental constraints on the predicted trajectories.
Given the segmentation map, the \emph{sa}-map variant can output predicted trajectories that better align with the environmental context conditions.
Here, we would like further analyze how grouping mechanism and group behavior will change after introducing environmental contexts as virtual agents.

\begin{figure}[h]
\centering
\includegraphics[width=1.0\textwidth]{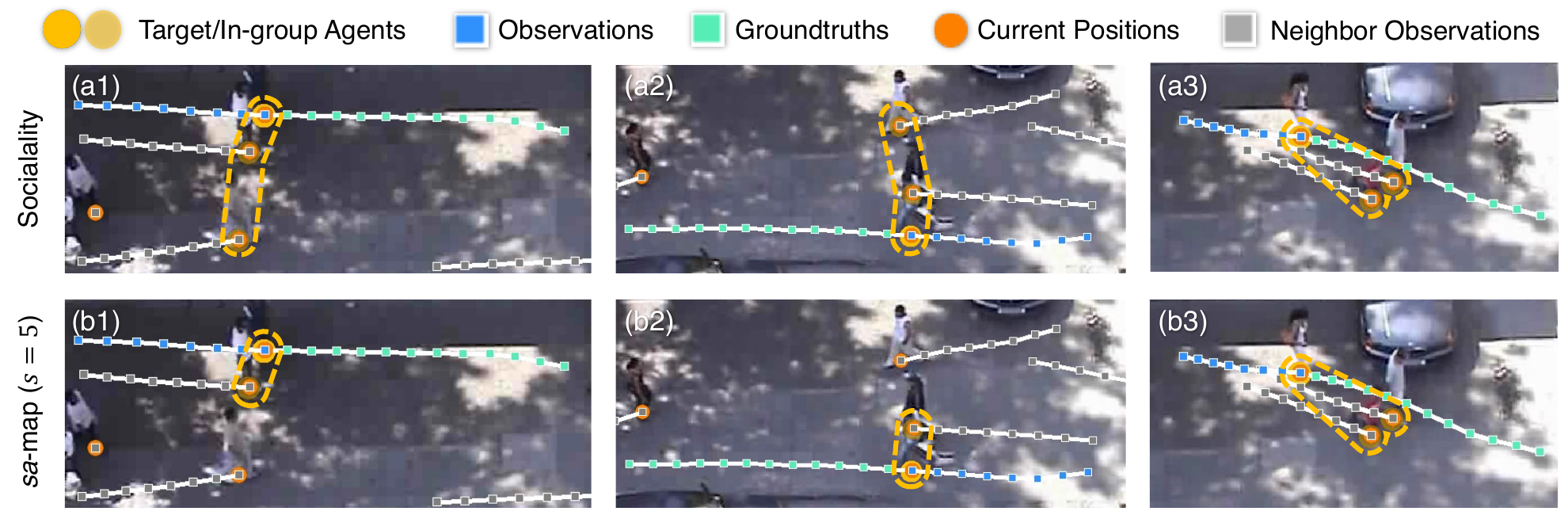}
\caption{
    Examples of different group priors provided by the base \NEWMODEL~and the \emph{sa}-map variants.
    }\label{fig_segmap_group}
\end{figure}

\begin{figure}[h]
\centering
\includegraphics[width=1.0\textwidth]{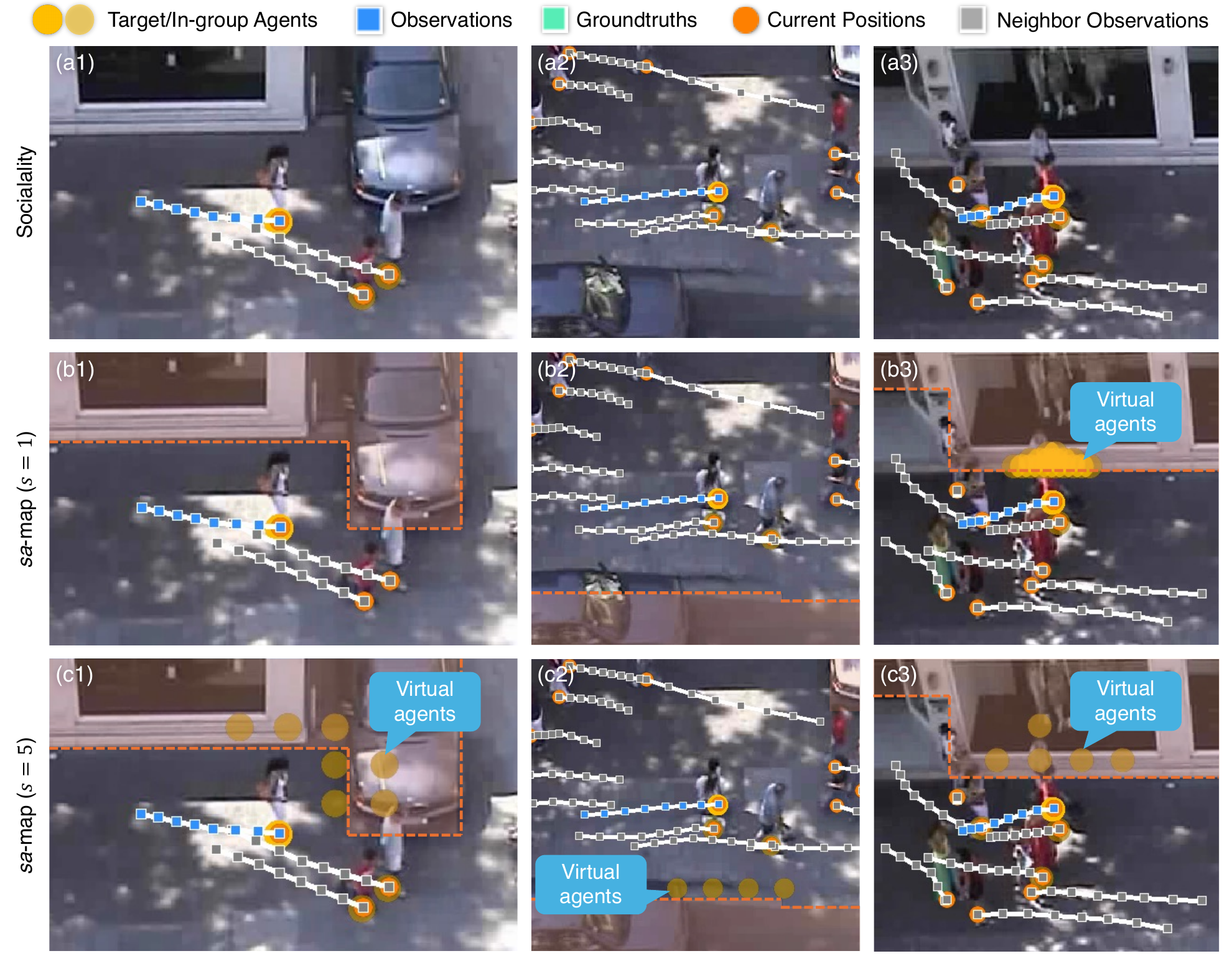}
\caption{
    Examples of different group priors (including virtual agents) provided by the base \NEWMODEL~and the \emph{sa}-map variants under different stride value.
    }\label{fig_segmap_grouping}
\end{figure}

\NEWHALFLINE\noindent\textbf{\emph{2) Grouping Analyses:}}
In \FIG{fig_segmap_grouping}, we visualize several examples to investigate whether the grouping results of the \emph{sa}-map variant differ after introducing virtual agents.
First, we can observe that the base \NEWMODEL~and the \emph{sa}-map ($s=5$) variant can generally achieve reasonable grouping of real agents.
For the case in (a3) and (b3), both models provide the same group priors to include two neighbors walking in front of the target agent\footnote{This is a classic case we discussed in our original \MODEL~\cite{zou2024who} in detail.}.
However, compared with (b1)-(b3), \FIG{fig_segmap_grouping} (a1)-(a3) exhibits a more inclusive grouping preference.
First, in (a1) and (b1), we can observe that the base \NEWMODEL~includes an approaching agent, while the \emph{sa}-map variant decides that a two-agent group is sufficient.
This phenomenon can also be observed in (a2) and (b2).
The different grouping preferences may reflect an inherent model tendency.
After introducing environmental contexts as virtual agents, the model is more inclined to shrink its group boundary, thus excluding some neighbors that would have originally been divided into the in-group agent set.
We would like to further analyze the reason behind this phenomenon.

Since the only difference between the base \NEWMODEL~and the \emph{sa}-map variant is that the latter utilizes virtual agents and treats them as real agents, we visualize several examples to investigate how these virtual agents change the model's grouping decisions in \FIG{fig_segmap_grouping}.
Moreover, we add a \emph{sa}-map variant under $s=1$ to provide more in-depth analyses of virtual agents.
By comparing (b1) and (c1), we find that \emph{sa}-map ($s=5$) grouped some virtual agents around the target agent, while \emph{sa}-map ($s=1$) chooses to consider the target agent as individually moving.
The \emph{sa}-map ($s=1$) variant exhibits an even more exclusive tendency compared to \emph{sa}-map ($s=5$).
Example in (b2) and (c2) also indicates that \emph{sa}-map ($s=5$) determines some in-group virtual agents, whereas in (b2), \emph{sa}-map ($s=1$) decides that no virtual agent should be included as an in-group agent.
Further comparing (b3) and (c3) reveals that \emph{sa}-map ($s=1$) might cluster a massive number of virtual agents into the group for a target agent walking on the edge of non-walkable regions.
In contrast, \emph{sa}-map ($s=5$) conducts the grouping for a much larger non-walkable region by incorporating only 5 virtual agents.
This validates the necessity of introducing the virtual agents max-pooling strategy.
Under the default setting, the number of virtual agents is far greater than the number of real agents in the scene.
If there is a non-walkable region around the target agent, the number of virtual agents will also be excessively large.
However, the entire \NEWMODEL~framework equally considers real and virtual agents.
This forces the model to decrease the probability of the target agent including neighbors, thus avoiding the inclusion of too many virtual agents that are useless for the final prediction.
Conversely, with a reasonable stride value, the number of virtual agents in the scene is moderate, allowing them to jointly participate with real agents as group priors of the target agent, thereby achieving better trajectory prediction performance.

\begin{figure}[h]
\centering
\includegraphics[width=1.0\textwidth]{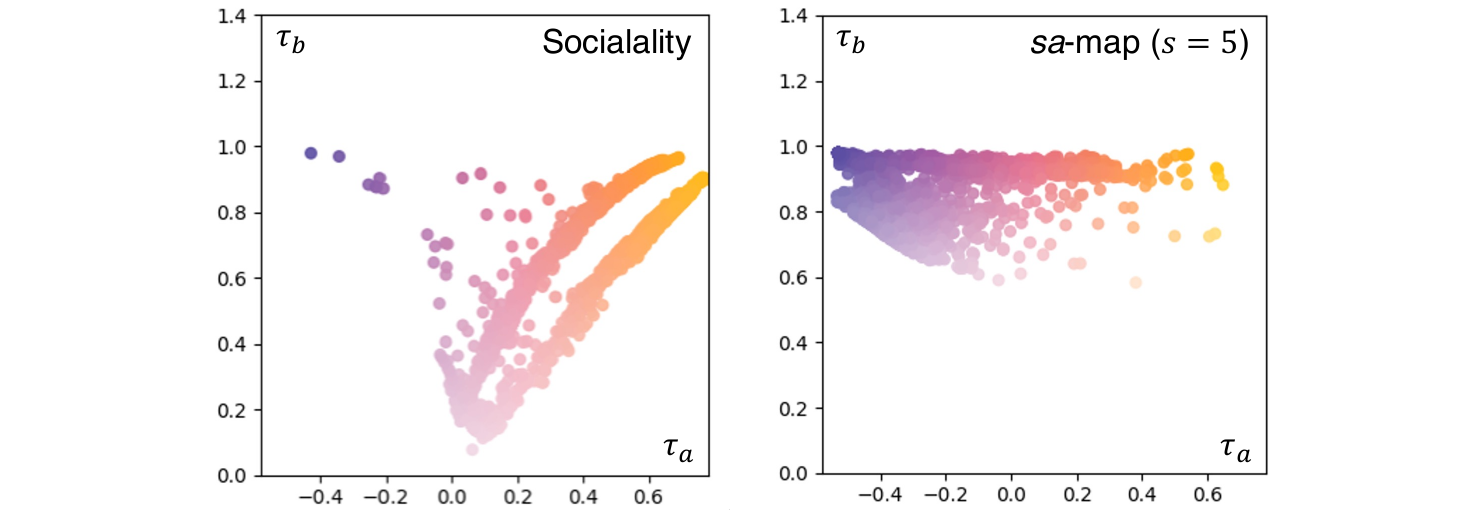}
\caption{
    Visualization of \ANCHOR~anchor distributions provided by the base \NEWMODEL~and the \emph{sa}-map variant.
    }\label{fig_segmap_anchor}
\end{figure}

\NEWHALFLINE\noindent\textbf{\emph{3) Anchor Analyses:}}
We are also wondering whether the \ANCHOR~anchors would change after introducing environmental contexts as virtual agents.
In \FIG{fig_segmap_anchor}, we can observe that the anchor distribution in both base model and the \emph{sa}-map ($s=5$) variant exhibit a distinct ``V''-shape structure.
However, the direction and distribution of the ``V''-shape structure have changed in the \emph{sa}-map ($s=5$) variant.
In our manuscript, we analyzed the relationship between the agent-specific anchor and the grouping rule from agent-level and group level perspective.
Meanwhile, we analyzed the anchor distributions across different dataset scenes and the corresponding agent behavior patterns in each region.
Finally, we demonstrated why more anchors are not needed from the statistical analysis.
We can observe in \FIG{fig_segmap_anchor} that the anchor distribution of the \emph{sa}-map ($s=5$) variant is more scattered, with many agents' anchor values falling into the middle region of the ``V''-shape.
Moreover, a large proportion of the anchor distribution in \emph{sa}-map ($s=5$) is concentrated in regions with a large speed tolerance $\tau_b$, while the distance tolerance $\tau_a$ is evenly distributed.
This corresponds to the situation where the target agent requires a larger $\tau_b$ to include stationary virtual agents as in-group agents.
It indicates that the model can adaptively adjust the anchor distribution and learning pattern to determine how to group neighboring agents more effectively.
Here, the anchor visualization of the \emph{sa}-map ($s=5$) variant further validates the effectiveness of our proposed \ANCHOR~anchors in representing the grouping preferences of agents.

\section{Higher-degree Group Members Analyses}

\subsection{Definition of Higher-degree Group Members}

We conducted two analyses on the higher degree members among the target agent's neighbors. 
We have cited a representative method in our original manuscript studying higher-order social relations \cite{kim2024higher}, which uses HighGraph to capture the higher-order dynamics of social interactions.
Similarly, to explicitly capture this chain effect of grouping relations without relying on graph, we follow how Kim \ETAL~\cite{kim2024higher} construct HighGraph to adopt a recursive approach to construct higher degree group members.
Specifically, for a target agent $i$, there might exist an in-group set containing its in-group neighbors, which we refer to as the first-degree group set, denoted as $\mathcal{G}^{(1)}_i$. 
Subsequently, each agent $j$ within this first-degree group set ($j \in \mathcal{G}^{(1)}_i$) possesses its own first-degree group set $\mathcal{G}^{(1)}_j$. 
The union of these individual sets is defined as the second-degree group set $\mathcal{G}^{(2)}_i = \bigcup_{j \in \mathcal{G}^{(1)}_i} \mathcal{G}^{(1)}_j $.
The third-degree group set is then defined as $\mathcal{G}^{(3)}_i = \bigcup_{j \in \mathcal{G}^{(2)}_i} \mathcal{G}^{(2)}_j$.
Following the same logic, the $p$-degree group set can be calculated from the $(p-1)$-degree group set:
\begin{equation}
    \mathcal{G}^{(p)}_i = \bigcup_{j \in \mathcal{G}^{(p-1)}_i} \mathcal{G}^{(1)}_j, p \geq 1.
\end{equation}
Specifically, when $p=1$, $\mathcal{G}^{(1)}_i = \bigcup_{j \in \mathcal{G}^{(0)}_i} \mathcal{G}^{(0)}_j$, which indicates that the zero-degree group set $\mathcal{G}^{(0)}_i$ is the target agent itself, \IE, $\mathcal{G}^{(0)}_i=\{i\}$.

\begin{figure}[h]
\centering
\includegraphics[width=1.0\textwidth]{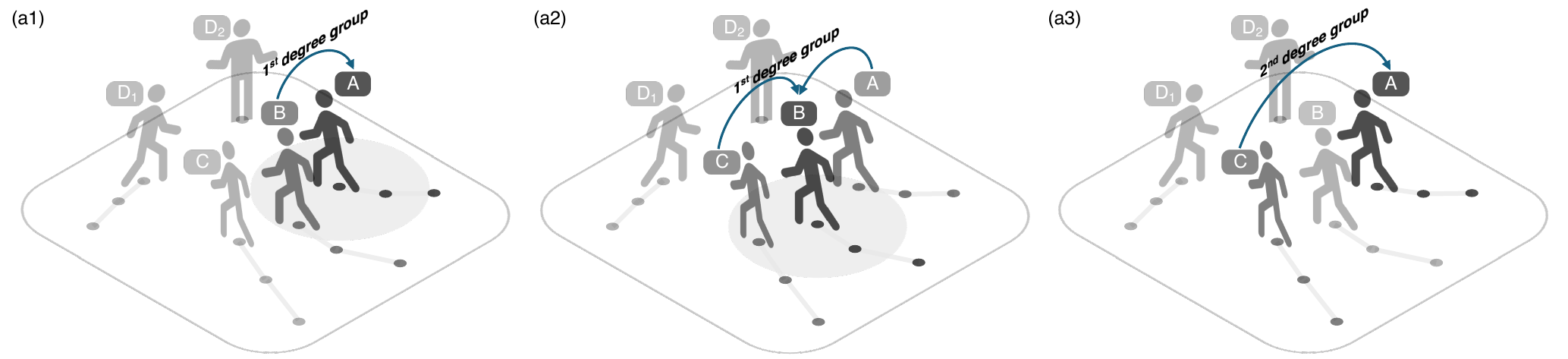}
\caption{
    Illustration of first-degree group neighbors and higher-degree group neighbors.
    }\label{fig_higher_degree_group}
\end{figure}

To illustrate the relation of these neighbors more directly, we visualize an example of second-degree group members.
As shown in \FIG{fig_higher_degree_group}, (a1) represents that agent B is target agent A's in-group neighbor.
(a2) shifts the target agent to B, where agent A and agent C are both agent B's in-group agents.
In this way, agent C is \emph{connected} with agent A via agent B, thus making itself agent A's second-degree group member.

In our \NEWMODEL, we apply two different interaction modeling strategies on in-group and out-of-group agents, as shown in original manuscript Sec. 3.3 (Perception Mechanism).
To take higher-degree group members into account, we modify the in-group agents with different degree group members.
The original in-group agents means considering up to the first-degree group members, which contains both the target agent $\{i\}=\mathcal{G}^{(0)}_i$ itself and the first degree group set $\mathcal{G}^{(1)}_i$.
Specifically, when taking second-order group members into account, the in-group members become the combination of original in-group agents set $\{i\} \cup \mathcal{G}^{(1)}_i$ and the second-degree group set $\mathcal{G}^{(2)}_i$, \IE, $\{i\} \cup \mathcal{G}^{(1)}_i \cup \mathcal{G}^{(2)}_i$.
Following the same reasoning, considering up to $q$-degree group members means considering all correspond degree group members set $\mathcal{U}_{i}^{q}$, which can be formulated as:
\begin{equation} 
    \mathcal{U}_{i}^{q} = \bigcup_{p=0}^{q} \mathcal{G}^{(p)}_i 
\end{equation}

\subsection{Different Degree Group Members Feature Energy}

To calculate the impact of introducing these higher-degree group members to our network, we treat the modification of the in-group set as a structural intervention. 
Specifically, we perform the intervention operation $do(\mathcal{G}_i = \mathcal{U}_i^q)$ to represent considering up to $q$-degree group members. 
Accordingly, the remaining neighbors set becomes $\mathcal{N}_i \setminus \mathcal{U}_i^q$, which are regarded as the out-of-group agents to be calculated through the perception mechanism.
In the original manuscript, the in-group agents finally form the in-group feature $\mathbf{f}_i^g$, which is modulated by the \ANCHOR~anchors and fused through an fusion encoder $n(\cdot)$.
Under $do(\mathcal{G}_i = \mathcal{U}_i^q)$ intervention, we calculate the feature energy of the modulated in-group feature, defined as the squared sum of the intervened feature tensors prior to the final fusion layer.
Formally, 
\begin{equation}
E_i^g(q) = \left( \Vert c_2 \mathbf{f}_i^g(do(\mathcal{G}_i = \mathcal{U}_i^q))\Vert_2 \right)^2.
\end{equation}

\begin{table}[h]
\caption{
Feature energy for up to $q$-degree group members set $\mathcal{U}_i^q$ evaluated on ETH-UCY zara1 across 3 trained weights and SDD datasets 
}\label{tab_feature_energy}%
\begin{tabular}{@{}l|c|lll|l@{}}
\toprule
$E_i^g(q)$ & $do(\mathcal{G}_i = \mathcal{U}_i^q)$&\#$1_\mathrm{zara1}$ & \#$2_\mathrm{zara1}$& \#$3_\mathrm{zara1}$&  \#$1_\mathrm{sdd}$\\
\midrule
$q=1$&$\mathcal{U}_i^1$&41247(+0.0\%)&51662(+0.0\%)&55459(+0.0\%)&239994(+0.0\%)\\
$q=2$&$\mathcal{U}_i^2$&52818(+28.1\%)&67837(+31.3\%)&68783(+24.0\%)&318733(+32.8\%)\\
$q=3$&$\mathcal{U}_i^3$&53527(+29.8\%)&69668(+34.9\%)&74646(+34.6\%)&329013(+37.1\%)\\
\botrule
\end{tabular}
\end{table}

We can observe from \TABLE{tab_feature_energy} that the feature energy $E_i^g(q)$ increases as we expand the degree $q$. 
Specifically, when extending the group from first-degree ($q=1$) to second-degree ($q=2$), the in-group feature energy experiences a notable increase across all models (e.g., $+28.1\%$ in $\#1_\mathrm{zara1}$ and $+32.8\%$ in $\#1_\mathrm{sdd}$). 
However, when we further intervene the in-group agents set to third-degree members ($q=3$), the gain of feature energy becomes marginal. 
For example, in $\#1_\mathrm{zara1}$, the energy only marginally increases from $+28.1\%$ to $+29.8\%$ ($1.7\%$). 
This phenomenon indicates that the second-degree group members might indeed need consideration, while the third or higher-degree group members are relatively marginal to the target agent.

\subsection{Different Degree Group Members Feature Sensitivity}

To validate whether the \NEWMODEL~captures and utilizes information from these exact higher-degree group members, we further analyse the final fused representation $\mathbf{f}_i$ in Eq. (25) before trajectory prediction backbones. 
For a target agent $i$, $\mathbf{X}_k(t_o, 0) \in \mathbb{R}^{t_o \times 2}$ denote the historical input trajectory of a neighboring agent $k$. 
We define the feature sensitivity $\Delta E_{i \leftarrow k}$ exerted by agent $k$ on target $i$ using the feature energy change.
Formally, 
\begin{equation}
    \Delta E_{i \leftarrow k} = \left| E_i(\Delta \mathbf{X}_k) \right|=\Vert \mathbf{f}_i(\Delta \mathbf{X}_k) \Vert_2^2.
\end{equation}
By computing this feature sensitivity over all agents $j \in \mathcal{N}_i$ and aggregating each belonging to different $q$-degree group members set $\mathcal{U}_i^q$, we can quantify the \NEWMODEL's feature energy allocation ratio towards different degree group members.

\begin{table}
    \caption{
    Feature energy allocation for different degree group members sets evaluated on ETH-UCY zara1 across 3 independent training runs and SDD datasets
    }\label{tab_feature_sensitivity}%
\begin{tabular}{@{}l|lll|l@{}}
\toprule
$\mathcal{U}_i^q$ & \#$1_\mathrm{zara1}$ & \#$2_\mathrm{zara1}$& \#$3_\mathrm{zara1}$&  \#$1_\mathrm{sdd}$\\
\midrule
$q=1$&99.89\%&70.58\%&99.39\%&36.04\%\\
$q=2$&99.89\%&98.45\%&99.99\%&58.44\%\\
$q=3$&99.89\%&100.00\%&100.00\%&73.37\%\\
\botrule
\end{tabular}
\end{table}

As shown in Table \ref{tab_feature_sensitivity}, we can observe that our proposed \NEWMODEL~consistently allocates a major share of feature energy to the first-degree group members ($q=1$, \IE, $\mathcal{U}_i^1=\{i\}\cup\mathcal{G}^{(1)}_i$) in most runs corresponding to zara1.
Although the in-group feature increases considerably when further including the second-degree group members, as shown in \TABLE{tab_feature_energy} and \TABLE{tab_feature_sensitivity}, the model allocates little concentration on them since the fused feature sensitivity does not increase accordingly from $q=1$ to $q=2$.
However, in SDD dataset, the model implicitly learns to allocate an higher energy sensitivity to higher-degree group members. 
This indicates that the perception mechanism could adaptively adjusts energy allocation with higher proportion of higher-degree group members and the remaining neighbors according to different scenes and distributions.

\section{Prediction Backbone Verifications}

The proposed \NEWMODEL~first obtains group priors via \NEWKERNEL~and utilizes them as conditions to derive the social interaction representation. 
Finally, a trajectory prediction backbone is employed to output the predicted trajectories based on the final social interaction representation. 
To further verify the generalization capability of \NEWMODEL, we replace the original MSN backbone with different backbones that are capable of modeling social interactions, and evaluate them on the ETH-UCY dataset to verify whether the prediction performance of \NEWMODEL~can be maintained across various backbones.
Specifically, we would like to briefly introduce the selected backbones below.
It should be noted that all backbones generate $K_f=20$ trajectories.

\NEWHALFLINE\noindent\textbf{MSN} \cite{wong2021msn} is a Transformer-based multi-style trajectory prediction network. 
To build our base \NEWMODEL~model (\IE, \emph{sa}-MSN), the original social-interaction modules in MSN are removed and replaced by our group-conditioned representations.

\NEWHALFLINE\noindent\textbf{Transformer} \cite{vaswani2017attention} is the simplest Transformer model used to predict trajectories.
Since MSN is a Transformer-based framework, we remove the MSN-related components and only keep the transformer part.
Notably, this is the only backbone variant that requires fewer parameters than the base \NEWMODEL.

\NEWHALFLINE\noindent\textbf{E-$\mathrm{V}^2$-Net} \cite{wong2023another} introduces the Fourier transform to trajectory prediction.
Similarly, its default social-interaction modules are removed when constructing the \emph{sa}-ev variant.

\NEWHALFLINE\noindent\textbf{Resonance} \cite{wong2024resonance} is motivated by the co-vibration phenomenon and forecasts trajectories as the superposition of independent vibrations separately.
Here, we replace the social-interaction part of Resonance with our final social interaction representation to compute the agents' reactions to social vibrations.
Since each vibration reaction is calculated independently, the \emph{sa}-re variant contains the highest number of parameters among all tested models.

\begin{table}[h]
\footnotesize
\caption{
    Prediction performance of the proposed \NEWMODEL~Model modified with different backbone prediction models on ETH-UCY dataset
    }\label{tab_backbone_models}%
\setlength{\tabcolsep}{6pt}
\begin{tabular}{@{}l|c|ccccc@{}}
\toprule
Model&Backbone & eth & hotel& univ& zara1& zara2 \\
\midrule
\emph{sa}-tran&Transformer \cite{vaswani2017attention}&0.272/0.454&0.128/0.212&0.277/0.516&0.174/0.306&0.137/0.230\\
\emph{sa}-ev&E-$\mathrm{V}^2$-Net \cite{wong2023another}&0.234/0.369&0.111/0.171&0.246/0.437&0.177/0.311&0.137/0.236\\
\emph{sa}-re&Resonance \cite{wong2024resonance}&\C{0.227}/\C{0.356}&0.106/0.169&\C{0.234}/\C{0.425}&0.180/0.303&\C{0.130}/\C{0.220}\\
\midrule
base&MSN \cite{wong2021msn} &0.233/0.369&\C{0.102}/\C{0.156}&0.240/0.428&\C{0.168}/\C{0.294}&\C{0.130}/0.223  \\
\botrule
\end{tabular}
\footnotetext{
    The base model, \IE, \NEWMODEL~uses MSN \cite{wong2021msn} as the prediction backbone, which includes a transformer network.
    \emph{sa}-tran only uses transformer \cite{vaswani2017attention} as the prediction backbone, removing the MSN components.
    \emph{sa}-ev uses E-$\mathrm{V}^2$-net \cite{wong2023another} as the prediction backbone.
    \emph{sa}-re uses Resonance \cite{wong2024resonance} as the prediction backbone.
    \C{Blue markers} denote the best results on each set.
    }
\end{table}

\begin{table}[h]
\footnotesize
\caption{
    Parameter counts of the proposed \NEWMODEL~Model modified with different backbone prediction models
    }\label{tab_backbone_models_parameters}%
\setlength{\tabcolsep}{6pt}
\begin{tabular}{@{}l|l|c@{}}
\toprule
Model&Backbone & Parameters \\
\midrule
\emph{sa}-tran&Transformer \cite{vaswani2017attention}&2,041,413 (-1.8\%) \\
\emph{sa}-ev&E-$\mathrm{V}^2$-Net \cite{wong2023another}&3,601,093 (+73.2\%)\\
\emph{sa}-re&Resonance \cite{wong2024resonance}&3,996,357 (+92.2\%)\\
\midrule
base&MSN \cite{wong2021msn} & 2,079,577 \\
\botrule
\end{tabular}
\end{table}

In \TABLE{tab_backbone_models}, we can observe that all three model variants achieve prediction performance comparable to the base \NEWMODEL~on the ETH-UCY dataset. 
Notably, \emph{sa}-tran still obtains considerable prediction performance despite removing the MSN-related modules and reducing the overall parameter count. 
Compared to the base model, the performance gap between \emph{sa}-tran and the base model is minimal on the zara1 subset. 
Furthermore, \emph{sa}-re achieves the best prediction accuracy, outperforming the base model in terms of ADE/FDE on the eth, univ, and zara2 datasets. 
These comprehensive evaluations robustly demonstrate that the performance gains of our framework are not tightly coupled to a specific backbone (such as MSN). 
Instead, the group-bounded representations generated by our \NEWMODEL~serve as priors that can effectively enhance diverse trajectory prediction architectures.

\section{Dataset Analyses}

First, we would like to analyse the distributions of pairwise distances between agents across different subsets.
Considering the ETH-UCY dataset lacks explicit ground-truth group annotations (SDD dataset does not have such annotation, either.), we further analyse the nearest-neighbor distance distribution across all subsets.
The nearest-neighbor distance indicates the shortest distance during all observation steps between any neighbor and the target agent.

\begin{table}[h]
\caption{
    Statistics for pairwise and nearest neighbor distances across ETH-UCY subsets (in meters)
    }\label{tab_pairwise_distance}%
\begin{tabular}{@{}l|ccccc@{}}
\toprule
Statistic & eth & hotel& univ& zara1& zara2 \\
\midrule

Mean (p)& 5.738 & 4.667 & 6.555 & 5.215 & 4.946 \\
Median (p)& 4.747 & 4.105 & 6.353 & 4.503 & 4.336 \\
Mean (n)& 1.881 & 1.652 & 0.803 & 1.571 & 1.231 \\
Median (n)& 1.098 & 1.184 & 0.632 & 0.911 & 0.756 \\

\botrule

\end{tabular}
\footnotetext{
    Mean (p) and Median (p) indicate pairwise mean distance and pairwise median distance.
    Mean (n) and Median (n) indicate nearest-neighbor mean distance and nearest-neighbor median distance.
    }
\end{table}

As presented in \TABLE{tab_pairwise_distance}, we can observe that the nearest-neighbor (n) distances significantly shift across different subsets. 
For example, the median nearest-neighbor distance in the hotel and eth subsets is noticeably larger than in univ and zara2. 
Notably, the median nearest-neighbor distance of the hotel scene(1.184) is almost twice as large as that of the univ (0.632).
This indicates that the acceptable social distance is highly sensitive to the specific scene and crowd density.
The univ dataset exhibits the largest pairwise median distance, implying a larger scene where agents are widely spread out globally. 
However, it presents the smallest nearest-neighbor median distance. 
This demonstrates that agents in univ move in small groups despite the vast global space.
Even within the exact same physical location, \IE, zara1 and zara2, the median nearest-neighbor distances are distinctively different.
This might be caused by different recording time.
The observations above highly align with the classic sociology research we cited in our original manuscript, which demonstrated that appropriate social distance varies with different agents.

\begin{figure}[H]
\centering
\includegraphics[width=1.0\textwidth]{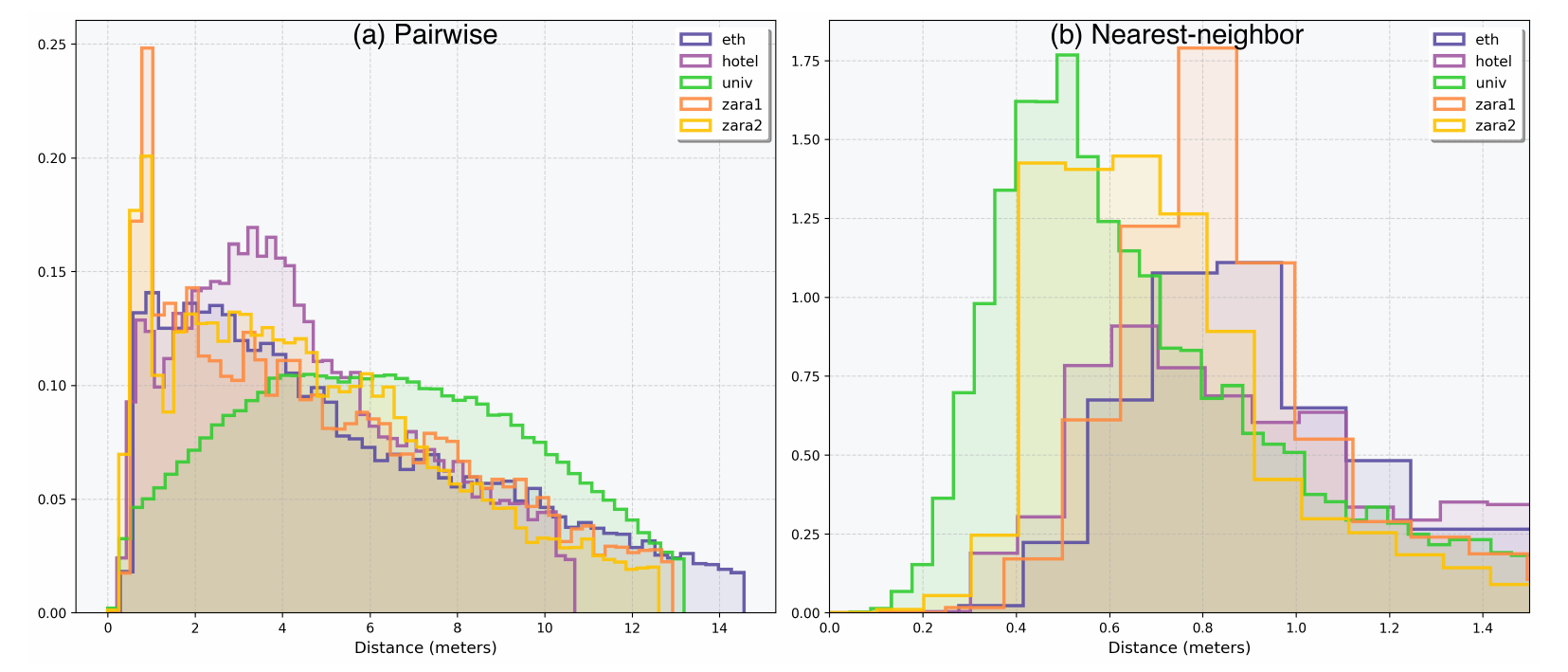}
\caption{
    Distributions of pairwise distance and nearest-neighbor distance across ETH-UCY scenes.
    }\label{fig_distance_distribution}
    
\end{figure}

We also visualize the distributions of both pairwise and nearest-neighbor distances in \FIG{fig_distance_distribution}.
We can observe that both the pairwise distance and nearest-neighbor distance distributions differ significantly across the ETH-UCY subsets.
Although the pairwise distance distributions of zara1 and zara2 are highly similar in (a), their nearest-neighbor distance distributions exhibit substantial differences in (b).
The observations above highly align with the classic sociology research we cited in our Introduction section, which demonstrated that appropriate social distance varies with different agents.

Classic theories in sociology argue that the appropriate social distance is culture-dependent rather than universal \cite{hall1973hidden}. 
In detail, such cultural-dependence could be interpreted from several simple but intuitive perspectives. 
Like people with different personalities usually keep different group-wise motion tendencies, and people from different culturalities also have different acceptable average social distance in a group.

The statistics above provide empirical evidence as to why a fixed threshold mostly fails to capture diverse grouping patterns. 
If we adopt a fixed spatial threshold of $1.0$ meter to determine group affiliation, it will be too strict in the hotel scene, where agents naturally maintain a larger social distance. 
Conversely, in a dense scene like univ, applying the exact same $1.0$m threshold might classify a massive number of out-of-group neighbors as in-group members.
Therefore, the statistics above further validate our motivation that a context-adaptive and agent-specific grouping boundary is necessary to handle the different real-world scenes with diverse spatial distributions.

\end{document}

\end{appendices}

\FloatBarrier

\bibliography{ref.bib}